\documentclass[sigconf]{acmart}

\AtBeginDocument{%
  \providecommand\BibTeX{{%
    Bib\TeX}}}

\setcopyright{acmlicensed}
\copyrightyear{2024}
\acmYear{2024}
\acmDOI{XXXXXXX.XXXXXXX}

\acmConference[CCS '24]{ACM SIGSAC Conference on Computer and Communication Security (CCS'24)}{October 14--18,
  2024}{Salt Lake City, Utah}
\acmISBN{978-1-4503-XXXX-X/18/06}

\usepackage{amsmath,amsfonts}
\usepackage{algorithm,algpseudocode}
\usepackage{multirow}
\usepackage{subfigure}
\usepackage{graphicx}
\usepackage{textcomp}
\usepackage{pifont}
\usepackage{xcolor}
\usepackage{xspace}
\usepackage{comment}
\usepackage{tikz}
\usepackage{listings}

\def\BibTeX{{\rm B\kern-.05em{\sc i\kern-.025em b}\kern-.08em
    T\kern-.1667em\lower.7ex\hbox{E}\kern-.125emX}}

\newcommand{\circlenum}[1]{\tikz[baseline=(myanchor.base)] \node[circle,fill=.,inner sep=1pt] (myanchor) {\color{-.}\bfseries\footnotesize #1};}

\newcommand{\etal}{et al.~}
\newcommand{\ie}{i.e,~}
\newcommand{\eg}{e.g.,~}
\newcommand{\xmark}{\text{\ding{55}}}
\newcommand{\cmark}{\text{\ding{51}}}

\long\def\authornote#1{%
	\leavevmode\unskip\raisebox{-3.5pt}{\rlap{$\scriptstyle\diamond$}}%
	\marginpar{\raggedright\hbadness=10000
		\def\baselinestretch{0.8}\tiny
		\it #1\par}}

\newcommand{\sysname}{{\em SpecGuard}\xspace}
\newcommand{\anomalyagent}{{Attack Agent}\xspace}

\begin{document}

\title{\sysname: Specification Aware Recovery for Robotic Autonomous Vehicles from Physical Attacks}
\author{Pritam Dash}
\email{pdash@ece.ubc.ca}
\affiliation{
  \institution{University of British Columbia}
  \country{Vancouver, Canada}
}
\author{Ethan Chan}
\email{echan@ece.ubc.ca}
\affiliation{
\institution{University of British Columbia}
  \country{Vancouver, Canada}
}
\author{Karthik Pattabiraman}
\email{karthikp@ece.ubc.ca}
\affiliation{
\institution{University of British Columbia}
  \country{Vancouver, Canada}
}

\renewcommand{\shortauthors}{Dash et al.}

\begin{abstract}
Robotic Autonomous Vehicles (RAVs) rely on their sensors for perception, and follow strict mission specifications (\eg altitude, speed, and geofence constraints) for safe and timely operations. 
Physical attacks can corrupt the RAVs’ sensors, resulting in mission failures.  
Recovering RAVs from such attacks demands robust control techniques that maintain compliance with mission specifications even under attacks to ensure the RAV's safety and timely operations.  

We propose \sysname, a technique that complies with mission specifications and performs safe recovery of RAVs. 
There are two innovations in \sysname. First, it introduces an approach to incorporate mission specifications and learn a recovery control policy using Deep Reinforcement Learning (Deep-RL).
We design a compliance-based reward structure that reflects the RAV's complex dynamics and enables \sysname to satisfy multiple mission specifications simultaneously.   
Second, \sysname incorporates state reconstruction, a technique that minimizes attack induced sensor perturbations. 
This reconstruction enables effective adversarial training, and optimizing the recovery control policy for robustness under attacks.    
We evaluate \sysname in both virtual and real RAVs, and find that it achieves $92\%$ recovery success rate under attacks on different sensors, without any crashes or stalls. 
\sysname achieves 2X higher recovery success than prior work, and incurs about 15\% performance overhead on real RAVs.
\end{abstract}




\maketitle

\section{Introduction}
\label{sec:introduction}
Robotic Autonomous Vehicles (RAV) such as drones and rovers are increasingly used in complex industrial scenarios and urban settings~\cite{rav-industry, mining-rv}. 
RAVs use on-board sensors such as GPS, camera, gyroscope,  for perception, and they use specialized algorithms for autonomous navigation. 
This involves a continuous feedback control loop, where the RAV's physical states are derived from sensors and fed to a controller~\cite{feedback-control}. 
To ensure both safety and timely operations, the RAV
autopilot software provides a set of mission specifications (\eg collision avoidance, operational boundary, and geofence constraints)~\cite{ardupilot,px4}.  
The controller regulates the behavior of the RAV as per the mission specifications and computes control commands to execute the mission goals. 
Thus, the correctness of sensors and compliance with mission specifications is critical for the RAV's safety, and for timely execution of the mission.  

Unfortunately, sensors are vulnerable to {\em physical attacks} that can feed malicious signals through physical channels \eg GPS spoofing~\cite{gpsspoofing2}, optical flow spoofing~\cite{opticalspoofing},  gyroscope and accelerometer tampering~\cite{injected-delivered}.
Physical attacks can evade traditional security measures~\cite{ci-choi}, corrupt RAV's perception, and result in erroneous control commands that violate mission specifications. 
This violation poses serious safety risks such as a drone unexpectedly nosediving, or a rover veering off its path leading to a collision. 
These outcomes not only lead to damage to the RAVs and mission failure, but also raise serious safety concerns as RAVs operate in the physical world. 

Many techniques have been proposed to address physical attacks in RAVs. These techniques fall into four categories:  i) sensor redundancy~\cite{ftc-tmr}, ii) resilient control~\cite{ftc-cps, ftc-non-linear}, iii) fail-safe, and (iv)  attack recovery~\cite{pid-piper, srr-choi, recovery-rl, unrocker}. 
Unfortunately, all of these techniques have significant limitations: 
(1) Sensor redundancy is not enough as physical attacks can compromise redundant sensors~\cite{srr-choi}. 
(2) Resilient control techniques are only effective against sensor noise or sensor faults, and cannot handle attacks~\cite{pid-piper}.  
(3) Activating fail-safe (e.g., landing a drone) as a response to attacks is not always safe as the RAV may land in adverse areas or end up falling into the attacker’s hands~\cite{reaper}.
(4) Prior attack recovery techniques~\cite{pid-piper, srr-choi, recovery-rl, recovery-mpc, delorean} focus on a narrow recovery objective such as preventing an immediate crash and restoring the RAV to a set point, but ignore the RAV's mission specifications in the RAV's autopilot software~\cite{px4, ardupilot}.
Disregarding mission specifications such as geofence, altitude constraints or operational boundaries could lead to unsafe recovery for the RAVs, and/or significantly delay the RAVs' missions. 

We propose \sysname, a {\em specification aware recovery} technique for RAVs, that complies with mission specifications even under attacks, and ensures the RAV's safety and timely operations similar to those achieved in attack-free conditions. 
We use {\em Deep Reinforcement Learning} (Deep-RL) in the design of \sysname due to its ability to learn policies to execute complex control tasks, governed by a reward structure~\cite{deeprl-control, drone-racing, deep-rl-paper, rl-book}. 
While various safe Deep-RL approaches have been proposed~\cite{rl-sheilding, rl-cbf, rl-saferl}, they are designed to handle environmental noise rather than attacks. 
Furthermore, the prior safe Deep-RL approaches rely on the correctness of sensors; therefore, they are ineffective under physical attacks. 

\sysname has two main innovations. 
First, it introduces an approach to translate the RAV's mission specifications to a Deep-RL reward function. This enables it to learn a {\em Recovery Control Policy} that complies with mission specifications.
Second, it incorporates {\em State Reconstruction}, a technique that minimizes the impact of sensor perturbations due to attacks.
This enables optimizing the recovery control policy, ensuring its robustness under physical attacks.



Recovering RAVs from physical attacks while satisfying multiple mission specifications is challenging for the following two reasons. 
First, a reward structure is the cornerstone for learning optimal Deep-RL policy~\cite{rl-book, deep-rl-paper}. 
Designing a robust reward structure that reflects the complex dynamics of RAV navigation and control is challenging especially when simultaneously enforcing multiple mission specifications. 
For example, it is crucial to ensure that \sysname steers an RAV around an obstacle to avoid collisions, while maintaining the RAV within a given operational boundary. 
Second, RAV dynamics under attacks cannot be modeled explicitly, because attacks can inject sensor perturbations of varying magnitudes and patterns~\cite{physical-attacks}. 
It is challenging to derive robust control commands for recovery under a diverse attack landscape. 

We address these two challenges in \sysname as follows. 

First, we define the mission specifications in Signal Temporal Logic~\cite{stl}, and design a {\em compliance-based reward structure}. 
This enables \sysname to learn the complex dynamics of the RAV, and derive control commands consolidating 
multiple mission specifications.   
STL allows us to formally define temporal and logical constraints in mission specifications (e.g., stay within the operational boundary $x$, maintain a safe distance $\tau$ from obstacles to avoid collisions). 
Our reward structure quantifies the degree of compliance for each mission specification, and adjusts the reward values accordingly (reward shaping). 
This allows \sysname to learn nuances of RAV's dynamics such as the proximity to obstacles, and make proactive course corrections to prevent collisions while maintaining compliance with other mission specifications.  
In contrast, a simple binary reward structure (\eg satisfied/violated) will limit \sysname's effectiveness in taking proactive measures as it offers no insights into an RAV's complex dynamics. 
\sysname is trained in various scenarios in simulation to derive control commands while complying with multiple mission specifications simultaneously. 



Second, \sysname undergoes adversarial training to optimize the recovery control policy to handle attacks. 
The optimization problem is to satisfy the mission specifications under attacks and safely maneuver the RAVs. 
While prior work in Deep-RL has explored adversarial training to mitigate input perturbation attacks~\cite{adv-agent-training, adv-train-deeprl}, these approaches have limited practicality against physical attacks~\cite{limit-adv-train-deeprl}. 
This is mainly because it is computationally expensive to generate adversarial samples representing the full range of potential sensor perturbations caused by physical attacks. 
This complexity limits the scope of adversarial training, restricting the policy's ability to generalize to diverse physical attacks.  
Instead of generating computationally expensive adversarial samples to cover the full range of physical attacks, we limit the magnitude of sensor perturbations to make adversarial training against physical attacks practical. 
To do so, we incorporate state reconstruction~\cite{checkpointing, delorean} in our adversarial training.
{\em State reconstruction} uses the RAV's historic physical state information to derive robust state estimates under attacks,  
thereby minimizing attack-induced perturbations.




\textbf{\em Contributions.} We make four contributions in this paper. 
\begin{itemize}
    \item Propose a reward structure that reflects the complex dynamics of RAV navigation and control. 
    This enables learning optimal Deep-RL control policy that simultaneously complies with multiple mission specifications of RAVs (\S~\ref{sec:reward-structure}). 
    \item Propose an approach to enhance the robustness of Deep-RL control policies against physical attacks. Our approach minimizes attack-induced sensor perturbations, making policy optimization through adversarial training feasible (\S~\ref{sec:adversarial-training}).
    \item Incorporate the above techniques in an integrated framework, \sysname, to recover RAVs from physical attacks.
    \item Evaluate two designs for specification aware recovery in RAVs (\S~\ref{sec:recovery}).
    (i) Proactive control, where \sysname is the RAV's main controller, proactively correcting control commands for recovery. 
    (ii) Reactive control, where \sysname operates as a secondary controller alongside the RAV’s main controller, and is activated only upon attack detection. 
\end{itemize}

\textbf{\em Results.}
Our main results are as follows: 
(1) Our compliance-based reward structure is highly effective in training \sysname for mission specification compliance, achieving 0 violations of mission specifications in the absence of attacks, unlike the baseline binary reward structure that results in over 25\% violations. 
(2) We find that the Reactive Control approach requires significantly less training time, converging at optimal policy 4X faster than the Proactive Control approach. 
Furthermore, it is 3X faster in recovering RAVs to safe states than Proactive Control. 
(3) Under attacks, \sysname prevents mission specification violations in \emph{87.5\% of the cases} and successfully recovers RAVs in 92.1\% of the cases. 
(4) We compare \sysname with five prior attack recovery techniquess~\cite{srr-choi, recovery-rl, pid-piper, unrocker, delorean}, and find that 
\sysname achieves $5X$ lower mission specification violation rate, and $2X$ higher successful recovery compared to 
 prior work. 
(4) \sysname is also effective in real RAVs achieving $90\%$ mission specification compliance and $>90\%$ recovery success rates, and (5) incurs a maximum overhead of 15.33\% 
 on the real RAVs.


\section{Background and Threat Model}
\label{sec:background}
\subsection{Feedback Control Loop in RAVs}
\begin{figure}[!ht]
    \centering
    \includegraphics[width=0.48\textwidth]{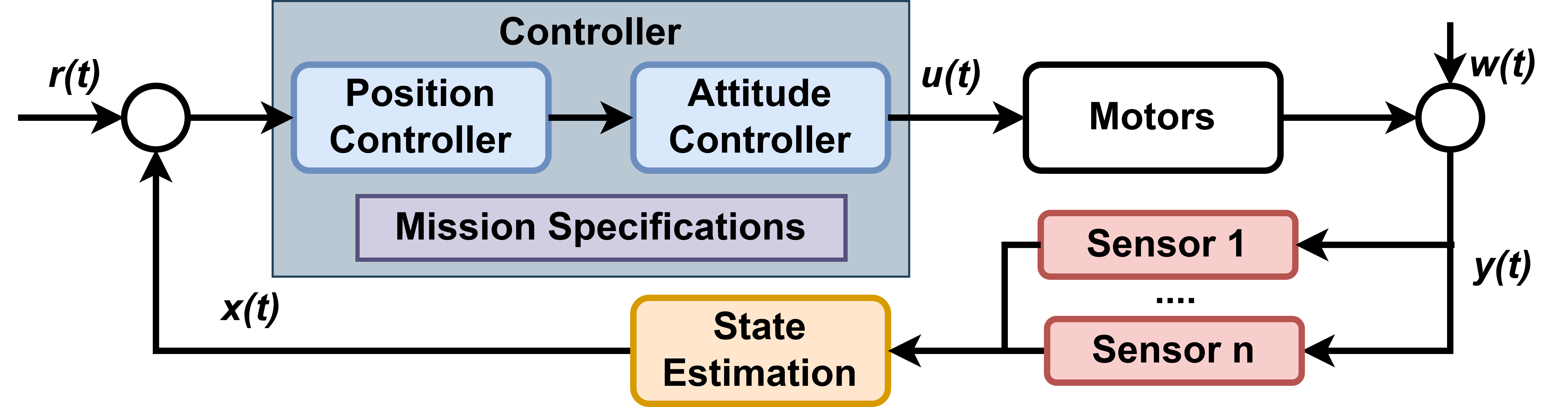}
    \caption{Feedback control loop in RAVs}
    \label{fig:rav-fbc}
\end{figure}

The perception and actuation process in RAVs occurs in a feedback loop as shown in Figure~\ref{fig:rav-fbc}. 
For example, GPS measures the position, gyroscope measures angular velocity, accelerometer measures velocity and acceleration, and camera and LiDAR provide visual perception. 
RAVs estimate their current physical states (\eg position, angular orientation, velocity) using sensor measurements, and they use sensor fusion techniques like
Extended Kalman Filter (EKF)~\cite{ekf} to enhance the physical state estimations by fusing measurements from multiple sensors. 
The controller measures the difference between the reference states and the current states of the vehicle, and calculates appropriate control commands (\eg steering angle, next position) to navigate toward the reference state. 

\subsection{Physical Attacks}
\label{sec:bg-faults-attacks}
Physical attacks manipulate sensor measurements from an external source via physical channels. These include 
(1) Gyroscope and accelerometer measurements, which can be manipulated through acoustic noise injection~\cite{gyroscopespoofing, injected-delivered} leading to unstable attitude control. 
(2) GPS measurements, which can be manipulated by transmitting false GPS signals~\cite{gpsspoofing1, tractor-beam} leading to unstable position control and the RAV deviating from its course. 
(3) Optical flow sensors that track motion, which can be manipulated by injecting fake image beams~\cite{opticalspoofing} leading to unstable motion. 
(4) Magnetometer, which can be manipulated by electromagnetic signal injection (EMI)~\cite{physical-attacks}. 

Physical attacks can be either overt or stealthy~\cite{pid-piper}. 
\textbf{\em Overt attacks} inject large bias in sensor measurements that immediately cause errors in state estimation, steering the RAV to unsafe states (\eg collision or RAV veering off operation boundaries).
In contrast, \textbf{\em stealthy attacks} employ a subtle approach by injecting controlled sensor bias~\cite{stealthy-attacks, stealthy-attacks2, savior}. Over time, the cumulative state estimation error steers the RAV to unsafe states, while evading detection. 



\subsection{Deep-RL in Control Tasks}
\label{sec:bg-deep-rl}
Reinforcement Learning (RL) follows a process of performing a sequence of actions through iterative trial and error to accomplish a goal by learning the optimal action for a given state~\cite{rl-book}.
Deep Reinforcement Learning (Deep-RL) combines reinforcement learning techniques with deep neural networks (DNN). Deep-RL uses DNNs to approximate policy and value functions~\cite{deep-rl-paper}. 

A {\em policy} defines the strategy the agent uses to derive control commands in a given state, while {\em value functions} estimate the expected cumulative reward when following a specific policy. 
The {\em reward function} defines the goals (\eg control objective) and guides the agent's behavior towards achieving the goals, and the {\em reward structure} dictates how and when the rewards are assigned. 

Deep-RL involves interaction between an agent in an environment. 
At each step, the agent observes state $s_t$, performs an action $a_t$ to transition to the next state $s_{t+1}$, and receives a reward $r_t$. 
The reward indicates the effectiveness of action $a_t$ in achieving the control objective.  
The policy $\pi_\theta$ maps state $s$ to a probability distribution over actions $a$.  $\pi_\theta(a|s)$ is iteratively updated to maximize the reward. 
The agent repeats the above steps until it converges on an optimal policy $\pi^*_\theta$ to select the best action and maximize the expected sum of rewards (Equation~\ref{eqn:deep-rl-policy}).
\begin{equation}
    \pi^*_\theta = \arg\max_{\pi_\theta} ~\mathbb{E}^\pi_\theta~[\Sigma_{0}^{t} r_t(s_t, a_t, s_{t+1})]
    \label{eqn:deep-rl-policy}
\end{equation}


\subsection{Threat Model}
\noindent
{\em \textbf{Adversary's Capabilities.}} 
We consider a white-box attack setting where the adversary has knowledge of the target RAV's hardware and software.  
The adversary can launch physical attacks that maliciously perturb the RAV's sensors' measurements through physical channels, and can optimize the attack parameters for maximum impact. 
The adversary can launch physical attacks at any location during the RAVs' missions. 
Furthermore, the adversary can launch stealthy attacks while evading detection for as long as possible~\cite{stealthy-attacks}.  

We consider only single-sensor attacks, as performing signal interference against multiple sensors simultaneously in a moving RAV is challenging~\cite{poltergeist}.
Further, attacks that exploit vulnerabilities in the RAV's software components and communication channels are out of our scope as they can be handled by existing techniques~\cite{cfi, cfh}. 
Moreover, \sysname executes in the RAV's trusted computing base to minimize exposure to such threats.
Finally, sophisticated LiDAR and RADAR~\cite{lidar-attack, radar-spoofing} spoofing attacks, and camera projection attacks~\cite{cam-projection-attack} are out of our scope.  
These attacks are designed against advanced driving assistance systems (ADAS) and object detection models in automobiles, and not for RAVs. 


\noindent
{\em \textbf{Assumptions}}. 
We assume that an adversary can neither tamper with the Deep-RL training environment nor 
the training process.   
These are reasonable assumptions as Deep-RL training is typically carried out in a simulated environment that is isolated from external modifications~\cite{ai-safety}. Consequently, attacks such as reward hacking~\cite{reward-hacking}, reward poisoning~\cite{reward-poisoning}, and policy induction~\cite{policy-induction}, which modify the training environment, are outside our scope. 



\section{Motivation and Approach}
\label{sec:motivation}
\subsection{Mission Specification and Safety in RAV}
An RAV mission plan typically contains waypoints and the mission path. In addition, RAV autopilot software such as PX4~\cite{px4} and ArduPilot~\cite{ardupilot} provide  mission specifications such as geofence, operational boundary, collision avoidance, altitude constraints, speed constraints. 
Mission specifications play a crucial role in ensuring the safety and timely operation of RAVs.  
For example:   
(i) Geofence creates a virtual barrier and ensures the RAV does not enter restricted airspace or areas with collision risks. 
(ii) Operational boundary constraints ensure that the RAV remains within a designated boundary. 
(iii) Minimum and maximum altitude constraints ensure the RAV flies above obstacles and complies with airspace regulations. 
(iv) When collision avoidance is activated, the RAV detects potential collision risks using rangefinder sensors (\eg ultrasonic, LiDAR, or infrared) and re-routes to avoid the collision. 

The controller receives the RAV's physical states, constantly monitors for violations of mission specifications, and derives control commands to execute mission goals while complying with those specifications. 
It applies trajectory corrections if the RAV deviates from the intended mission plan or violates mission specifications. 

\subsection{Motivating Example}
\label{sec:motivation-ex}
Prior recovery techniques fall into two categories: 
(1) {\em Sensor measurement correction}: These techniques use a learned model of sensors to minimize corruption in sensor measurements due to attacks, and employ the original controller to derive control commands~\cite{srr-choi, unrocker}. 
(2) {\em Control command correction}: These techniques either learn the vehicle's physical dynamics to design a specialized recovery controller~\cite{pid-piper, recovery-rl, recovery-mpc} to steer the RAV towards estimated future states, or bound the values of safety-critical control parameters~\cite{scvmon} to apply corrective control commands and recover RAVs. 

\begin{figure}[!ht]
	\centering
	\subfigure[]{
		\includegraphics[width=0.45\linewidth]{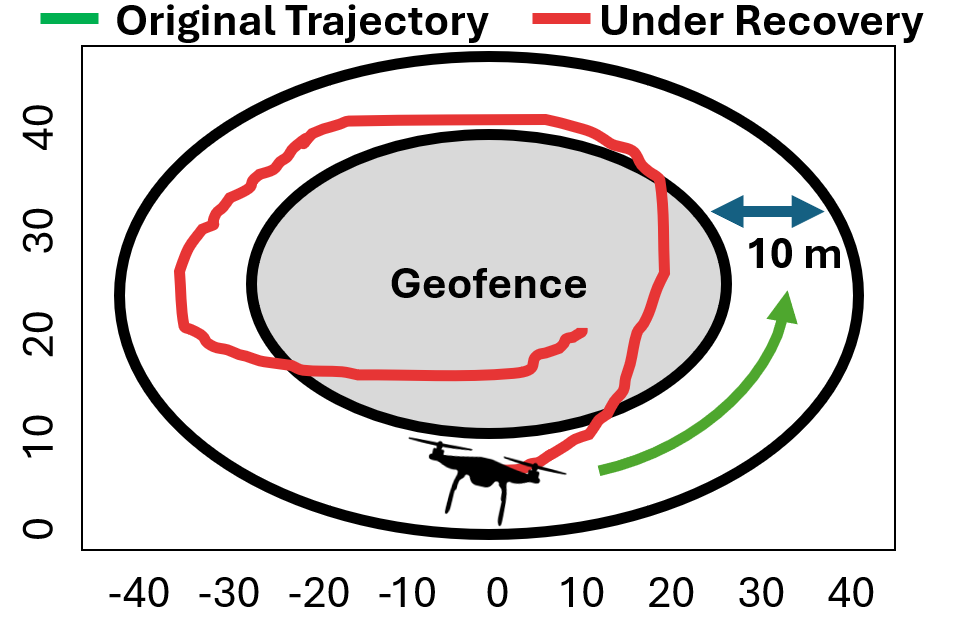} 
		\label{fig:motivation-1}
	}
	\subfigure[]{
		\includegraphics[width=0.45\linewidth]{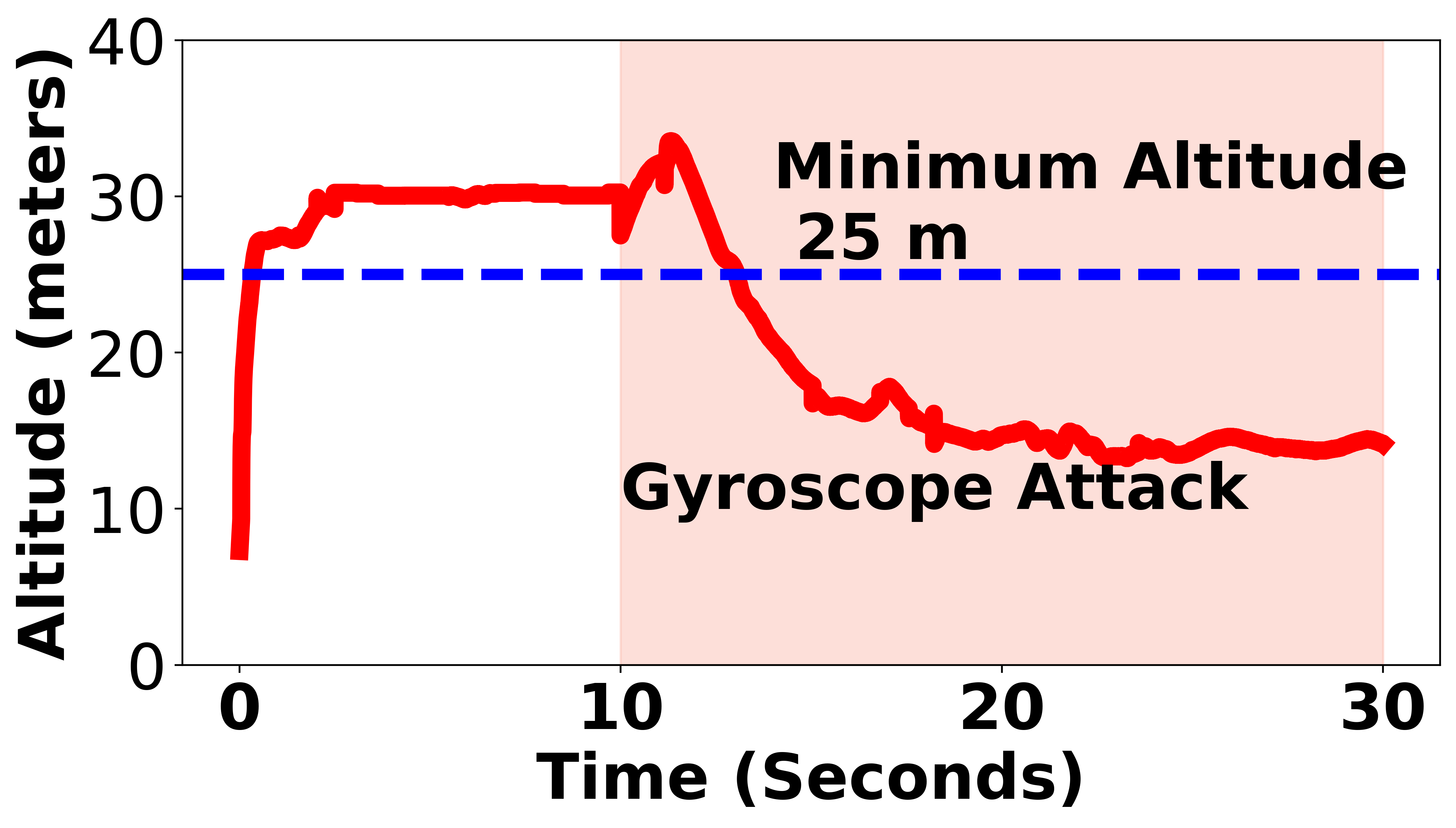}
		\label{fig:motivation-2}
	}
	\caption{Mission specifications violations resulting in unsafe recovery. 
		The drone violated
		(a) Operational boundary and geofence specifications. 
		(b) minimum altitude specification.}
	\label{fig:motivation}
\end{figure}

To demonstrate the limitations of the prior attack recovery techniques, we conducted two experiments on PXCopter (details in \S~\ref{sec:experimental-setup}). 
In the first experiment, a drone covers a circular path as shown in Figure~\ref{fig:motivation-1}, at an altitude of 30m. 
As part of the mission plan three mission specifications are provided: 
($S_1$) do not violate a geofence, 
($S_2$) do not veer off the operating boundary (10m), and 
($S_3$) maintain altitude above 25m. 
The drone is equipped with the Fault Tolerant Control (FTC)~\cite{recovery-rl} recovery method.  
We consider FTC as a representative of prior methods - a detailed comparison with four additional techniques is presented in 
\S~\ref{sec:res-comparison-recovery}.
FTC uses a control policy to derive corrective control commands. 

Figure~\ref{fig:motivation-1} shows the results of the first experiment. 
At t=10s we launch a GPS spoofing attack (attack parameters in Table~\ref{tab:rv-sensors}). 
In response, FTC derives control commands to recover the drone. 
While FTC prevented an immediate crash, it violated the geofence ($S_1$) and operating boundary ($S_2$) specifications. 
The red line in Figure~\ref{fig:motivation-1} shows the drone's trajectory during recovery. 
These violations may result in collisions in scenarios with obstacles. 

In the second experiment, we repeat the same mission with the same mission specifications, 
This time we attacked the gyroscope sensor starting t=10s, which forced the drone to nosedive. 
The red line in Figure~\ref{fig:motivation-2} shows the drone's altitude. 
While FTC again prevented an immediate crash, the drone's altitude dropped to $15m$, and
FTC maneuvered the drone at an altitude of 15m, 
violating $S_3$. 
This violation increases the risk of collisions in scenarios with obstacles below the set safe altitude ($S_3$: minimum altitude specification).

Thus, prior recovery techniques only mitigate the immediate disruptions caused by an attack and prevent crashes. They do not enforce mission specifications such as geofences, operational boundaries, and altitude constraints. 
Thus, the narrow focus of prior techniques leads to unsafe recovery, endangering the RAV's safety. 

\subsection{Research Gap and Challenges}
\noindent
\textbf{\em Research Gap}. 
We establish four criteria essential for successful attack recovery in RAVs:
(1) Mitigate disruptions: prevent disruptions such as a crash, or deviations from the set path. 
(2) Complete mission: ensure the RAV can complete its task. 
(3) Timeliness: minimizing mission delays due to attacks. 
(4) Safe recovery: ensuring mission specification compliance, no collisions, crashes, and stalls.

Table~\ref{tab:prior-work} summarizes the capabilities of prior attack recovery techniques. 
Prior methods such as Software Sensor based Recovery (SSR)~\cite{srr-choi}, Fault Tolerant Policy (FTC)~\cite{recovery-rl}, SCVMON~\cite{scvmon}, and SemperFi~\cite{semperfi} focus on a narrow recovery objective \ie preventing crashes and drastic deviations.
Techniques such as PID-Piper~\cite{pid-piper}, DeLorean~\cite{delorean}, and UnRocker~\cite{unrocker} focus on a slightly broader recovery scope \ie enabling RAVs to complete the mission despite malicious actions as well as preventing crashes. 
Thus, the prior techniques focus on a narrow recovery objective, making these techniques applicable in only a limited set of scenarios. 

\begin{table}[!ht]
\centering
\footnotesize
\caption{Comparison of \sysname with prior techniques. Safe Recovery means mission specification compliant recovery without any crashes and stalls.}
\begin{tabular}{l|c|c|c|c}
\hline
\multicolumn{1}{c|}{\textbf{\begin{tabular}[c]{@{}c@{}}Recovery \\ Techniques\end{tabular}}} & \multicolumn{1}{c|}{\textbf{\begin{tabular}[c]{@{}c@{}}Mitigate \\ Disruptions\end{tabular}}} & \multicolumn{1}{c|}{\textbf{\begin{tabular}[c]{@{}c@{}}Complete\\ Mission\end{tabular}}} & \multicolumn{1}{c|}{\textbf{\begin{tabular}[c]{@{}c@{}}Timely\\ Recovery\end{tabular}}} & \multicolumn{1}{c}{\textbf{\begin{tabular}[c]{@{}c@{}}Safe \\ Recovery\end{tabular}}} \\ \hline
SSR~\cite{srr-choi}         & $\cmark$  & $\xmark$  & $\xmark$      & $\xmark$   \\ 
FTC~\cite{recovery-rl}      & $\cmark$  & $\xmark$  & $\xmark$      & $\xmark$   \\ 
PID-Piper~\cite{pid-piper}  & $\cmark$  & $\cmark$  & $\xmark$      & $\xmark$   \\ 
SemperFI~\cite{semperfi}    & $\cmark$  & $\xmark$  & $\xmark$      & $\xmark$   \\ 
UnRocker~\cite{unrocker}    & $\cmark$  & $\cmark$  & $\xmark$      & $\xmark$   \\ 
DeLorean~\cite{delorean}    & $\cmark$  & $\cmark$  & $\xmark$      & $\xmark$   \\ 
\textcolor{blue}{\textbf{\sysname}} & \textcolor{blue}{$\cmark$}  & \textcolor{blue}{$\cmark$}  & \textcolor{blue}{$\cmark$}  & \textcolor{blue}{$\cmark$}   \\ \hline

\end{tabular}
\label{tab:prior-work}
\end{table}

\smallskip
\noindent
\textbf{\em Our Approach.}
RAVs operate under strict constraints such as geofence, operational boundaries, collision avoidance, altitude constraints, and speed constraints.
These constraints are provided as mission specifications along with the mission plan in RAV autopilot software e.g., PX4~\cite{px4} and ArduPilot~\cite{ardupilot}. 
Because RAVs rely on mission specifications to safely operate in attack-free conditions, it is important to comply with the specifications when taking recovery actions. 
{\em Thus, a specification-aware recovery technique is required to ensure the safe and timely operation of RAVs under physical attacks.}

Prior recovery techniques~\cite{srr-choi, recovery-rl, pid-piper, unrocker, delorean} cannot dynamically adjust actuator commands based on these constraints. 
Extending these techniques to include constraints is non-trivial. For example, incorporating constraint outcomes (satisfaction/violation) as input signals is not enough, as it does not ensure that these outcomes influence actuator commands to satisfy mission specifications.

\smallskip
\noindent
\textbf{\em Challenges.}
The challenges in designing \sysname are: 

\circlenum{C1} {\em Incorporating mission specifications into recovery control}: 
How to derive recovery control commands for maneuvering the RAV that comply with the mission specifications? 
To achieve this \sysname has to incorporate the mission specifications in the recovery control policy learning process, and ensure that multiple mission specifications can be satisfied simultaneously. 

\circlenum{C2} {\em Reliable state estimation under attack}: How to minimize attack induced sensor perturbations and obtain reliable state estimations (position, attitude) of the RAV under attack?
This is essential for \sysname to enforce mission specifications during recovery.

\smallskip
\noindent
\textbf{\em Scope of this work.}
Our focus is recovery from physical attacks, and hence attack detection and attack diagnosis are out of our scope. 
Prior work has proposed methods to  detect~\cite{ci-choi, savior, avmon}, and localize sensor anomalies in RAVs~\cite{delorean} through diagnosis. 
Our approach is designed to seamlessly integrate with prior attack detection and diagnosis techniques, and perform recovery under attacks.

\subsection{\sysname Overview}
\label{sec:overview}

We design \sysname, a specification aware attack recovery technique for RAVs that addresses all four recovery criteria in Table~\ref{tab:prior-work}.   
It learns a {\em Recovery Control Policy}, and uses it to maneuver RAVs under attacks while complying with mission specifications, thus, ensuring safe and timely recovery. 
\sysname integrates with existing attack detection and diagnosis techniques~\cite{ci-choi, savior,delorean} that detect and identify the compromised sensor respectively. 

\smallskip
\noindent
\textbf{\em Mission Specifications.}
We manually convert the mission specifications provided in RAVs' autopilot software~\cite{px4, ardupilot} to Signal Temporal Logic (STL)~\cite{stl}, and incorporate them into \sysname. 
STL allows us to formally express the constraints specified in mission specifications in natural language. 
We design a reward structure for recovery control policy training by incorporating the STL formulated mission specifications.
As a result, \sysname learns a policy to derive optimal control commands that comply with the mission specifications (addressing \circlenum{C1}). 
\emph{Our technique can work with any set of mission specifications  provided they are expressed using STL.}

\smallskip
\noindent
\textbf{\sysname~{\em Training.}}
We formulate the control problem as a Markov Decision Process (MDP). An MDP is defined by a state space $X$, action space $U$, a scalar reward function $R(x_t,x_{t+1})$, and the transition probability $P(x_{t+1}|x_t, u_t)$. 
\sysname selects an action $u_t$ using a policy $\pi_\theta(u_t|x_t)$ and receives a reward $r_t$. 
The training objective is to find an optimal policy $\pi^*_\theta$ that maximizes rewards.

\sysname undergoes a two-phase training in simulated environments using Deep-RL. 
First, it is trained in attack-free scenarios to learn the vehicle dynamics: $x_{t+1} = f(x_t, u_t) + w_t$, where $x_t$, $u_t$, $w_t$ are RAVs current states, control commands, and environmental noise respectively. 
In this phase, \sysname learns a recovery control policy $\pi_\theta(u'_t|x'_t)$ that derives control commands $u'_t$ complying with the mission specifications.
Second, \sysname undergoes adversarial training in the presence of simulated attacks. 
We use a {\em State Reconstruction} technique~\cite{delorean, recovery-lp, checkpointing} (details in \S\ref{sec:adversarial-training}) to minimize the impact of attack-induced sensor perturbation. 
This makes adversarial training feasible against physical attacks. 
In this phase, \sysname optimizes the policy $\pi_\theta(u'_t|x'_t)$ to derive control commands ensuring both safe recovery of RAVs under attacks and continued compliance with mission specifications (addressing \circlenum{C2}).  

\section{\sysname: Design}
\label{sec:design}
In this section, we present the five steps involved in \sysname design: 
First, we explain how mission specifications are defined as STL. 
Next, how we monitor conditions in STL. 
Then, we present our compliance based reward structure, 
followed by \sysname's two training phases.  
Finally, we present two variants of \sysname for attack recovery - proactive and reactive control based approaches. 

\subsection{Mission Specifications as STL}
\label{sec:ms-stl}

We use Signal Temporal Logic (STL) to formally define the {\em mission specifications}. 
STL is a formal language for expressing temporal properties of a system~\cite{stl}. 
Mission specifications include conditions such as maintaining a safe distance from obstacles, operating within a specific bound $a$, 
or reaching a waypoint within a time frame. 
Traditional temporal logic forms such as Linear Temporal Logic (LTL) and Computational Tree Logic (CTL) do not support expressing conditions over continuous signals that hold within a time bound~\cite{monitoring-tl}.  
In contrast, STL allows expressing conditions over continuous signals that hold (only) within a given time bound. 
This capability of STL is crucial for defining mission specifications provided in the RAV autopilot software such as reaching a waypoint within a certain time.
Furthermore, LTL and CTL express boolean outcomes of conditions, i.e., they focus on whether a condition is true or false without accommodating the continuous nature of the signals. 
In contrast STL expresses boolean outcomes over real-valued continuous signals, which is useful for defining mission specifications over dynamic RAV trajectories~\cite{why-stl}. 
Thus, STL is suitable for defining nuanced and time-bounded conditions~\cite{stl-robustness}. 
We explain temporal logic in detail in Appendix~\ref{appn:temp-logic}.


\smallskip
\noindent
\textbf{\em STL Notations.} We use the following temporal operators to represent temporal properties of mission specifications~\cite{stl}: 
$G$: properties that should always hold true, and $F$: properties that should eventually hold true. 
Finally, the boolean operators $\lnot$ and $\land$ are used to express negation and conjunction respectively. 

We illustrate how we define mission specifications in STL considering the following RAV mission specifications as examples (where $a$ and $b$ are constant parameters): 
$M_1$: avoid collisions, 
$M_2$: do not veer off a designated bound $a$.  
$M_3$: maintain altitude above $b$. 
$M_4$: navigate through all the given waypoints $[w_1..w_n]$ within a time bound. 
We use the following template to express mission specifications in STL:
\texttt{<temporal operator(condition(parameters))>}, 
where the temporal operator defines the temporal properties, the condition expresses the constraints to be satisfied, and parameters are specific values within the condition. 

Table~\ref{tab:stl-examples} shows the above mission specifications in STL. 
$M_1$ is expressed as $S_1: G(obstacle\_distance(x_t) > \tau)$, where $x_t$ is the RAV's current state and $\tau$ is a threshold.
To ensure collision avoidance, we add a condition in the STL formulation of  $M_1$ to maintain a safe distance $\tau$ from obstacles in the surroundings. 
We specify the condition in $S_1$ as $obstacle\_distance(x_t) > \tau$. 
This approach is more robust than simply checking for a binary collision status (true or false). 
$M_1$, $M_2$, and $M_3$ should always hold true, thus, expressed with operator $G$. 
$M_4$ is expressed using operator $F$ because it is expected to be true at some point in the future i.e., the RAV must eventually cover all the waypoints completing the mission.

\begin{table*}[!ht]
\centering
\footnotesize
\caption{Mission Specifications in Signal Temporal Logic (STL). The RAV must satisfy $S_1 \land S_2 \land S_3 \land S_4.$}
\begin{tabular}{l|l|l|l|l|l}
\hline
\textbf{ID} & \textbf{Mission Specification}        & \textbf{Operator} & \textbf{Condition} $f(c)$          & \textbf{Parameters} & \textbf{Mission Specification in STL}                 \\ \hline
$S_1$       & Avoid collisions                      & $G$               & $obstacle\_distance(x_t)$   & $\tau$              & $G(obstacle\_distance(x_t) > \tau)$       \\ 
$S_2$       & Do not veer off bound - $a$           & $G$               & $checkCurrentPosition(x_t)$ & $a$                 & $G(checkCurrentPosition(x_t) < a)$        \\ 
$S_3$       & Maintain altitude above $b$         & $G$               & $altitude(x_t)$             & $b$                 & $G(altitude(x_t)> b$  \\ 
$S_4$       & Navigate via waypoints $w_1..w_n$ & $F$               & $dist(w_i, x_t)$            & $[w_1..w_n], R$     & $F_{[t_{i_1}, t_{j_1}]}(dist(w_1, x_t) < R)..\land F_{[t_{i_n}, t_{j_n}]} dist(w_n, x_t) < R)$     \\ \hline
\end{tabular}
\label{tab:stl-examples}
\end{table*}

\subsection{Monitoring Conditions in STL}
\label{sec:monitor-ms}
We use the existing functions and APIs in the RAV's autopilot software e.g., PX4~\cite{px4} and ArduPilot~\cite{ardupilot} to monitor conditions in the STL formulated mission specifications (henceforth, referred to as STL specifications). 
For example, in $M_1$ to ensure collision avoidance, its STL formulation $S_1$ requires monitoring the distance between RAV's current state $x_t$ and obstacles in the surroundings. 
We use the following function in PX4 defined in \texttt{pos\_control} module
to monitor distance from obstacles that calculates the distance from obstacles: 
\texttt{CollisionPrevention::obstacle\_distance()}. 

Specifications expressed with the $F$ operator are monitored within a specific time interval $[t_i,t_j]$, and it must be true at least once in the given interval.
Specifications expressed with the $G$ operator are monitored continuously, and must be true throughout the mission. 
Furthermore, specifications expressed with the $G$ operator within a time bound must be true in the entire time frame.

\subsection{Compliance-based Reward Structure}
\label{sec:reward-structure}
\noindent
\textbf{\em Reward Function.}
The reward function reflects either the satisfaction or violations of the STL specifications.
By mapping STL specifications into a reward function, \sysname incorporates the temporal and logical constraints specified in the mission specifications.
The reward function assigns a reward value at every timestep $t$ indicating compliance or violation of STL specification during \sysname's training. 
Finally, a cumulative reward expresses how well \sysname satisfies all the mission specifications overall. 

\smallskip
\noindent
\textbf{\em Reward Structure.}
Reward structure defines a numerical scoring system to assign dynamic and intermediate rewards for efficient Deep-RL policy training~\cite{reward-shaping}.
Assigning rewards based on binary outcomes (satisfied/violated) of each STL specification leads to sparse reward conditions (infrequent rewards)~\cite{reward-shaping}, e.g., a reward of 1 will be given when an STL specification is satisfied, and 0 only when the specification is violated. 
This approach fails to capture the complexities of RAV dynamics, such as distances from obstacles or how close the RAV is to breaching its operational boundaries. 
These factors significantly influence optimal control commands. 
This issue becomes even more pronounced when enforcing multiple mission specifications simultaneously. Consequently, \sysname may fail to comply with multiple mission specifications simultaneously.

\begin{figure}[!ht]
    \begin{minipage}{0.24\linewidth}
        \begin{equation}
        \rho(S) = \frac{1}{1+e^{-S}}
        \label{eqn:sigmoid}
        \end{equation}
    \end{minipage}
    \hspace{0.5em}
    \begin{minipage}{0.35\linewidth}
        \subfigure[]{\includegraphics[width=\linewidth]{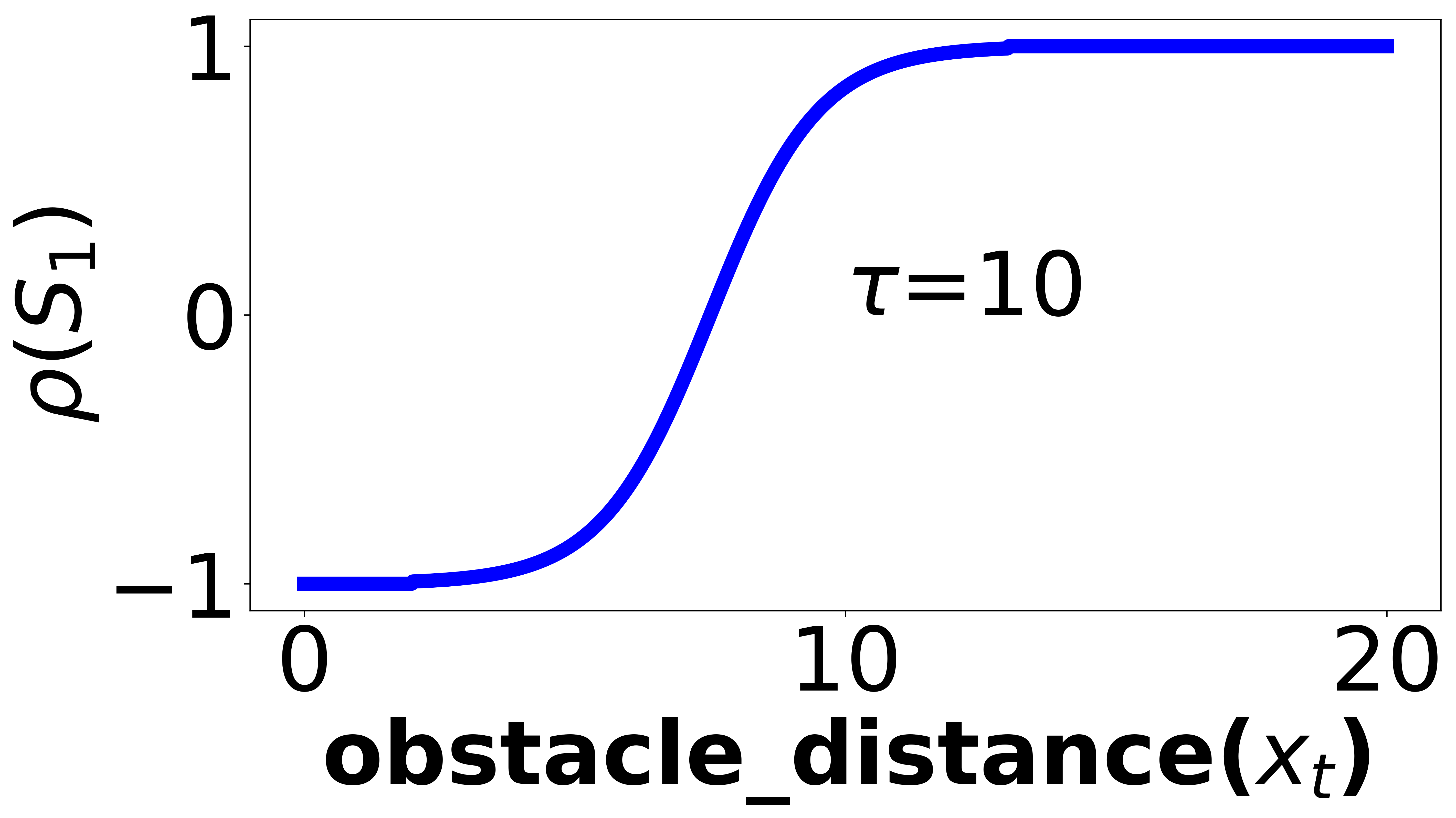}\label{fig:stl-s1}}
    \end{minipage}%
    \begin{minipage}{0.35\linewidth}
        \subfigure[]{\includegraphics[width=\linewidth]{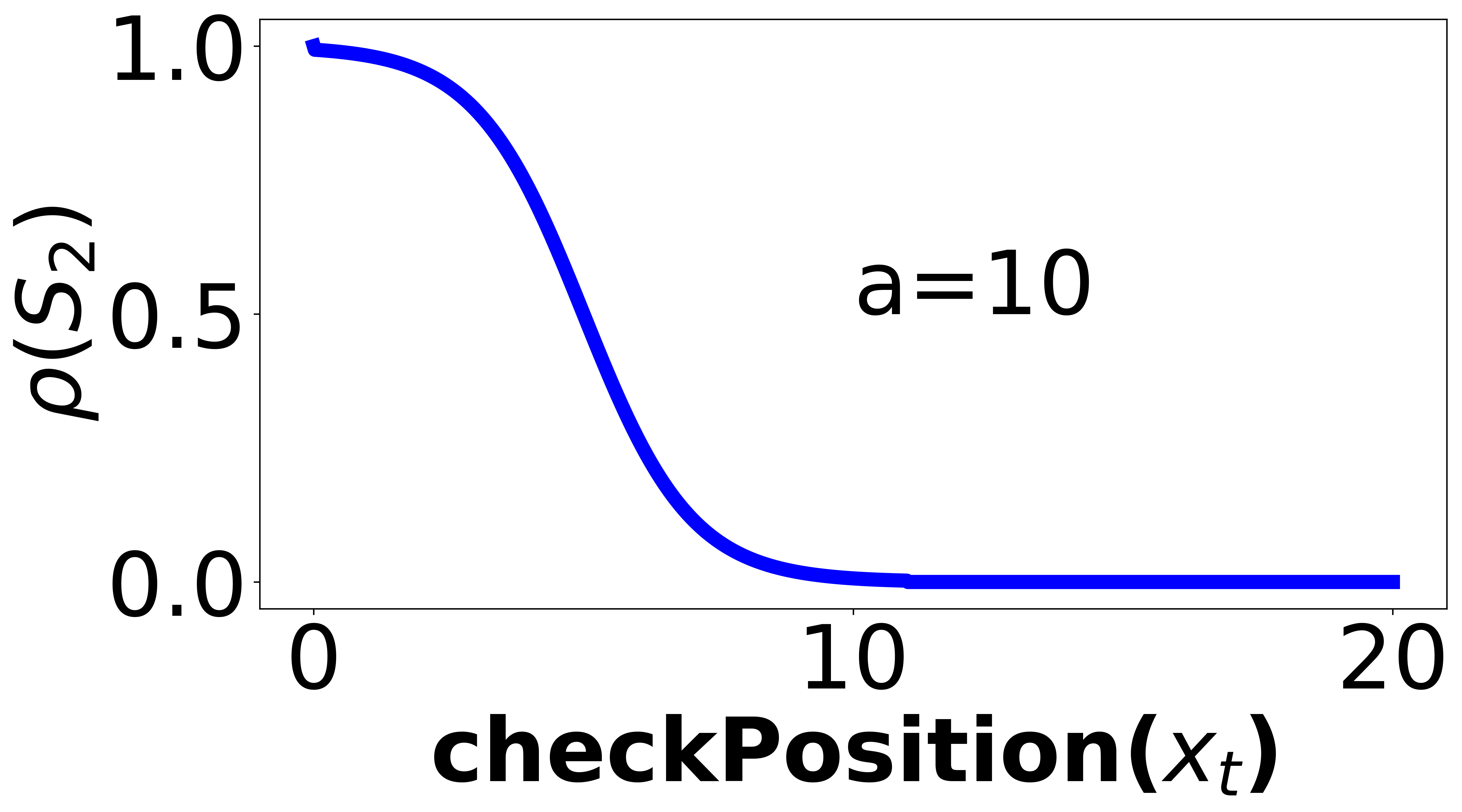}\label{fig:stl-s2}}
    \end{minipage}
    \caption{Sigmoid function based reward assignment for $S_1, S_2$.}
    \label{fig:stl-robustness}
\end{figure}

We design \sysname with a {\em Compliance-based} reward structure inspired by the concept of STL robustness~\cite{stl-robustness} that addresses the limitations of binary reward structures. 
Our reward structure reflects the complexities of the RAV's dynamics. 
This enables \sysname to learn a policy to formulate optimal control commands to satisfy multiple mission specifications. 
Our reward structure adjusts the reward values (reward shaping) based on the degree to which STL specifications are satisfied. 
We design a Sigmoid function (Equation~\ref{eqn:sigmoid}) based framework to design reward functions and perform reward shaping, where the outcome of the reward function $\rho$ quantifies the degree to which STL specifications are satisfied. 

Our Sigmoid function-based framework for 
reward shaping provides the following advantages: 
(i) It bounds the reward values in a range providing a good balance between exploration and exploitation. 
(ii) The Sigmoid function can be tailored to represent the conditions within any STL specification. 
(iii) It provides a continuous and smooth transition of reward values between satisfaction and violation of mission specifications enabling fine-grained adjustments to control commands.
iv) It effectively balances the satisfaction of multiple mission specifications simultaneously.

We explain how our reward structure enables satisfying multiple specifications with the following example. 
Figure~\ref{fig:stl-robustness} visualizes the Sigmoid based reward assignment for STL specifications $S_1$ and $S_2$.
As shown in Figure~\ref{fig:stl-s1}, if the RAV's distance from the obstacle is $> \tau$, the $\rho(S_1)$ is set to 1, indicating a high degree of satisfaction of the specification $S_1$. As the distance from the obstacle becomes $< \tau$, the  $\rho(S_1)$ gradually becomes negative indicating a violation of specification $S_1$. 
Similarly, in Figure~\ref{fig:stl-s2}, if the RAV's deviation is $0$, $\rho(S_2)$ is set to 1.
As the deviation increases, $\rho(S_2)$ indicates a low degree of satisfaction of $S_2$, and when the deviation is $> a$, $\rho(S_2)$ becomes 0, indicating violation of $S_2$. 
Thus, the rewards for $S_1$, and $S_2$ (reward shaping) is derived based on $\rho({S_1})$, and $\rho({S_2})$. 



Our compliance-based reward structure enables learning a recovery control policy that dynamically makes fine-grained adjustments to control commands to balance multiple mission specifications. 
When satisfying one specification conflicts with another, our reward structure ensures that the recovery control policy prioritizes the most critical specification. 
For example, if the RAV is close to violating $S_1$ (collision risks), \sysname derives a control command that decreases the immediate reward for $S_2$ to prevent violation of $S_1$.  
Thus, \sysname optimizes the control commands for individual STL specifications, while simultaneously satisfying multiple mission specifications.
This is achieved by maximizing the cumulative reward across all STL specifications.

The following equations show the Sigmoid functions $\rho(S_1)$ and $\rho(S_2)$ for calculating rewards for STL specifications $S_1$ and $S_2$. 
Where $f(c)$ denotes the outcome of the conditions in the STL specifications, and $a$ and $\tau$ are the parameters in the STL specifications as shown in Table~\ref{tab:stl-examples}. 
The reward functions $\rho(S_1)$ and $\rho(S_2)$ represent the quantitative measures of the degree of satisfaction of the STL specifications. 
Finally, $k_1,..,k_n$ are constants for constructing a smooth piecewise function.

\begin{align}
& \rho({S_1}) = \left\{\begin{matrix} \frac{2}{1+e^{-k_1(f(c)-\tau)}}-1,   & 0<f(c)<\tau\\ 1 & f(c)>\tau \end{matrix}\right.
\\
& \rho({S_2}) = \left\{\begin{matrix} \frac{1}{1+e^{k_2(f(c)-a)}}, & 0<f(c)<a\\ 0    & f(c)>a \end{matrix}\right. 
\end{align}

Finally, a cumulative reward $r_t$ that indicates \sysname's overall compliance is calculated by summing all the reward values:
$r_t=\rho(S_1)+\rho(S_2)+ .. +\rho(S_n)$.
The algorithm for calculating cumulative reward considering compliance with each STL specification in Table~\ref{tab:stl-examples} is presented in Appendix~\ref{appn:reward-function}.
The reward functions for all the STL specifications in Table~\ref{tab:stl-examples} are presented in Appendix~\ref{appn:reward-func}.

\subsection{Phase 1: Training for Mission Specification Compliance}
\label{sec:recovery-control-policy}

Recall that \sysname undergoes a two-phase training. 
In the first Phase, it is trained to learn a policy $\pi_\theta(u'_t|x'_t)$ that derives control commands $u'_t$ in compliance with the mission specifications. 
\sysname's inputs are represented as a vector $x'_t$ shown in Equation~\ref{eqn:input}.
The inputs include the RAV's current state: position $(x, y, z)$, velocity $(\dot{x}, \dot{y}, \dot{z})$, acceleration $(\ddot{x}, \ddot{y}, \ddot{z})$, and angular orientation $(\phi, \theta, \psi)$. 
Given the input $x'_t$, \sysname outputs $u'_t \in U$, where $U$ is a set of $\{+x, -x, +y, -y, +z, -z\}$. 
These actions dictate how the RAV's trajectory should be adjusted along its relative axis. 
For example, $+x$ indicates forward movement along the RAV's x-axis relative to its current position, and $-x$ indicates backward movement along the same axis.
The output $u'_t$ is then translated to low-level actuator commands (\eg thrust force) - details are in Appendix~\ref{appn:act-commands}.
At each timestep $t$, \sysname's input and output are calculated as below:

\begin{equation}
    x'_t = \{x, y, z,\dot{x}, \dot{y}, \dot{z}, \ddot{x}, \ddot{y}, \ddot{z}, \phi, \theta, \psi\}
    \label{eqn:input}
\end{equation}
\begin{equation}
    u'_t = \arg \max_u~ \pi_\theta (u'_t|x'_t) \cdot R(u'_t,x'_t), \: u'_t \in U
    \label{eqn:output}
\end{equation}

\sysname is trained in various simulated environments. 
Note that $x'_t$ denotes the inputs to \sysname, while $x_t$ denotes the RAV's current state. 
The state transition dynamics of the RAV is denoted as $P(x_{t+1}|x_t, u_t)$. 
As a result of the action $u'_t$, the RAV transitions to the next state $x_{t+1}$, and \sysname receives a reward $r_t$.
The function $R(.)$ assigns a cumulative reward $r_t$ that shows how the action $u'_t$ complies with the mission specifications.
The reward functions play a pivotal role in ensuring \sysname adapts its policy as per the mission specifications. 
In each rollout, the cumulative discounted reward is calculated as $R = \sum_{t=0}^{T} \gamma^{t-1}r_t(x'_t, u'_t)$ at the end of the episode at time $T$. 
The goal is to optimize the policy $\pi_\theta$ to maximize the average total reward ($J$) received when actions $u'_t$ are sampled from the policy $\pi_\theta$, calculated as $J = E_{u'_t \sim \pi_\theta} [R]$.  

Note that RAV autopilot software typically uses a cascading controller design for position control and attitude control.  
However, \sysname is designed to subsume these two as a single control entity (\ie Recovery Control Policy).  

\subsection{Phase 2: Adversarial Training}
\label{sec:adversarial-training}

In the second phase of training, \sysname undergoes adversarial training to enhance the robustness of the policy learned in the first phase. 
This is because policies trained in non-adversarial conditions are typically not robust against attacks~\cite{adv-agent-training, adv-train-deeprl}. 

Physical attacks can introduce biases of various magnitudes and patterns to the RAV's sensor measurements~\cite{physical-attacks, stealthy-attacks}. 
Thus, modeling the attacks explicitly for adversarial training is challenging due to the large state space.
For example, GPS spoofing can inject biases up to $\pm$ 50m along all of the RAV's position axes ($x, y, z$). 
Generating adversarial state samples by manipulating RAV's position, ranging from $y\pm 0$ to $y\pm 50$ for each position axis in various vehicle dynamics (\eg take off, cruising at a constant altitude, turning, landing, etc.), and repeating this process for attacks against all the sensors, imposes significant computational costs on adversarial training.

\smallskip
\noindent
\textbf{\em State Reconstruction.}
\label{sec:state-recon}
Our approach to address this challenge is to bound the attack induced sensor perturbations within a manageable range such as $y \pm \epsilon$, where $\epsilon$ is a small bias. 
This allows us to design a practical and efficient adversarial training approach. 
We use a technique called State Reconstruction~\cite{checkpointing, recovery-lp, delorean} to achieve this. 
State Reconstruction estimates the RAV's physical states under attacks using trustworthy historic state information.
This process involves two components (1) a Checkpointer, and (2) an Estimator. 
The Checkpointer records RAV's historic state information in a sliding window. 
It relies on the attack detector to ensure that the recorded states are uncorrupted.   
The Estimator models the RAV's non-linear physical dynamics using system identification~\cite{system-dentification}. 

We incorporate the state reconstruction technique proposed in our prior work, DeLorean~\cite{delorean} into our adversarial training due to its selective state reconstruction capabilities. 
The Delorean approach strategically reconstructs \sysname's input vector (shown in Equation~\ref{eqn:input}) for sensors compromised by an attack,
using its historic states, while simultaneously preserving the accuracy of states derived from unaffected sensors. 
We empirically compare this state reconstruction technique with alternative approaches for minimizing attack induced sensor perturbations such as sensor fusion~\cite{ekf}, and denoising autoencoders~\cite{unrocker} (details in \S~\ref{sec:implementation}). 
Our results show that DeLorean's state reconstruction~\cite{delorean} significantly outperforms the alternative techniques in bounding the attack induced sensor perturbations. 
Thus we use state reconstruction in designing \sysname.  
The implementation details are in Appendix~\ref{appn:state-recon}.

We present our adversarial training algorithm in Appendix~\ref{appn:multi-agent-training} due to space constraints. 
We frame \sysname's adversarial training as a multi-agent interaction involving two players namely \sysname and an \anomalyagent, a known strategy in Deep-RL~\cite{rarl, adv-agent-training}. 
\sysname's objective is to recover RAVs from attacks, while complying with the mission specifications. 
In contrast, the \anomalyagent's goal is to strategically launch attacks with varying intensity and timing to thwart \sysname.
We simultaneously train \sysname and \anomalyagent framing the training as a zero-sum game. 
This means every time \sysname fails in its objectives, the \anomalyagent gains a reward, and vice versa \ie the total return of both players sums to 0. 
If the change in the average reward for both players becomes smaller over time \ie  
$(\frac{d}{dt} \overline{R}_{\text{SG}}(t))| < \epsilon$, and $(\frac{d}{dt} \overline{R}_{\text{AA}}(t))| < \epsilon$, this indicates both \sysname and \anomalyagent have learned optimal policies. 
More details of \anomalyagent are in Appendix~\ref{appn:multi-agent-training}.

\subsection{\sysname Recovery}
\label{sec:recovery}
We design two variants of \sysname for attack recovery in RAVs: 
(1) Reactive Control: In this variant, \sysname operates as a secondary controller that is activated only upon attack detection. Once activated, it derives control commands to maneuver the RAV while complying with the mission specifications.
(2) Proactive Control: In this variant,  \sysname replaces the RAV's original controller, and it derives control commands to maneuver RAVs, while complying with the mission specifications even in the absence of attacks. 
The reactive control variant is trained to only take recovery
actions post-attack detection, while the proactive control variant
is trained to maneuver the RAV both with and without attacks

\begin{figure}[!ht]
    \centering
    \includegraphics[width=0.49\textwidth]{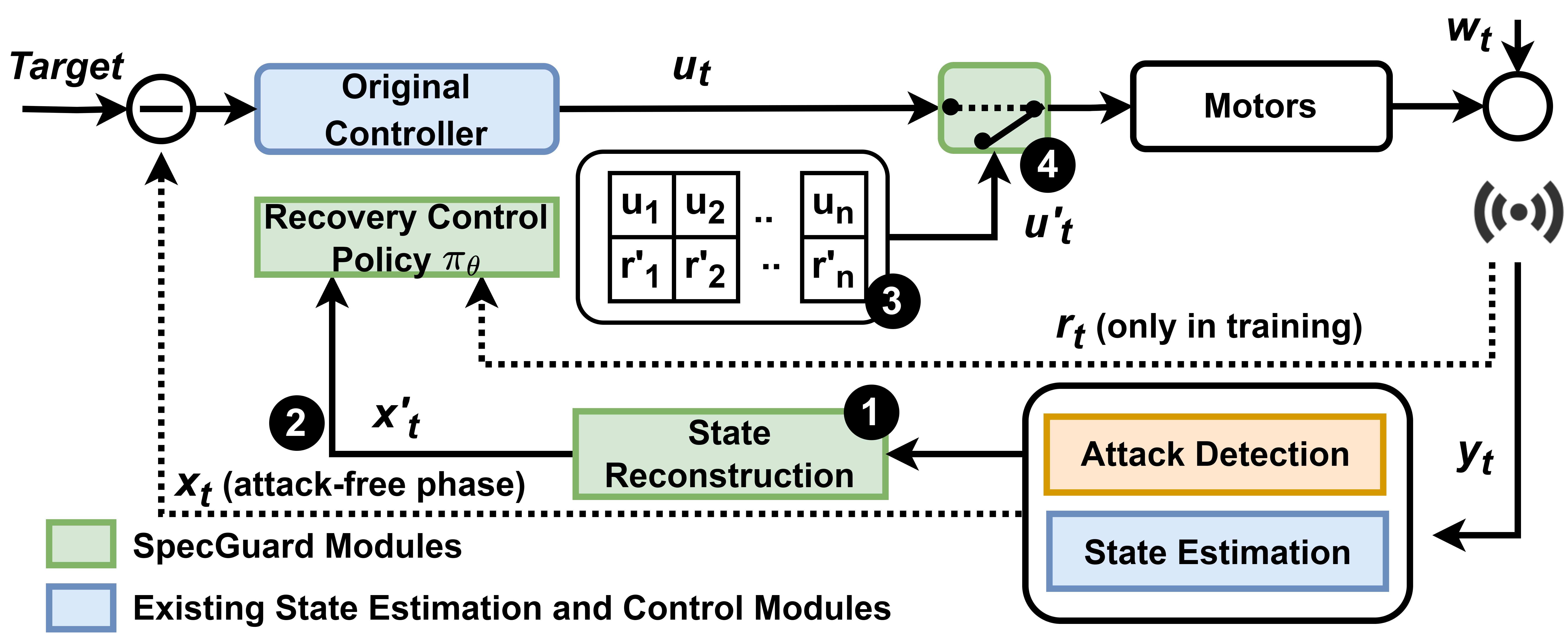}
    \caption{\sysname Architecture and steps in recovery.}
    \label{fig:sysname}
\end{figure}

\smallskip
\noindent
\textbf{\em Reactive Control.}
Figure~\ref{fig:sysname} shows the Reactive Control variants of \sysname, and its recovery process.  
\circlenum{1} The {\em State Reconstruction} is activated after a canonical attack detector raises an alert. 
It isolates the compromised sensors, and uses the trustworthy uncompromised sensors to derive a robust state estimation of the RAV. 

\circlenum{3} Given the reconstructed states, \sysname determines optimal control commands. 
It does so by evaluating potential control commands and the resulting state transitions using a value function $V(x)$. 
The value function estimates potential rewards associated with state transitions considering both immediate and future rewards.  
The state with the maximum $V(x)$ exhibits the strongest compliance with the multiple mission specifications.
Equation~\ref{eqn:value-function} shows the $V(x)$ calculation, 
$R(.)$ is the reward function generated from the mission specifications, $x'_t$ and $x'_{t+1}$ are \sysname's current and next state inputs, $u'_t$ is \sysname's control command, and $\gamma$ is a discount factor that balances both immediate and future rewards.  

\begin{equation}
    V(x) = \max_{u'_t}~\Sigma_{x\in X} [R(x'_{t+1}, u'_t, x'_t)] + \gamma V(x'_{t+1}) 
    \label{eqn:value-function}
\end{equation}

\sysname uses the learned policy $\pi_\theta(u'_t|x'_t)$ to derive control commands $u'_t$ for maneuvering the RAV through a sequence of states that indicates strong compliance with mission specifications.   
Recall that the control command $u'_t$ simultaneously enforces multiple mission specifications 
(Equation~\ref{eqn:output}).
\circlenum{4} Once the attack detector indicates that the attack has subsided, the {\em Recovery Switch} restores the control to the original controller from \sysname.  

\smallskip
\noindent
\textbf{Proactive Control.}
\sysname's proactive control variant derives control commands using the learned policy $\pi_\theta(u'_t, x'_t)$, similar to that in the reactive control variant. 
The proactive control variant simplifies the recovery process by eliminating the need for a recovery switch. 
Since \sysname acts as the main controller, it needs to ensure mission compliance even in the absence of attacks.
 
\section{Experimental Setup}
\label{sec:experimental-setup}
\begin{table*}[]
\centering
\footnotesize
\caption{Mission Specifications for evaluating \sysname. The Params show the parameters to be used in enforcing the mission specifications, and the Control Parameters are parameters in PX4 autopilot that sets them.}
\begin{tabular}{l|l|l|l|l|l}
\hline
\textbf{ID} & \textbf{Mission Specification}  & \textbf{Control Parameters}                                                      & \textbf{Params}                                     & \textbf{Condition}                                                                                 & \textbf{Mission Specification in STL}  \\ \hline
$S_1$     & Avoid collisions                & CP\_DIST                                                                         & 5m                                                  & $obstacle\_distance(x_t)$                                                                          & $G(obstacle\_distance(x_t) > 5)$       \\ 
$S_2$     & Do not veer off  a boundary     & NAV\_MC\_POS\_TH                                                                 & 10m                                                 & $checkCurrentPosition(x_t)$                                                                         & $G(checkCurrentPosition(x_t) < 10)$    \\ 
$S_3$     & Maintain minimum altitude       & NAV\_MIN\_LTR\_ALT                                                               & 10m                                                 & $altitude(x_t)$                                                                                    & $G(altitude(x_t)> 10$                  \\ 
$S_4$     & Maintain maximum altitude       & MIS\_TAKEOFF\_ALT                                                                & 20m                                                 & $altitude(x_t)$                                                                                    & $G(altitude(x_t)<20$                   \\ 
$S_5$     & Navigate through waypoints      & CMD\_NAV\_WAYPOINT                                                          & 5m                                          & $dist\_to\_waypoint(x_t)$                                                                                        & $F_{[t_i, t_j]}(dist\_to\_waypoint(x_t) < 5)$  \\ 
$S_6$     & Maintain distance from obstacle & CP\_DIST                                                                         & 5m                                                  & $obstacle\_distance(x_t)$                                                                          & $G(obstacle\_distance(x_t)>5)$         \\ 
$S_7$     & Maintain minimum velocity       & MPC\_XY\_VEL\_MIN                                                                & 5m/s                                                & $velocity\_xy(x_t.velocity)$                                                                       & $G(velocity\_xy(x_t.velocity)>5)$      \\ 
$S_8$     & Maintain maximum velocity       & MPC\_XY\_VEL\_MAX                                                                & 12m/s                                               & $velocity\_xy(x_t.velocity)$                                                                       & $G(velocity\_xy(x_t.velocity)<12)$     \\ 
$S_9$     & Precision landing               & MPC\_LAND\_SPEED                                                                 & 0.5 m/s                                             & $velocity\_z(x_t.velocity)$                                                                        & $G_{[t_i, t_j]}(velocity\_z(x_t.velocity \le 0.5))$ \\ 
$S_{10}$  & Takeoff speed                   & MPC\_TKO\_SPEED                                                                  & 2m/s                                                & $updateRamp(x_t.z)$                                                                                & $G_{[t_i, t_j]}(updateRamp(x_t.z) \ge 2)$           \\ 
$S_{11}$  & Maintain loiter radius $R$      & NAV\_MC\_ALT\_RADIUS                                                             & 8m                                                  & $allows\_position(x_t, R)$                                                                         & $G(allows\_position(x_t, R) < 8)$      \\ 
$S_{12}$  & Stay within geofence            & \begin{tabular}[c]{@{}l@{}}GF\_MAX\_HOR\_DIST\\ GF\_MAX\_VER\_DIST\end{tabular} & \begin{tabular}[c]{@{}l@{}}50m, \\ 10m\end{tabular} & \begin{tabular}[c]{@{}l@{}}$isCloserThanMaxDist(h,v)$\\ h:horizontal, v:vertical dist\end{tabular} & $G(isCloserThanMaxDist(h, v)>1)$       \\ \hline
\end{tabular}
\label{tab:mission-specs}
\end{table*}

In this section, we present our experimental setup, the implementation details of \sysname\footnote{\sysname's code is available at \url{https://github.com/DependableSystemsLab/specguard}}, and the various types of mission specifications \sysname enforces. 
Finally, we discuss the evaluation setup, and the metrics used to evaluate \sysname. 

\subsection{Subject RAVs}
\label{sec:mission-scenarios}
We evaluate \sysname on two virtual RAVs: 
(1) PX4's quadcopter (PXCopter), and 
(2) ArduPilot's ground rover (ArduRover)~\cite{ardupilot}. 
PXCopter uses PX4 firmware version 1.13.0, and ArduRover uses ArduPilot firmware version 4.3.5.
We also deploy \sysname in two commercial real RAVs: (1) Tarot drone and (2) Aion rover - details are in \S~\ref{sec:real-ravs}.
Table~\ref{tab:rv-sensors} lists the sensors present in the RAVs. 

\begin{table}[!ht]
\footnotesize
\centering
\caption{Sensors in the RAVs used for evaluation. TD: Tarot drone, PC: PXCopter, R1: Aion rover, AR: ArduRover}
\begin{tabular}{l|cccc|l}
\hline
\multirow{2}{*}{\textbf{Sensor Type}} & \multicolumn{4}{l|}{\textbf{Numer of Sensors}}                                                                                            & \multicolumn{1}{c}{\multirow{2}{*}{\textbf{\begin{tabular}[c]{@{}c@{}}Bias Values\\ Manipulating Sensors\end{tabular}}}} \\ \cline{2-5}
                                      & \multicolumn{1}{l|}{\textbf{TD}} & \multicolumn{1}{l|}{\textbf{PC}} & \multicolumn{1}{l|}{\textbf{R1}} & \multicolumn{1}{l|}{\textbf{AR}} & \multicolumn{1}{c}{}                                                                                                           \\ \hline
GPS                                   & \multicolumn{1}{c|}{1}           & \multicolumn{1}{c|}{1}           & \multicolumn{1}{c|}{1}           & 1                                & Position: 1-50m                                                                                                                \\ 
Optical Flow                                & \multicolumn{1}{c|}{1}           & \multicolumn{1}{c|}{2}           & \multicolumn{1}{c|}{-}           & 1                                & Optical flow:  1-7.07 px/frame                                                                                                   \\ 
Gyroscope                             & \multicolumn{1}{c|}{3}           & \multicolumn{1}{c|}{3}           & \multicolumn{1}{c|}{1}           & 3                                & Attitude: 0.5-9.47 rad                                                                                                         \\ 
Accelerometer                         & \multicolumn{1}{c|}{3}           & \multicolumn{1}{c|}{3}           & \multicolumn{1}{c|}{1}           & 3                                & Acceleration: 0.5-6.2$rad/s^2$                                                                                                   \\ 
Magnetometer                          & \multicolumn{1}{c|}{3}           & \multicolumn{1}{c|}{3}           & \multicolumn{1}{c|}{1}           & 3                                & Heading: 90-180 deg                                                                                                               \\ 
Barometer                             & \multicolumn{1}{c|}{1}           & \multicolumn{1}{c|}{2}           & \multicolumn{1}{c|}{1}           & 1                                & Pressure: 0.1 kPa                                                                                                              \\ \hline
\end{tabular}
\label{tab:rv-sensors}
\end{table}

\subsection{\sysname Implementation}
\label{sec:implementation}

\noindent
\textbf{\em Mission Specifications.}
We construct \sysname to enforce 12 mission specifications shown in Table~\ref{tab:mission-specs}. 
These mission specifications are derived from two RAV autopilot software, namely PX4~\cite{px4} and ArduPilot~\cite{ardupilot}.
These are typical mission specifications provided along with the mission plan. 
We use the functions and APIs in RAV autopilot software to express the condition in the mission specifications. 
The functions in Table~\ref{tab:mission-specs} are available in \texttt{pos\_control} and \texttt{attitude\_control} modules in PX4. The corresponding functions for ArduPilot are in Appendix~\ref{appn:mission-specs-ardupilot}.
By {\em mission specification violation}, we mean any of the $S_n$ is violated. 
We consider \textbf{$S_1, S_2, S_{12}$ as critical mission specifications} due to the significant consequences associated with their violation such as collisions and mission failure. 

\smallskip
\noindent
\textbf{\em Supplementary Modules.}
As mentioned before, \sysname relies on existing techniques for attack detection and diagnosis.  
For detection, we use our prior work PID-Piper's attack detection module~\cite{pid-piper}, which is a feed-forward control based detector. 
PID-Piper~\cite{pid-piper} 
is effective against both overt and stealthy attacks. 
For attack diagnosis, we use DeLorean's~\cite{delorean} causal analysis to identify compromised sensors.
We do not measure the effectiveness of the attack detection and diagnosis modules (true positives, and false positives) -  this is reported in our respective prior work~\cite{pid-piper, delorean}.  

\smallskip
\noindent
\textbf{\em State Reconstruction.}
We use the implementation of {\em State Reconstruction} from our prior work~\cite{delorean}. 
We also compare the effectiveness of state reconstruction in limiting attack-induced sensor perturbations with alternative techniques namely sensor fusion~\cite{ekf}, and sensor denoising~\cite{unrocker}. 
We find that {\em state reconstruction incurs 3X lower state estimation error} compared to the alternative techniques. 
The details of this experiment are in Appendix~\ref{appn:res-state-recon}.

\smallskip
\noindent
\textbf{Training.} We implement \sysname using stable-baselines 3~\cite{stable-baselines}. 
We use the Proximal Policy Optimization (PPO)~\cite{ppo} algorithm for training both the reactive and proactive control variants of \sysname.
The reactive control variant is trained to take recovery actions post-attack detection, while the proactive control variant is trained to maneuver the RAV both with and without attacks (\S~\ref{sec:recovery}).  
We use domain randomization techniques~\cite{domain-random} to inject noise (randomness) such as wind, thrust, drag, and friction to improve \sysname's robustness to environmental noise. 
We use the PPO design and hyperparameters from~\cite{drone-racing} - details are in  Appendix~\ref{appn:sysname-training}.

\subsection{\sysname Evaluation}
\label{sec:evaluation}

\noindent
\textbf{\em Physical Attacks for Evaluation.}
We evaluated \sysname under both \textbf{overt} attacks and \textbf{stealthy} attacks (details in \S~\ref{sec:bg-faults-attacks}): 
These attacks targeted six different sensors in the subject RAV's  - GPS, optical flow sensor, gyroscope, accelerometer, magnetometer, and barometer.
We used the code and methodology from prior work to launch attacks~\cite{stealthy-attacks, pid-piper, physical-attacks}. 
Due to the challenges associated with launching physical attacks via real signal injection, all the prior defense techniques~\cite{pid-piper, savior, ci-choi, srr-choi, recovery-lqr, recovery-rl, delorean} use a software-based method to simulate attacks. We followed a similar approach. 
Specifically, we added our attack code into the sensor interface that transmits the sensor measurements to the feedback control loop. 

\smallskip
\noindent
\textbf{\em Sensor Bias Values.}
We inject constant, gradually increasing, and gradually decreasing sensor bias values for each sensor type within the allowable limits for each sensor type. 
We modify the sensor signal $y_t$ as $y^a_t = y_t+b$, where $b$ is either a constant bias or gradually increasing and decreasing bias within the allowable limit.
For example, the maximum hopping distance of most GPS receivers is 50m (update frequency $\times$ maximum velocity)~\cite{gps-datasheet}
Thus, for GPS, we set the range of the bias value to be 1-50m, which is its operating limit. 
We derive the allowable bias values as per the respective sensor specifications and prior work~\cite{injected-delivered, cam-lidar, tractor-beam}. 
Table~\ref{tab:rv-sensors} shows the ranges of the bias values we set for each sensor type.

\begin{figure}[!ht]
\centering
\subfigure{
        \includegraphics[width=0.23\linewidth]{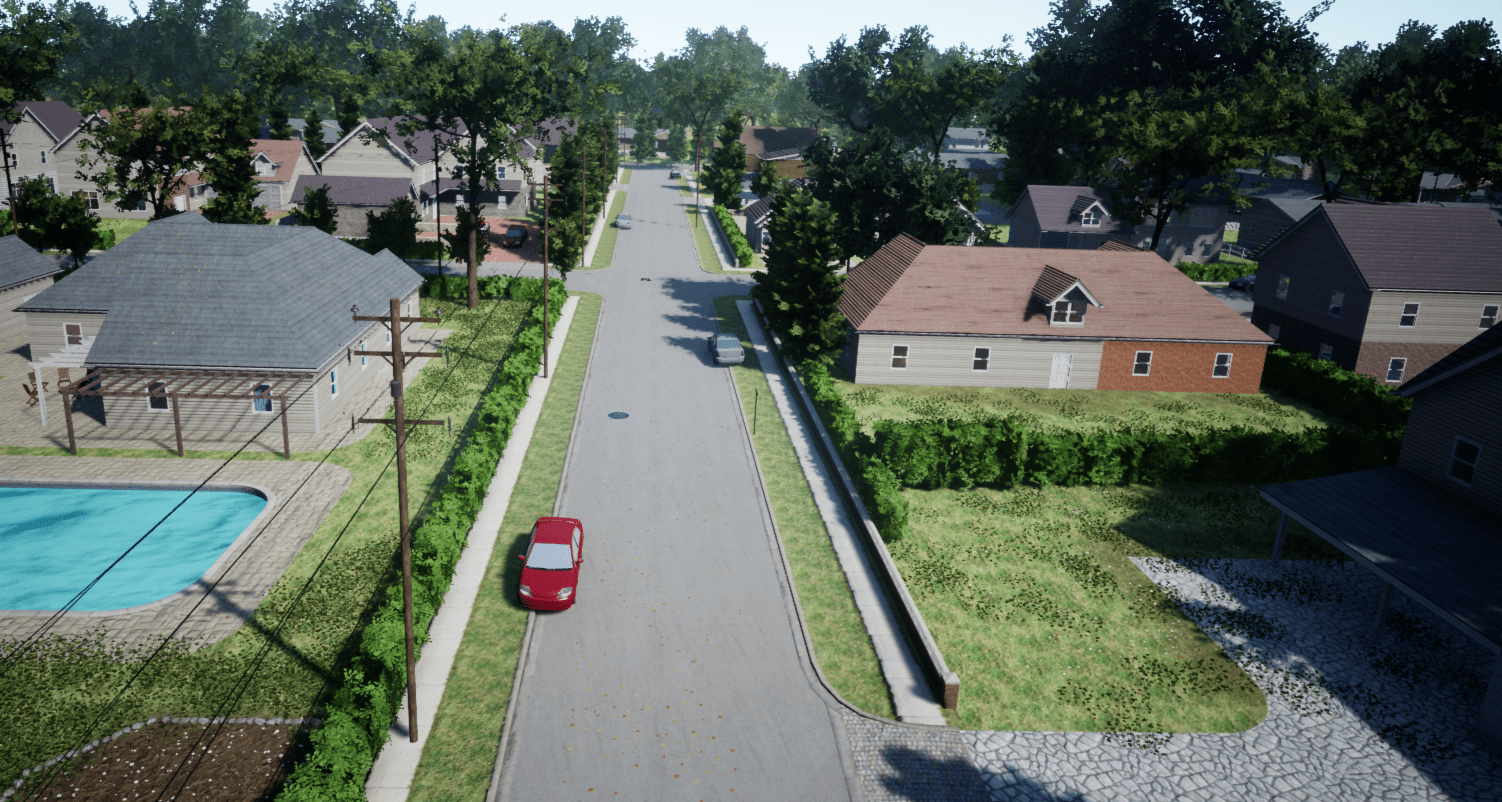} 
    }%
\subfigure{
    \includegraphics[width=0.23\linewidth]{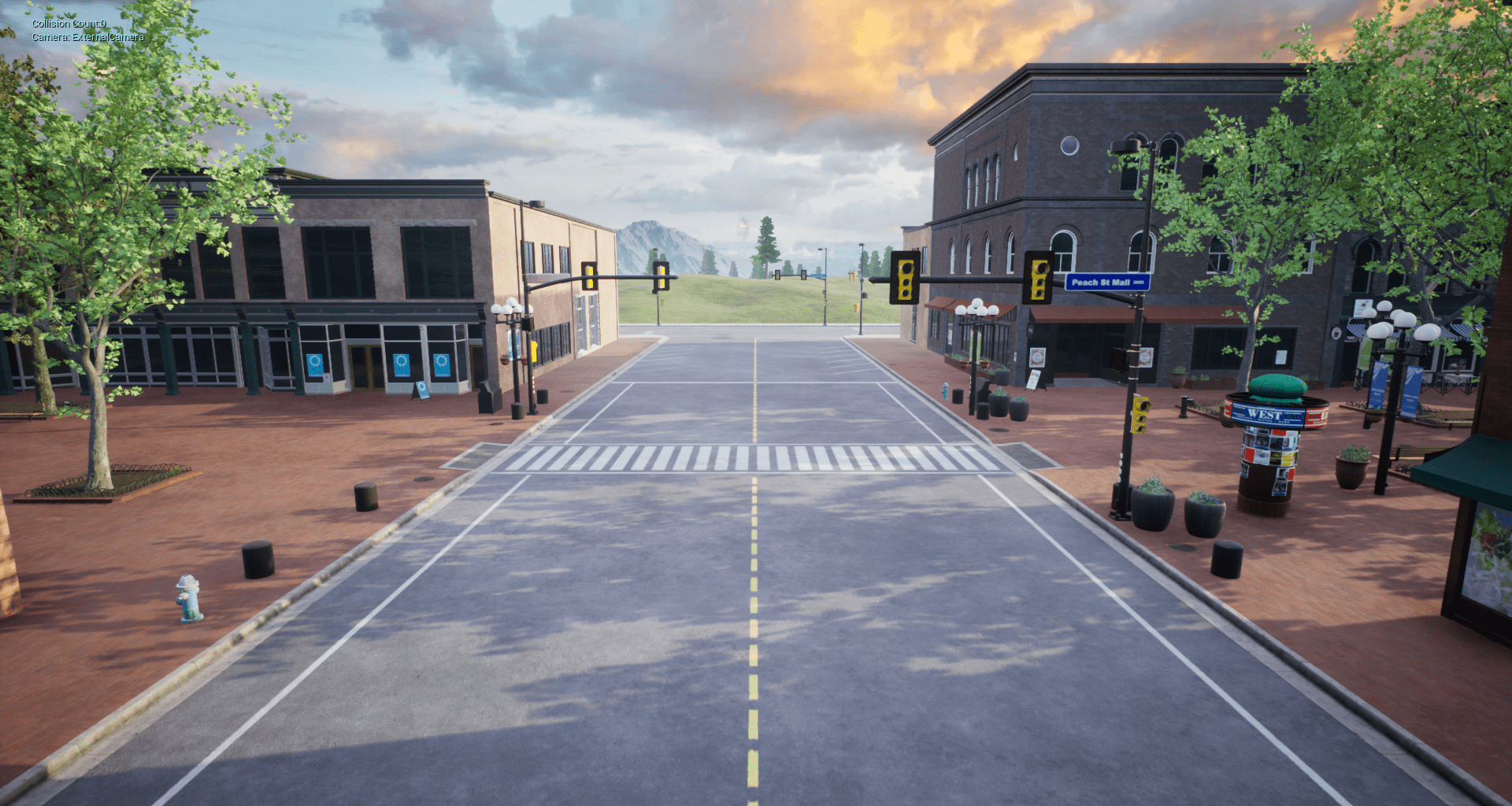}
}%
\subfigure{
        \includegraphics[width=0.23\linewidth]{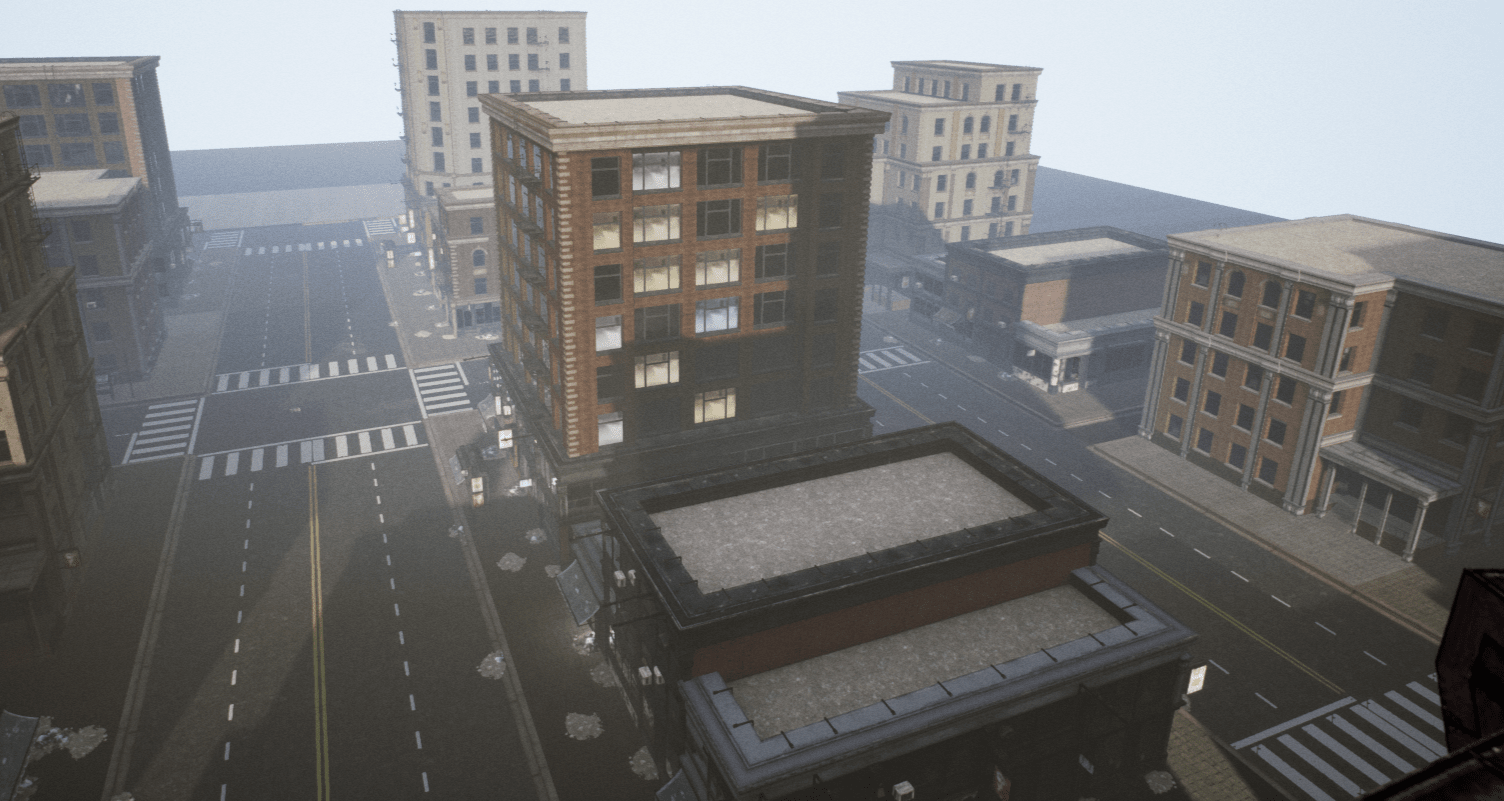} 
    }%
\subfigure{
    \includegraphics[width=0.23\linewidth]{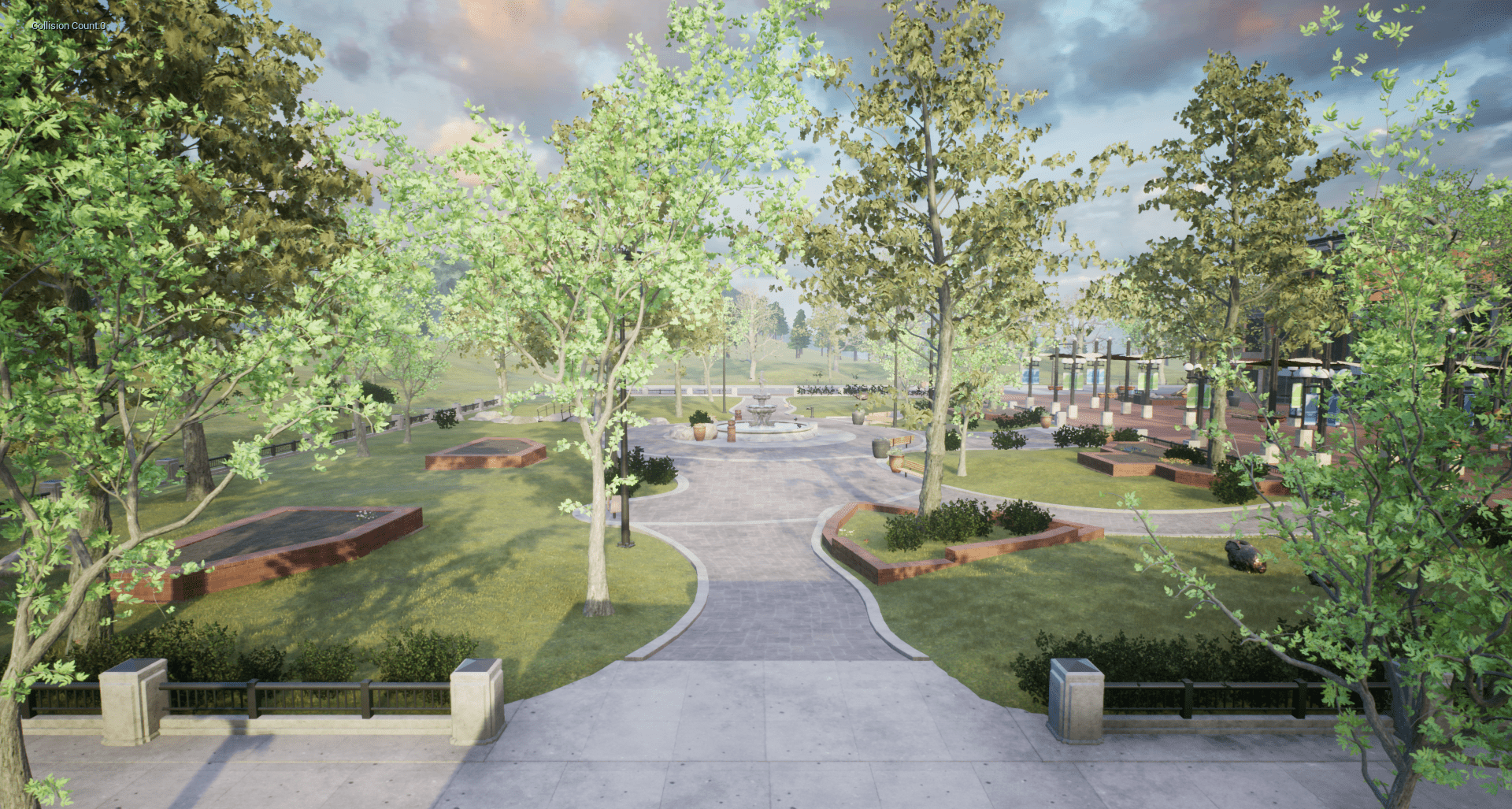}
}
\caption{Operating environments (left to right) - suburban, urban city areas, urban high-rise areas, urban green areas.}
\label{fig:environments}
\end{figure}

\smallskip
\noindent
\textbf{\em Operating Environments.}
We evaluated \sysname in ten different types of environments, ranging from suburban areas, urban areas, urban parks, urban high-rise areas, indoor settings, and harsh weather conditions.  
Figure~\ref{fig:environments} shows examples of a few operating environments, and Appendix~\ref{appn:mission-envs} shows the full list. 
We use Microsoft AirSim~\cite{airsim} for simulating vehicle dynamics and real-world operating environments. 
AirSim is an open-source platform that accurately represents reality (narrowing Sim2Real gap). 
We used the Unreal Engine environments in AirSim~\cite{airsim}, which simulate 3D realistic depictions of the above operating environments. 

To evaluate \sysname, we run more than 18000 RAV missions covering various types of trajectories (\eg complex geometric patterns, challenging paths with multiple obstacles, etc.). 
Each mission lasted for a time duration between 1 and 5 minutes.

\subsection{Evaluation Metrics}
\label{sec:metric}
\noindent
We use three metrics to evaluate \sysname, as follows.

\textbf{1. \em Specification Violation Rate (SVR):}
In our experiments, we introduce attacks intended to cause violations of mission specifications. The SVR measures the percentage of these intended violations that actually occur,  
as shown below. Lower values are better. 

\begin{equation}
    \footnotesize
    \thinspace
    \text{SVR} = \frac{\mathrm{Number~of~observed~specification~violations}}{\mathrm{Total~intended~specification~violations}} \times 100
\end{equation}

\noindent
\textbf{2. \em Recovery Success Rate (RSR):}
Similar to prior work~\cite{pid-piper} we consider a recovery to be successful if the RAV completes its mission and reaches the final waypoint within a $5m$ position error margin. 
This error margin accommodates the standard GPS offset~\cite{gps-offset}. 
A recovery fails if the RAV crashes or stalls.  
The RSR calculation is shown below. Higher values are better. 

\begin{equation}
     \footnotesize
     \thinspace
     \text{RSR} = \frac{\mathrm{Numer~of~sucessful~Recovery}}{\mathrm{Total~missions~under~attacks}} \times 100
    \label{eqn:rsr}
\end{equation}

\noindent
\textbf{3. \em Mission Delay (MD):}
We compare the mission completion time of a mission under attacks $T_{SG}$, and an attack-free ground truth mission $T_{GT}$ on the same trajectory.  
This allows us to calculate potential mission delays (MD) caused by \sysname.  
However, there might be minor variations in the mission completion times even in the same trajectory. We account for these variations using the baseline mission completion time $T_{b}$ 
which is the average of the minimum and maximum mission completion times in attack free missions.  
MD calculation is shown below. Lower values are better. 

\begin{align}
    \footnotesize
    \thinspace
     \text{MD} = \frac{T_{SG}-T_{GT}}{T_{b}} \times 100~~, &&
     T_b = \frac{T_{min} + T_{max}}{2}
    \label{eqn:pmd}
\end{align}
\vspace{-5mm}
\section{Results}
\label{sec:results}
\subsection{Effectiveness in Mission Specification Compliance in Absence of Attacks}
\label{sec:res-mission-specification}

First, we evaluate \sysname's effectiveness in mission specification compliance in the absence of attacks. 
To this end, we compare our compliance-based reward structure (\S~\ref{sec:reward-structure}) in training \sysname with a baseline binary reward structure. 
The binary reward structure uses a binary reward assignment \ie it assigns a reward of 1 if a mission specification is satisfied, and 0 if violated. 
Finally, the overall satisfaction of mission specifications is determined by summing the rewards for individual mission specifications (Table~\ref{tab:mission-specs}).

Our goal is to assess the effectiveness of each \sysname variant in simultaneously complying with multiple mission specifications in attack-free scenarios when trained with these different reward structures.
We train both the Proactive Control (\sysname-PC) and Reactive Control (\sysname-RC) variants of \sysname using both compliance-based and binary reward structures. 
Recall that \sysname-RC is designed as a secondary controller - thus we trigger false detection alarms to activate it purposely. \sysname-PC is the main controller and hence does not require activation.

We run 500 missions in PXCopter for each \sysname variant and reward structure combination.
We find that when trained with the binary reward structure, both the \sysname variants incur $\approx$25\% SVR even in the absence of attacks.   
In contrast, when trained with our compliance-based reward structure both \sysname variants achieve 0\% SVR i.e., there were no mission specification violations. 
{\em Thus, our compliance based reward structure is highly effective in training \sysname to comply with multiple mission specifications.}


\subsection{Proactive vs Reactive Control under Attacks}
\label{sec:res-proactive-reactive}
\begin{figure}[!ht]
\centering
\subfigure[]{
    \includegraphics[width=0.45\linewidth]{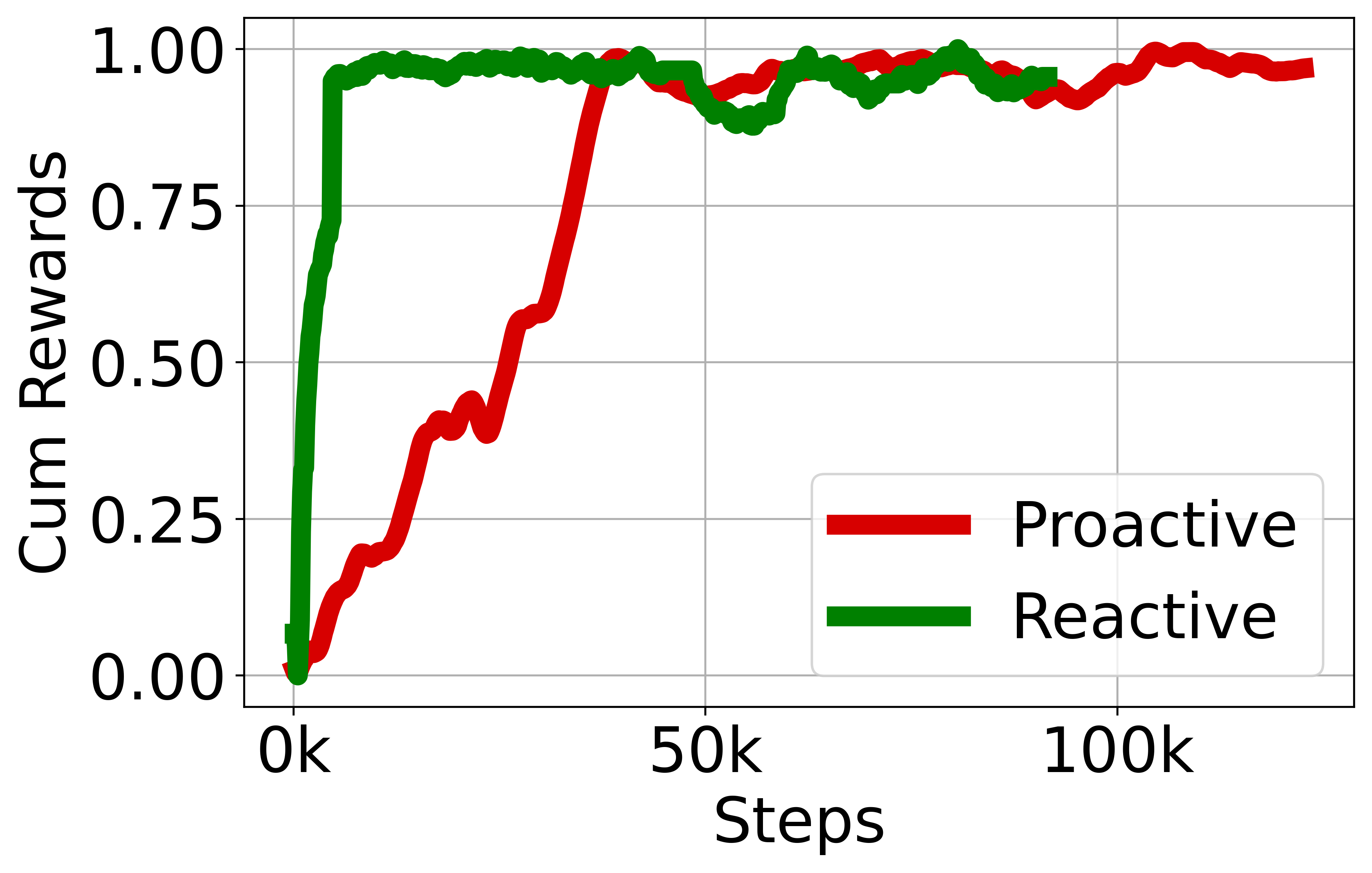}
    \label{fig:proactive-reactive-training}
}
\subfigure[]{
        \includegraphics[width=0.45\linewidth]{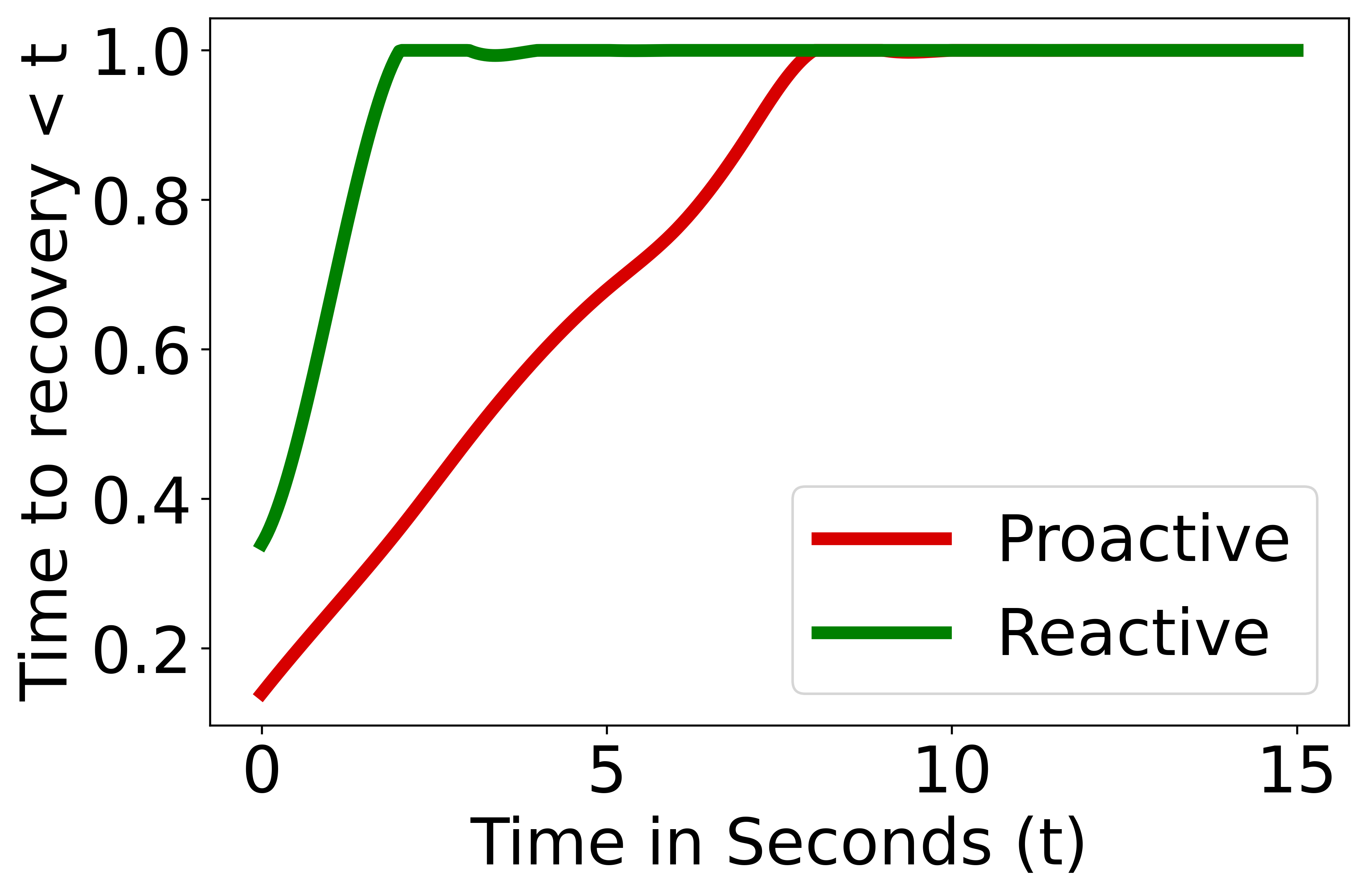} 
        \label{fig:proactive-reactive-md}
    }
\caption{
(a) Training time for optimal policy \sysname-PC vs \sysname-RC.
(b) CDF of time to recovery for \sysname-PC and \sysname-RC under overt and stealthy attacks. }
\label{fig:proactive-reactive}
\end{figure}

First, we assess the training duration for optimal policy learning for both variants of \sysname (\S~\ref{sec:adversarial-training}). 
Figure~\ref{fig:proactive-reactive-training} shows the training time.
As shown in the figure, the cumulative reward per episode for \sysname-PC plateaus after 40k steps ($\sim$20 hours), while the cumulative reward for \sysname-RC plateaus after 10k steps ($\sim$5 hours). 
Recall that cumulative reward is a measure of how well the recovery control policy is complying with multiple mission specifications (\S~\ref{sec:recovery-control-policy}). 
The plateau in cumulative reward indicates an optimal policy is learned. 
We find that \sysname-RC learns an optimal policy $4X$ faster than \sysname-PC. 
This is because \sysname-RC is only active under attacks, and thus does not need to learn the vehicle dynamics in the absence of attacks. 
On the other hand, \sysname-PC as the main controller, is required to learn the vehicle dynamics both in the absence and presence of attacks.

Next, we evaluate both \sysname variants under overt and stealthy attacks to understand their effectiveness in mission specification compliance under attacks. 
First, we observe the impact of the attacks without any protection. 
Then, we record \sysname-PC and \sysname-RC's effectiveness in preventing mission specification violations. 
We run 2400 missions for each \sysname variant and launched 400 instances of attacks against each sensor (Table~\ref{tab:rv-sensors}). 

\smallskip
\noindent
\textbf{\em Impact of Attacks.}
We launch attacks targeting various sensors in virtual RAVs. 
We aborted the mission at the first violation of \emph{any} mission specification, and recorded the violated mission specification. Table~\ref{tab:msc-pc-rc} shows the details. 
\emph{We find that without any protection, all the missions result in mission specification violations.} Thus, all the attacks cause mission failure and endanger the RAV's safety. 

\smallskip
\noindent
\textbf{\em Effectiveness in Recovery.}
Table~\ref{tab:msc-pc-rc} shows the SVR of both variants of \sysname under overt and stealthy attacks. 
We find that \sysname-PC and \sysname-RC significantly reduced the mission specification violations under overt attacks, achieving SVR values of 14.8\%, and 13.5\% respectively.  
Moreover, both 
variants prevented all violations of the critical mission specifications $S_1, S_2,$ and $S_{12}$. 

Under stealthy attacks, however, \sysname-RC achieved an SVR of 14.83, which is slightly better than \sysname-PC of 18.66\%. 
The distribution of SVR for each mission specification in Table~\ref{tab:mission-specs} under both overt and stealthy attacks is presented in Appendix~\ref{appn:mission-spec-compliance}.

\begin{table}[!ht]
\centering
\footnotesize
\caption{Comparison of SVR under overt and stealthy attacks: No protection, \sysname-PC and \sysname-RC.}
\begin{tabular}{c|c|c|c}
\hline
\textbf{Attack}  & \textbf{No Protection} & \textbf{\sysname-PC} & \textbf{\sysname-RC} \\ \hline
Overt Attacks    & 100                    & 14.8                 & 13.5                \\ 
Stealthy Attacks & 100                    & 18.66                & 14.83                \\ \hline
\end{tabular}
\label{tab:msc-pc-rc}
\end{table}

\smallskip
\noindent
\textbf{\em Time to Recovery.}
We define time to recovery (T2R) as the time between attack detection and recovery actions ensuring 0 SVR. 
Figure~\ref{fig:proactive-reactive-md} shows the Cumulative Distribution Function (CDF) of T2R for both \sysname-RC, and \sysname-PC. 
We find that \sysname-RC is 3X faster than \sysname-PC in recovering RAVs. 

\smallskip
\noindent
\textbf{\em Summary.}
Although both \sysname variants achieve comparable effectiveness in mission specification compliance under attacks, \sysname-RC demonstrates significantly faster policy learning, and it also performs faster recovery. 
Faster recovery is crucial for safeguarding RAVs under attacks and minimizing mission delays. 
Consequently, we consider only  \sysname-RC in the rest of this paper. 
\textbf{Henceforth, \sysname refers to \sysname-RC.}

\subsection{Comparison with Prior Techniques}
\label{sec:res-comparison-recovery}

We compare \sysname with five prior recovery techniques namely: SSR~\cite{srr-choi}, FTC~\cite{recovery-rl}, PID-Piper~\cite{pid-piper}, UnRocker~\cite{unrocker}, and DeLorean~\cite{delorean}. 
We measure the effectiveness of all the above methods including \sysname in mission specification compliance and attack recovery. 

\begin{figure}[!ht]
    \centering
    \includegraphics[width=0.9\linewidth]{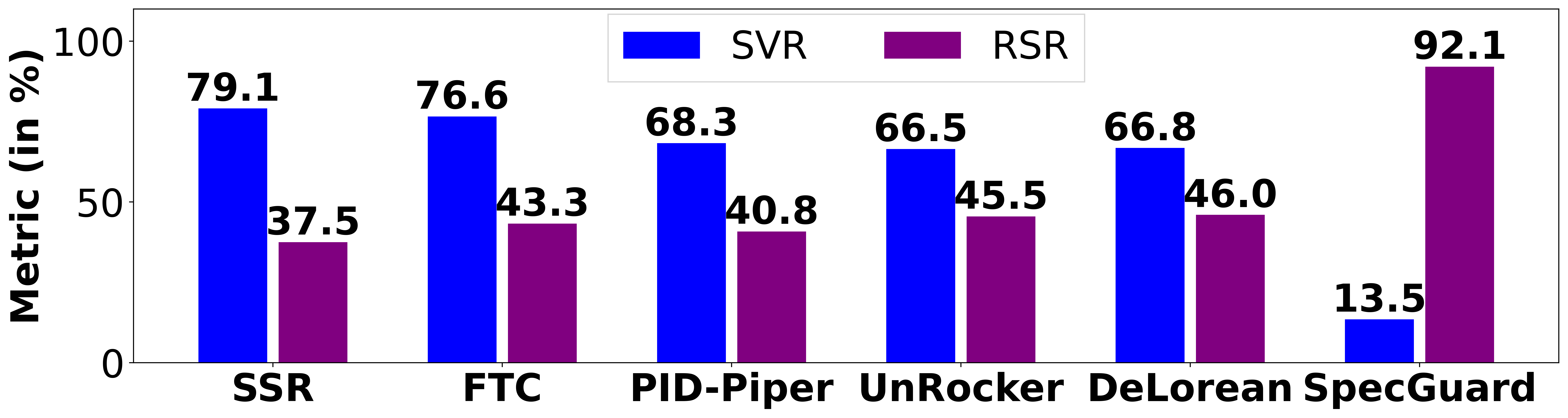}
    \caption{Comparison of \sysname with five prior recovery techniques. \sysname achieves 5X lower specification violation (SVR), and 2X higher recovery success (RSR).}
    \label{fig:recovery-comparison}
\end{figure}

\smallskip
\noindent
\textbf{\em Comparison under Overt Attacks.}
First, we compare the effectiveness of \sysname with prior recovery techniques under overt attacks. 
We run 1200 missions in virtual RAVs for each technique.  
Figure~\ref{fig:recovery-comparison} shows the SVR and RSR for all the above techniques.  
We find that the prior recovery techniques (SSR, FTC, PID-Piper, UnRocker, DeLorean) achieve high SVR between 66.8\% and 79.1\%.  Consequently, the prior recovery techniques achieve low RSR between 37.5\% and 46\%. 
Thus, a high SVR often results in low RSR.
Among the prior recovery techniques, UnRocker~\cite{unrocker} and DeLorean~\cite{delorean} performed the best -  both techniques achieved similar SVR of 66\%, and similar RSR of $\approx$ 46\%.  
In contrast, \sysname achieves an SVR of 13.5\% which is $\approx 5X$ lower compared to UnRocker and Delorean. 
Similarly, \sysname achieves an RSR of 92.1\%, which is a $2X$ higher recovery success compared to both UnRocker and DeLorean. 

Furthermore, we observed that for SSR, FTC, and PID-Piper, more than $90\%$ of the failed recoveries resulted in collisions. 
In the case of DeLorean and UnRocker,  $81\%$ of failed recoveries resulted in collisions.
In contrast, \sysname incurred 0 collisions even when the recovery failed (8\% of the cases). 
This is because prior techniques use a simpler recovery strategy (Table~\ref{tab:prior-work}), leading to unsafe recovery in a broad range of operating environments.  
In contrast, \sysname does not violate the critical mission specifications ($S_1, S_2$, and $S_{12}$). 
Thus, it maneuvers the RAVs within the given operation boundary and maintains a safe distance from obstacles even where recovery failed.  
{\em Thus, compliance with mission specifications leads to \sysname's higher effectiveness in ensuring safety during recovery.}

\noindent
\textbf{\em Comparison under Stealthy Attacks.}
Next, we compare \sysname with the prior recovery techniques under stealthy attacks. 
We run 1200 missions for each technique, and launched stealthy attacks targeting various sensors in the RAVs in each mission. 

We find that the two best prior techniques for overt attacks, FTC and UnRocker achieve SVR values of 90\% and 86\% respectively, and both the techniques achieved RSR of 17\%. 
These values indicate a significant decline in effectiveness compared to that observed under overt attacks. 
This is because these techniques are designed to ignore transient sensor perturbations or fluctuations that may happen due to noise, and thus, they fail to distinguish between noise and stealthy attacks. 
The other three techniques,  
 SSR, PID-Piper, and DeLorean achieve an SVR of 81\%, 70\%, 69.5\%, and RSR of 35\%, 39.75\%, and 43\% respectively. 
These values are comparable to those observed under overt attacks for these techniques.  

In comparison, \sysname achieves an SVR of 14.83\% and an RSR value of 90.41\%. 
Thus, \sysname achieves 4.5X higher mission specification compliance and 2X higher recovery success over DeLorean,  which is the best technique among all the prior techniques. 
Moreover, the SVR and RSR values of \sysname under stealthy attacks are comparable to its SVR and RSR values under overt attacks.

\subsection{Attacks targeting different RAV sensors}
\label{sec:res-different-sensors}
\begin{figure}[!ht]
\centering
\subfigure[]{
        \includegraphics[width=0.48\linewidth]{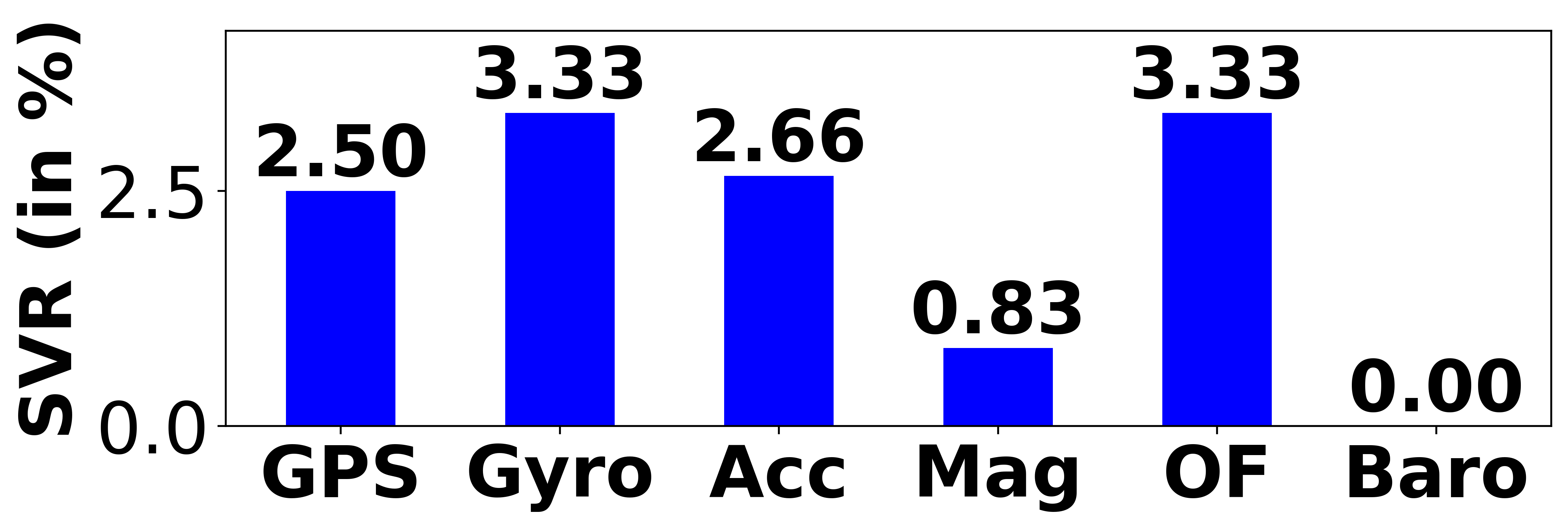} 
        \label{fig:sysname-svr}
    }%
\subfigure[]{
    \includegraphics[width=0.48\linewidth]{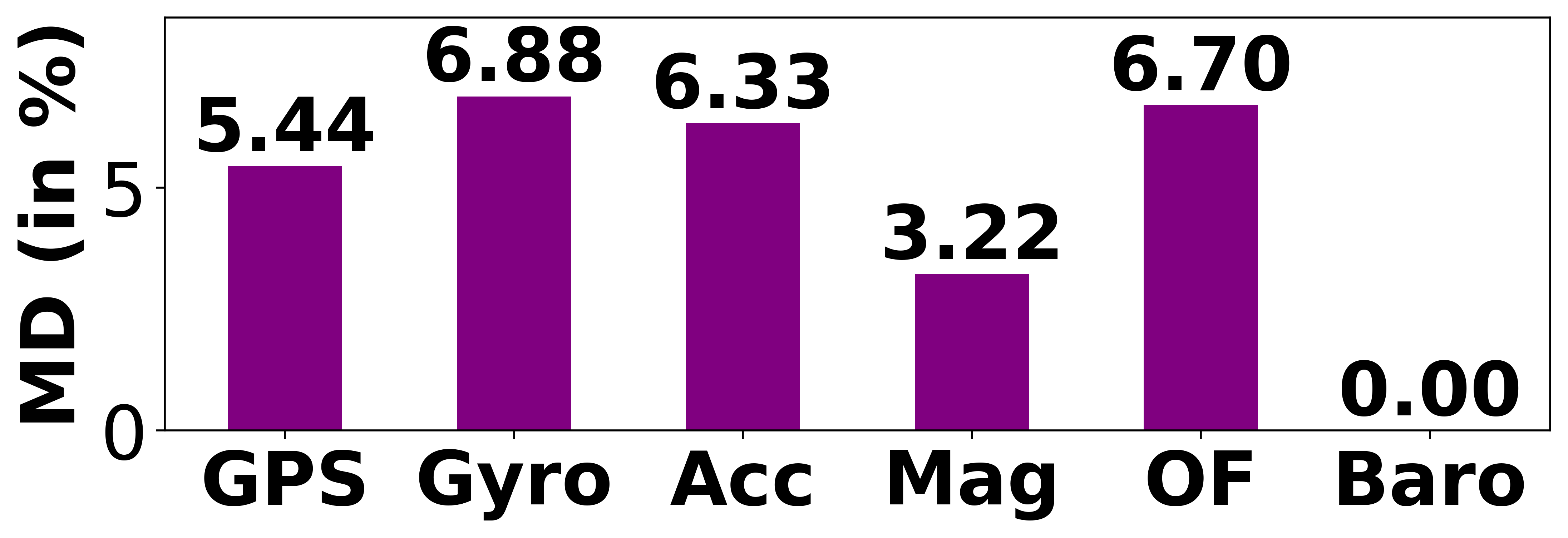}
    \label{fig:sysname-md}
}
\caption{ (a) SVR of \sysname under attacks targeting various sensors in RAVs, (b) MD due to \sysname recovery.}
\label{fig:sysname-svr-md}
\end{figure}

Figure~\ref{fig:sysname-svr} shows \sysname's SVR for attacks targeting the different types of sensors in RAVs namely GPS, Gyroscope, Accelerometer, Magnetometer, Optical Flow (OF), and Barometer. Table~\ref{tab:rv-sensors} shows the attack details used for evaluation. 

We find that \sysname is effective in preventing mission specification violations regardless of which sensor is targeted by the attack, achieving low SVR ($\le3.33\%$).
Notably, attacks targeting the barometer, incurred 0\% SVR, because the state reconstruction leverages GPS and optical Flow (OF) sensors to determine accurate z-axis position even without the barometer. 
Finally, \sysname incurs a modest MD of 3.2\%-6.88\% (Figure~\ref{fig:sysname-md}), excluding attacks targeting Barometer, that incurred no delays  
as they had no impact on the RAV \ie they incurred 0\% SVR as shown in Figure~\ref{fig:sysname-svr}.  

\subsection{\sysname in Real RAV Systems}
\label{sec:real-ravs}
We evaluate \sysname in two real RAVs: 
(1) Tarot 650 drone~\cite{tarrot} (Tarot drone),
(2) Aion R1 ground rover~\cite{aion} (Aion rover),
Both the real RAVs are based on the 
Pixhawk microcontroller~\cite{pixhawk}. 

We modified the RAV's firmware to integrate \sysname. In particular, 
we added \sysname as a new sub-system in the firmware, and flashed the modified firmware on the RAV's microcontroller.   
Our Deep-RL recovery control policy (RCP) is developed using the stable-baselines 3~\cite{stable-baselines} Python library. 
We export the trained RCP to ONNX (Open Neural Network Exchange) format~\cite{onnx} to execute it on the Pixhawk microcontroller of the RAV. 
ONNX allows models to be ported across frameworks and hardware platforms. 
We use the ONNX runtime C++ APIs~\cite{onnx-runtime} for running inference on the trained RCP on the Pixhawk microcontroller.


As running experiments on real RAVs requires elaborate safety precautions in public places, we had to limit the number of experiments. 
We run $20$ missions on each real RAV. 
Each RAV was subject to overt attacks in 10 missions, and stealthy attacks in the remaining 10 missions. 
These missions each 
took 1 to 3 minutes. 

Table~\ref{tab:real-ravs} shows the results of the experiments. 
\sysname achieved an SVR between 5 and 10\%, and RSR $>$ 90\% in both the real RAVs, and incurred 0 collisions/crashes. 
We observed less than 6.6\% MD in both the real RAVs
These results are consistent with the SVR, RSR, and MD observed in the virtual RAVs (\S~\ref{sec:res-comparison-recovery} and \S~\ref{sec:res-different-sensors}). 

\begin{table}[!ht]
\centering
\footnotesize
\caption{\sysname's recovery in Real RAVs.}
\begin{tabular}{c|c|c|c|c|c}
\hline
\textbf{Real-RAVs} & \textbf{SVR} & \textbf{RSR} & \textbf{MD} & \textbf{\begin{tabular}[c]{@{}c@{}}CPU\\ Overhead\end{tabular}} & \textbf{\begin{tabular}[c]{@{}c@{}}Space\\ Overhead\end{tabular}} \\ \hline
Tarot drone        & 10           & 90           & 6.66        & 15.33\%                                                            & 1.8\%                                                             \\ 
Aion Rover         & 5            & 100          & 5.88        & 13.66\%                                                         & 1.5\%                                                             \\ \hline
\end{tabular}
\label{tab:real-ravs}
\end{table}

\noindent
\textbf{\em CPU Overhead}: We measure the CPU overhead incurred by \sysname. 
The RAV autopilot software (PX4, ArduPilot) has a scheduler that tracks the total CPU time incurred by each task and module. 
We calculate the CPU overhead of \sysname by analyzing the additional CPU time recorded by the scheduler when the RAV is equipped with \sysname.
We find that \sysname incurred a CPU overhead of $15.33\%$  in the Tarot drone, and $13.66\%$ in the Aion Rover. 
The CPU overhead is 2X higher compared to DeLorean~\cite{delorean}, the best recovery technique among prior work (Figure~\ref{fig:recovery-comparison}). 
However, \sysname achieves a 2X higher successful recovery compared to Delorean. 
Finally, the addition of \sysname to RAV's autopilot resulted in a  $\approx$1.8\% increase in the firmware size (space overhead). 

We discuss two real RAV experiments in detail: (1) recovery under an overt attack on Aion rover, and (2) recovery under a stealthy attack on Tarot drone, both with and without \sysname. 

\begin{figure}[!ht]
\centering
\subfigure[]{
        \includegraphics[width=0.49\linewidth]{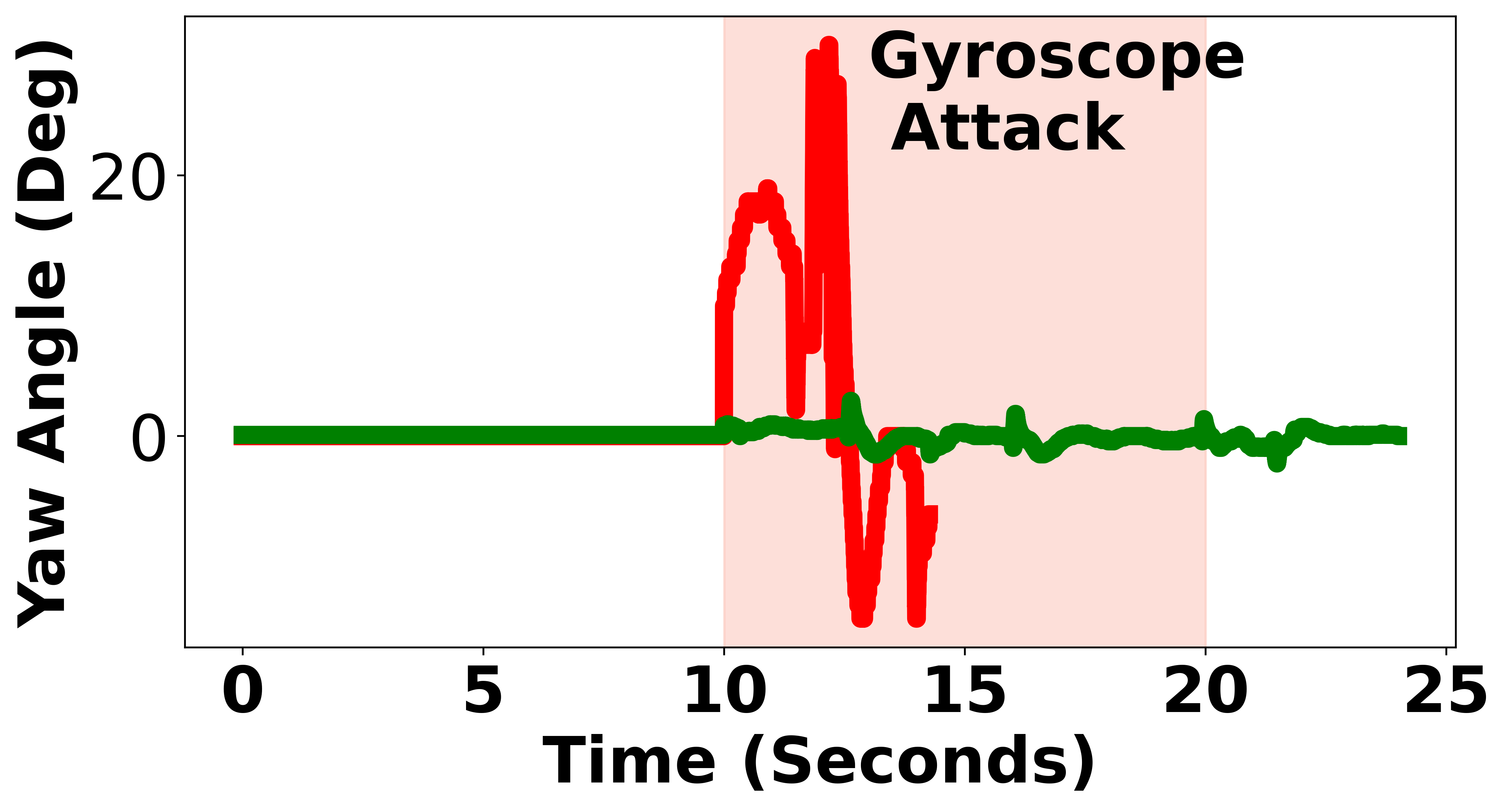} 
        \label{fig:rover-attack}
    }%
\subfigure[]{
    \includegraphics[width=0.49\linewidth]{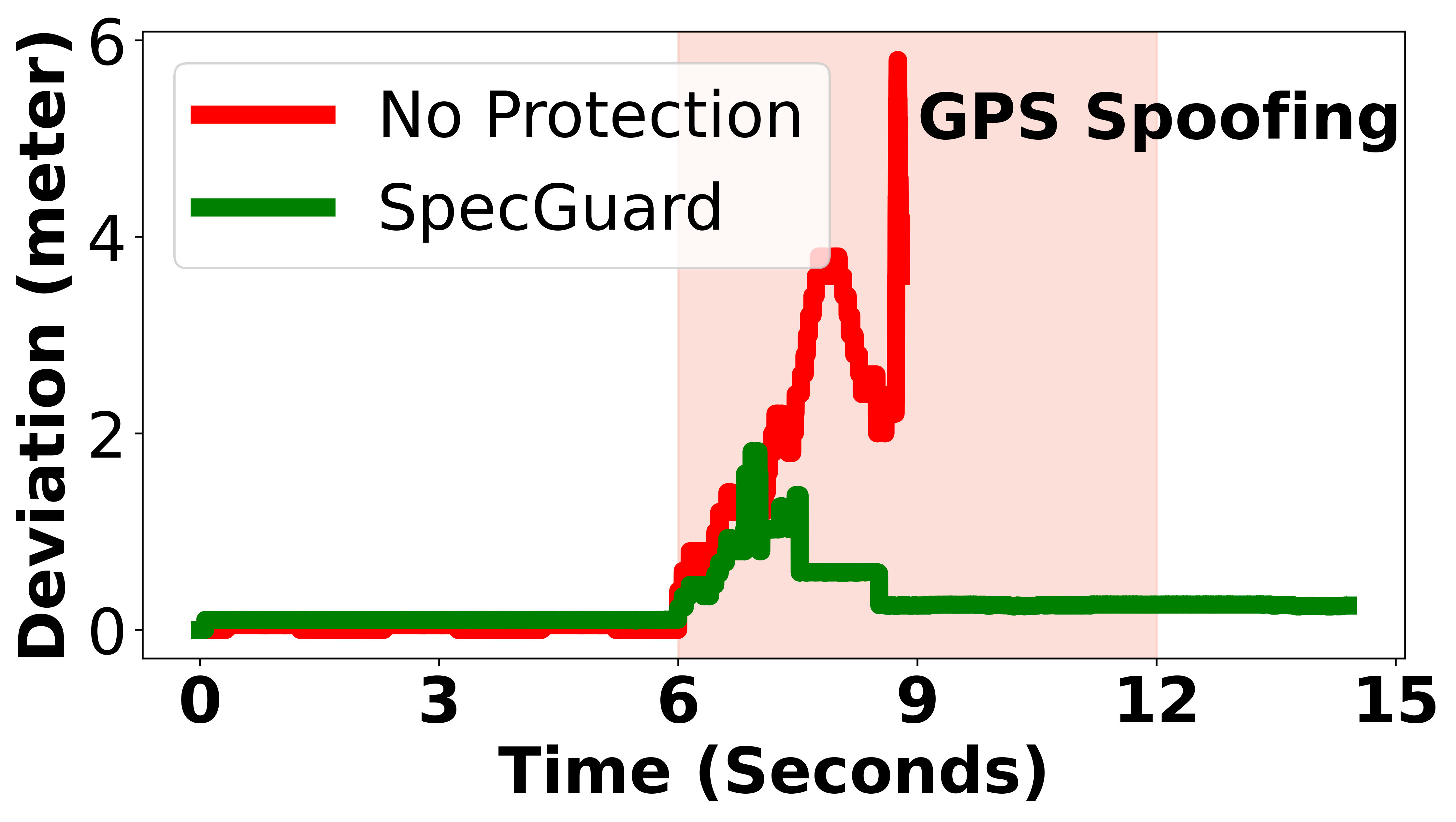}
    \label{fig:drone-attack}
}
\caption{Real RAVs under attacks, comparing scenarios with and without \sysname. 
(a) Aion rover under overt attack.
(b) Tarot drone under stealthy attack.}
\label{fig:real-ravs}
\end{figure}

\smallskip
\noindent
\textbf{\em Aion Rover.}
In this experiment, the rover was navigating a straight line path with an obstacle all along its right side.
We launched an overt attack targeting the gyroscope sensors. 
The red line in Figure~\ref{fig:rover-attack} shows the rover's yaw angle without \sysname. 
As there are no turns and the rover is going straight, the yaw angle (Z-axis movements) should be 0 (desired state), as observed in the attack-free phase (t=0-10s). 
The attack starts at t=10s; the gyroscope perturbations result in erroneous yaw angles. 
A sharp fluctuation in the yaw angle occurs at t=10s, and the rover starts moving right. 
The yaw angle fluctuates between -10 and 30 degrees as the attack continues. 
Eventually, at t=15s, the rover collides with the obstacle. 

The green line in Figure~\ref{fig:rover-attack} shows the rover's yaw angle with \sysname in place. 
\sysname complies with the mission specifications and maintains the yaw angle close to 0 (desired state) through the duration of the attack (10-20s), thus preventing the rover from veering off its path and colliding with the obstacle on its right.  

\noindent
\textbf{\em Tarot Drone.}
In this experiment, the drone takes off, attains an altitude of 10m, and navigates through four waypoints in a rectangular path. 
The red line in Figure~\ref{fig:drone-attack} shows the trajectory error during the drone mission without \sysname. 
We launch a stealthy GPS spoofing attack at t=8s when the drone is on course towards waypoint 1. 
As shown in Figure~\ref{fig:drone-attack}, the drone gradually deviated from the set path at a time between 8-12s. 
At t=12s, the drone abruptly deviates downwards to the left, causing it to 
violate $S_2, S_3$ and $S_5$. 

When \sysname is deployed, it minimizes the deviations under stealthy attacks (between t=8-18s) as shown in the blue line in Figure~\ref{fig:drone-attack}.
\sysname also maintains the altitude of the drone near 10m, and prevents violations of {\em any} of the mission specifications. 

We observed negligible MD in the above experiments. 
{\em These experiments show that the \sysname is also effective in real RAVs.}
We have made the videos of our experiments publicly available~\cite{sc-videos}.

\subsection{Ablation Studies}
\label{sec:res-ablation-study}

We perform ablation studies to understand the importance of two key components in \sysname: (1) recovery control policy, and (2) adversarial training with state reconstruction.
We perform two experiments where we systematically remove one component in each experiment and compare the resulting recovery technique (naive recovery) with \sysname that contains both components. 

\begin{figure}[!ht]
\centering
\subfigure[]{
        \includegraphics[width=0.49\linewidth]{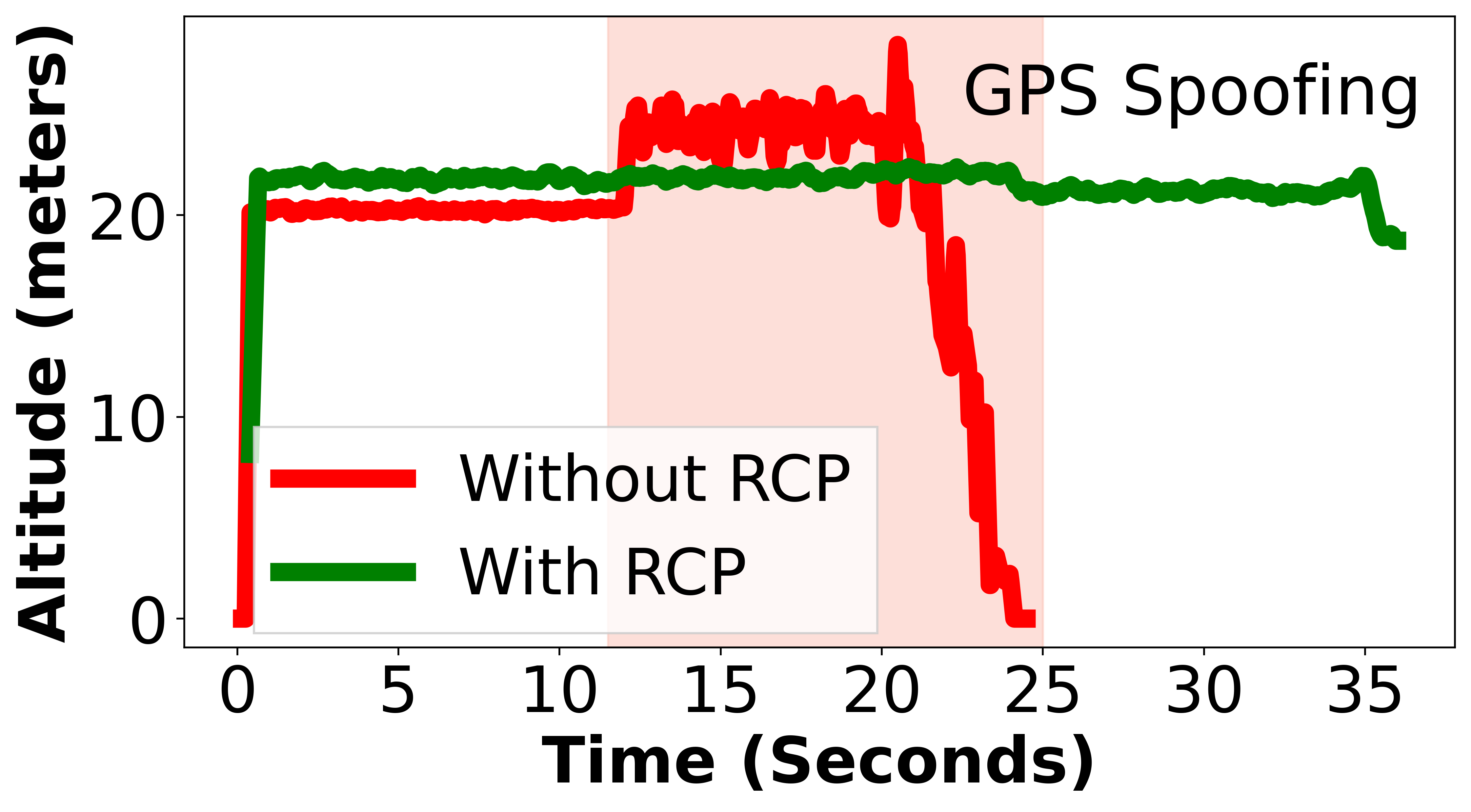} 
        \label{fig:without-rp}
    }%
\subfigure[]{
    \includegraphics[width=0.49\linewidth]{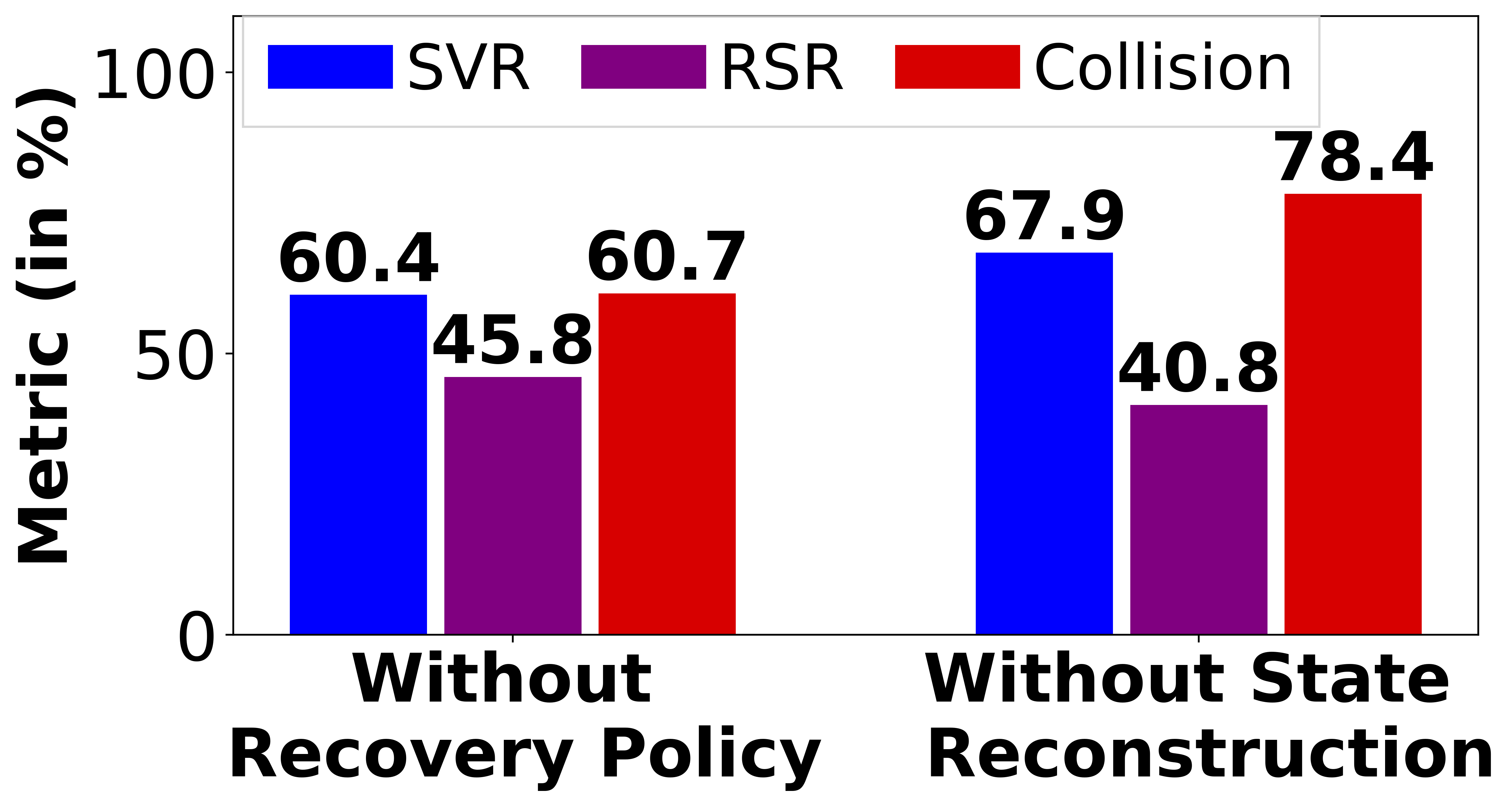}
    \label{fig:ablation-svr-rsr}
}
\caption{Ablation study to evaluate \sysname's design components. 
(a) PXCopter under GPS spoofing attack - the attack causes a crash without the Recovery Control Policy (RCP).
(b) SVR, RSR, and Collision Rate when one of the design components of \sysname is systematically removed.}
\label{fig:ablation}
\end{figure}

\smallskip
\noindent
\textbf{\em Without Recovery Control Policy (RCP).} In this experiment we compare \sysname with a naive recovery technique that uses state reconstruction to limit attack induced sensor perturbations but uses the original controller for recovery~\cite{pid}. 
We run a total of 1200 missions in virtual RAVs to evaluate the naive recovery technique. 
Figure~\ref{fig:ablation-svr-rsr} shows the results. 
The naive recovery technique resulted in an SVR of 60.4\% which is $4X$ higher compared to \sysname (Figure~\ref{fig:recovery-comparison}). 
The high SVR resulted in the naive recovery achieving an RSR of 45.8\% which is $2X$ lower than \sysname. 

Figure~\ref{fig:without-rp} shows a comparison of \sysname and the naive recovery in a mission. 
A GPS spoofing attack was launched during a mission in PXCopter (from t=10-25s).  
The naive recovery (red line in the figure) failed to maintain the drone's set altitude, and eventually, the drone crashed. 
This is because the naive recovery relies on the original controller which is known to overcompensate under attacks~\cite{pid-piper}. 
In contrast, \sysname maintained the drone's altitude at 20m (green line in the figure).
Thus, \sysname comfortably maneuvered the drone resulting in a successful recovery. 

\noindent
\textbf{\em Without State Reconstruction (SR).}
In this experiment, we compare \sysname with a naive recovery technique that uses a recovery control policy upon attack detection. 
However, the recovery control policy is adversarially trained \emph{without state reconstruction}. 

We again run 1200 missions to evaluate this naive recovery technique.  Figure~\ref{fig:ablation-svr-rsr} shows the results. 
We find the naive recovery technique results in an SVR of 67.9\%, which is $5X$ higher compared to \sysname. 
It also achieves $2X$ lower RSR compared to \sysname (Figure~\ref{fig:recovery-comparison}). 
This drop in effectiveness in both mission specification compliance and recovery success stems from the exclusion of state reconstruction. 
Without state reconstruction, the recovery control policy does not generalize to the diverse attacks launched.  

{\em Thus, both the recovery control policy and state reconstruction are important for \sysname's resilience under attacks.}

\section{Discussions}
\label{sec:discussions}
\subsection{Adaptability in New Operating Conditions}
We have evaluated \sysname in ten different operating environments. 
However, in real-world mission scenarios, \sysname may encounter environments it has not seen in training. 
To measure \sysname's adaptability to new operating environments, we design two experiments:  
(A) Suburban to Urban transition: \sysname is trained in suburban environments, which are less dense, and deployed in high-rise urban areas, which are more complex and dense. 
(B) Urban to Suburban transition: Conversely, \sysname is trained in high-rise urban environments and deployed in a suburban environment. 
Both A and B represent different environments from those in which the \sysname was trained. 

We find that \sysname achieved a 14\% SVR in experiment A, and a 12\% SVR in experiment B, under attacks. 
The detailed results are in Appendix~\ref{appn:res-adaptability}.
Thus, \sysname is largely consistent in preventing mission specification violations in both experiments. 
This consistency is due to our reward structure that enables learning optimal policy and deriving appropriate control commands. 
{\em Thus, \sysname adapts reasonably well to new operating environments.}

\subsection{Development Effort and Practicality}

Recall that there are five steps in developing \sysname (\S~\ref{sec:ms-stl} to \S~\ref{sec:adversarial-training}): 
(1) Defining mission specifications as STL, 
(2) Monitoring conditions in STL, 
(3) Constructing reward function using our compliance based reward structure, 
(4) Training for specification compliance, and finally, 
(5) Adversarial training for robustness to attacks.

Manual effort is only needed in the first two steps. In step 1, to 
formally define mission specifications, originally specified in natural language, using STL. In step 2, to 
identify the API calls in RAV autopilot software to monitor conditions in STL specifications. 
We explain the manual effort in detail in Appendix~\ref{appn:effort}.
Note that this manual effort is a one-time cost for a new RAV or new specification.

No manual effort is needed in the other three steps, however. 
For step 3, we provide the reward structure (\S~\ref{sec:reward-structure}), and our Algorithm (Appendix~\ref{appn:reward-function}) automates the construction of the reward function.
Steps 4 and 5 are automated by our implementation (\S~\ref{sec:implementation}).


To deploy \sysname in RAVs, manufacturers can select a standard set of mission specifications for either a drone or a rover in an existing RAV autopilot (PX4, ArduPilot) without writing new specifications. 
Furthermore, our STL based approach can be easily extended to new mission specifications if needed (\S~\ref{sec:design}).

\subsection{Limitations}

\textbf{\em Effectiveness in Recovery.} 
\sysname successfully recovered RAVs from attacks in 92\% of the cases (\S~\ref{sec:res-comparison-recovery}). 
In the remaining 8\% of the cases, the recovery failed because \sysname could not maneuver the RAVs within the allowable position error margin of 5m from the target, This 5m position error margin accommodates GPS offset~\cite{gps-offset}. 
However, even in these cases, \sysname prevented collisions, crashes, or stalls due to the attacks, thereby ensuring safety, unlike prior attack recovery techniques~\cite{srr-choi, pid-piper, recovery-rl, delorean, unrocker}.

\sysname's effectiveness can be improved by optimizing its recovery control policy using imitation learning~\cite{imitation-policy}.
Imitation learning is the supervised learning equivalent of Deep-RL~\cite{imitation-learning}. 
Optimizing \sysname through imitation learning involves collecting data that demonstrates the optimal path in challenging scenarios. 
We explain two such challenging scenarios in Appendix~\ref{appn:improving-effectiveness}.

\smallskip
\noindent
\textbf{\em Multi-sensor Attacks.} 
We only consider single sensor attacks in our threat model. 
Nonetheless, we conducted an experiment to evaluate \sysname under multi-sensor attacks. 
We run 600 missions on PXCopter and launched attacks targeting both GPS and gyroscope sensors simultaneously. 
We found the \sysname was unable to prevent mission specification violations in 24\% of the cases - this is an almost $2X$ higher violation rate than single sensor attacks (Figure~\ref{fig:recovery-comparison}). 
However, multi-sensor attacks are challenging to mount in practice as they require precise and simultaneous signal interference across multiple sensors targeting a moving RAV~\cite{poltergeist}.

\smallskip
\noindent
\textbf{\em Scalability.} 
When introducing new mission specifications, the corresponding STL specifications need to be defined. 
Our algorithm automates the reward structure generation and reward shaping for updating the \sysname.
However, the \sysname policy needs to be fine-tuned for compliance with the new mission specifications~\cite{finetuning}.

\section{Related Work}
\label{related-work}
\noindent
\textbf{\em Attack Detection and Diagnosis}
techniques have been proposed to detect physical attacks~\cite{ci-choi, savior, deep-sim, avmon}, and identify the compromised sensor through diagnosis~\cite{ftc-diagnosis, delorean}. 
However, these techniques do not perform recovery, which is our focus. 

\smallskip
\noindent
\textbf{\em Resilient Control} techniques have been proposed to handle sensor faults and environmental noise~\cite{ftc-cps, ftc-non-linear, ftc-diagnosis}. 
However, these techniques are effective against transient faults, environmental noise like wind or friction, and sensor noise caused by mechanical imperfections. 
They cannot be used to handle physical attacks~\cite{physical-attacks}. 

\smallskip
\noindent
\textbf{\em Resilient Hardware Designs.}
Resilient hardware techniques take the following approaches to mitigate physical attacks: (i) physical shields to deter malicious signal interference, (ii) sensor redesign, and (iii) sensor signal filter algorithms.  
Physical shields are not a viable option as they create heat build-up that affects the RAV's circuit and performance~\cite{emi-attack-drone}. 
Attack resilient sensors have been proposed to counter GPS spoofing~\cite{semperfi} and acoustic signal injection in gyroscope~\cite{fiber-gyroscope}. 
These resilient sensors require costly hardware redesign, and such techniques are only effective for attacks on GPS and gyroscope. 
Finally, sensor filtering algorithms~\cite{opticalspoofing, unrocker} also require hardware redesign, and thus they are not widely adopted~\cite{sok-sensor-security}. 

\smallskip
\noindent
\textbf{\em Attack Recovery.}
Many software-based attack recovery techniques have been proposed~\cite{srr-choi, recovery-rl, recovery-lp, pid-piper, unrocker, delorean, scvmon}. 
These techniques take the following recovery actions: 
(1) Use a learned model of sensors to minimize attack induced sensor manipulations, and employ the original controller to derive control commands~\cite{srr-choi, unrocker}. 
As we showed in Figure~\ref{fig:ablation}, RAV's original controller overcompensates under attacks and thus is not suitable for recovery~\cite{pid-piper}. 
Thus, relying on sensor measurement correction alone is not enough for attack recovery. 
(2) Learn the RAV's physical dynamics to design a specialized recovery controller~\cite{pid-piper, recovery-rl, recovery-mpc} or bound values of certain safety-critical control parameters~\cite{scvmon} to apply corrective control commands to steer the RAV towards estimated future states. 
Due to their narrow recovery focus, prior recovery techniques perform unsafe recovery. 
We compared \sysname with these techniques in \S~\ref{sec:res-comparison-recovery} and found that \sysname considerably outperforms them. 

DeLorean~\cite{delorean} proposes a technique for the diagnosis of physical attacks on RAVs to facilitate recovery. 
While both DeLorean~\cite{delorean} and \sysname use state reconstruction, they differ in their recovery approaches.  
DeLorean isolates sensors under attacks and uses state reconstruction to substitute the missing sensor inputs. 
DeLorean then relies on the original controller to recover RAVs. 
In contrast, \sysname uses a recovery control policy (RCP) instead of the original controller to ensure specification compliance and safe recovery. 
Our results show that simply using state reconstruction along with the original controller is not enough for specification compliance and safe recovery (Figure~\ref{fig:recovery-comparison}). 
\sysname uses state reconstruction to minimize attack induced manipulations for adversarial training and optimizing Deep-RL control policy for robustness under attacks. 

\section{Conclusion}
We propose \sysname, a specification aware attack recovery technique for RAVs. 
Unlike prior recovery techniques that ignore the RAV's mission specifications, and perform an expedient but unsafe recovery, \sysname ensures compliance with RAVs' mission specifications even under attacks, and ensures safe and timely operation 
similar to that in attack-free conditions. 
\sysname learns a recovery control policy using Deep-RL using a compliance-based reward structure. 
Furthermore, \sysname incorporates a state reconstruction technique to minimize attack induced sensor perturbations.
Our evaluation under different attacks on virtual RAVs finds that \sysname achieves $2X$ higher recovery success rate than five prior attack recovery techniques. 
Further, \sysname achieves $5X$ higher mission specification compliance than prior techniques. 
Finally, \sysname incurs a CPU overhead of 13.6 to 15.3\% on real RAVs. 

\section*{Acknowledgements}
This work was partially supported by the Natural Sciences
and Engineering Research Council of Canada (NSERC), the Department of National Defense (DND), Canada, and
a Four Year Fellowship from UBC. 
We also thank the anonymous reviewers of CCS 2024. 

\bibliographystyle{ACM-Reference-Format}
\bibliography{bibliography}

\appendix
\section{Research Methods}

\subsection{Temporal Logic - STL, LTL, CTL}
\label{appn:temp-logic}

Signal Temporal Logic (STL) is used to express properties of a system over continuous-time signals. 
An STL formula typically takes the form $f(x(t)) > \tau$, where $x(t)$ represents the value of the signal $x$ at time $t$, and function $f$ is applied to the signal. 
STL uses temporal operators such as $G$: Globally, $F$: Eventually, $U$: Until. 
STL can express time bound constraints, for example, $G_{[a, b]} f(x(t)) > \tau$. 
This means $f(x(t))>\tau$ is always true in the interval $[a, b]$. 

Linear Temporal Logic (LTL) is used to express the properties of a system over a linear sequence of states in discrete time.
An LTL formula typically takes the form $p_i > \tau$, where $p_i$ is a boolean predicate. 
LTL also has the same operators $G, F$, and $U$ as STL.  
LTL does not handle time bounds; it rather expresses properties over an unbounded sequence of timesteps, for example, $G(p_i) > \tau$. 
This means $p_i$ holds true at all discrete time points $t_i$. 

Computation Tree Logic (CTL) is used to express properties of a system over branching time. 
It allows the expression of complex properties that involve multiple paths originating from a single state. 
CTL uses a tree-like structure to reason about different possible future states of the system. 
\subsection{Deep-RL Reward Function}
\label{appn:reward-function}

\noindent
Algorithm~\ref{algo:reward-gen} shows the reward function generation process incorporating the compliance based reward structure. 
First, we generate the reward function template for each STL specification (Line 7-8). 
For calculating the reward for each STL specification, we monitor the conditions expressed within each STL specification (Line 11). 
For example, for $S_1$, we monitor if the RAV  is maintaining a specified distance from obstacles.
If so, we calculate the reward value $\rho_{S_1}$ based on the degree of compliance (Line 12-13) using the sigmoid framework (Line 20).
Line 13 shows the reward assigned for individual STL specifications. 
We repeat the condition monitoring and compliance based reward calculation for all the STL specifications and calculate a cumulative reward (Line 17).
The cumulative reward determines the overall satisfaction of mission specifications. 

\begin{algorithm}[!ht]
\footnotesize
\caption{Deep-RL Reward Function Generation}
\label{algo:reward-gen}
\begin{algorithmic}[1]
\State $S_n \leftarrow$ set of STL specifications
\State $RF_n \leftarrow$ corresponding reward functions
\State $S_n.c \leftarrow$ condition in the STL specification
\State $S_n.p \leftarrow$ parameters in the STL specification
\State $RF_n.\Vec{k} \leftarrow$ constants in $RF_n$

\Procedure{GenerateRewardFunction}{$S_i$}
    \State $F_{S_i}(x) \leftarrow$ RewardTemplate($S_i.c, S_i.p$) 
    \Comment{Reward function for $S_i$}
    \State \Return $F_{S_i}(x)~\cup$ GenerateRewardFunction($i \in S_n$) 
\EndProcedure

\Procedure{CalculateReward}{$S_i, x_t$}
    \State $V \leftarrow$ $S_i.c(x_t)$ 
    \Comment{Monitoring condition in $S_i$}
    \If {$V \in$ Satisfied}
    \Comment{Reward Shaping}
        \State $\rho_{S_1}$ $\leftarrow$ SIGMOID($V, R_i.\Vec{k}$)
        \Comment{Reward value of $S_i$}
    \Else
        \State $\rho_{S_1} \leftarrow 0 $
    \EndIf
    \State \Return Cumm\_Reward = normalize($\rho_{S_i} +$ CalculateReward($S_i \in S_n, x_t$))
\EndProcedure

\Procedure{Sigmoid}{$x, \Vec{k}$}
\State \Return $1 / (1 + \exp(-k_1 \times (x - k_2)))$
\Comment{Calculate degree of compliance}
\EndProcedure

\end{algorithmic}
\end{algorithm}

\subsection{Generating Actuator Commands}
\label{appn:act-commands}

Recall that \sysname outputs control command  $u'_t \in U$, where $U = \{+x,-x,+y,-y,+z,-z\}$. 
We explain how we derive low-level actuator commands (\eg throttle) from $u'_t$. 
First, we calculate the movement resolution for the RAV in the $x, y, z$ axis. 
Movement resolution means the smallest incremental adjustment that can be made to control the RAV (\ie smallest possible change in RAV's position, velocity, acceleration, and angular orientation)~\cite{gen-act-signal-1}. 
The RAV's movement resolution is denoted as $x', y', z'$.
Then, we compute RAV's earth frame velocity and angular orientation using trigonometric transformations~\cite{gen-act-signal-1, gen-act-signal-2}.  
Earth frame reference means the RAV's orientation relative to Earth's north. 
The following equations show the transformations. 

\begin{align}
    & \dot{x} = \dot{x'} \times cos(\psi) - \dot{y'} \times sin(\psi) \\
    & \dot{y} = \dot{x'} \times sin(\psi) - \dot{y'} \times cos(\psi) \\
    & \dot{z} = \dot{z'} \\
    & \psi = arctan2(\dot{y},\dot{x}) \\
    & \theta = \sin^{-1} \left( \frac{{\ddot{y}^2 + \ddot{z}^2}} {{\sqrt{\ddot{x}^2 + \ddot{y}^2} + \ddot{z}^2}} \right) \\
    & \phi = \sin^{-1} \left( \frac{{\ddot{x}^2 + \ddot{z}^2}}{{\ddot{y}}} \right)
\end{align}

These commands $\dot{x}, \dot{y}, \dot{z}, \ddot{x}, \ddot{y}, \ddot{z}$ are forwarded to the RAV autopilots (PX4, ArduPilot) as override commands.
The autopilots calculate thrust force to maneuver the RAV as per the desired change in orientation (shown in the Equations above). 

\subsection{Implementation of State Reconstruction}
\label{appn:state-recon}

We incorporate the state reconstruction technique proposed in our prior work~\cite{delorean} into our adversarial training due to its selective state reconstruction capabilities. 
This approach strategically reconstructs \sysname's input vector (shown in Equation~\ref{eqn:input}) for sensors compromised by an attack,
using trustworthy historic states, while simultaneously preserving the accuracy of states derived from unaffected sensors. 
During adversarial training, \sysname's reconstructed inputs are denoted as $x^r_{t_a}$, and the true inputs are denoted as $x'_{t_a}$ ( RAV's states without the attack's influence).   
The state reconstruction process ensures that the error between the reconstructed inputs and true inputs is minimized, such that $|x^r_{t_a} - x'_{t_a}| \rightarrow \epsilon$, thus bounding the attack-induced perturbations. 

\subsection{Multi-Agent Training}
\label{appn:multi-agent-training}

\noindent
\textbf{\em Attack Agent.}
A naive approach to introduce attacks during adversarial training would be to perturb sensor measurements at random locations for arbitrary durations during a training episode.
This naive approach has two limitations. 
(1) Limited attack coverage: the naive approach will not cover the full range of possible attacks. Consequently, \sysname will lack exposure to complex attack patterns, and stealthy attacks that gradually build up over time~\cite{stealthy-attacks}. 
(2) Limited attack scenarios: the naive approach may under-represent critical or more vulnerable segments of the mission (\eg mode changes from take off to steady state, or vulnerable scenarios with multiple obstacles around).
Consequently, \sysname may not effectively recover RAVs in these overlooked segments of the mission. 

To overcome these limitations we use a game-theoretic adversarial training approach to enhance coverage and generalizability of \sysname~\cite{rarl, adv-agent-training}. 
The main idea of our approach is that both players, namely \sysname and the \anomalyagent are trained simultaneously.  
\sysname's objective is to recover RAVs from attacks complying with the mission specifications. 
In contrast, the \anomalyagent's goal is to strategically launch attacks with varying intensity and timing to thwart \sysname.

The reward function, inputs, and outputs of \sysname are discussed in \S~\ref{sec:recovery-control-policy}.
The \anomalyagent's inputs include $u'_t$, the actions taken by \sysname, RAV's current states $x'_t$, the past sensor biases $B$ injected into sensor measurements, and the duration of the past attack: $S = \{u'_t, B_{[0..t], x'_t, t_a}\}$. 
The \anomalyagent outputs the sensor to the target, bias values to be injected for the target sensor, and the duration of the attack: $A = \{ \mathrm{target\_sensor_{1..n}}, B_{[1..n]}, t_a\}$. 
  
\begin{algorithm}[!ht]
\footnotesize
\caption{\sysname Adversarial Training}
\label{algo:adv-training}
\begin{algorithmic}[1]
\State $R_{SC} \leftarrow$ \sysname reward function
\State $u'_t \leftarrow$ \sysname's output
\State $R_{AA} \leftarrow$ \anomalyagent reward function
\State $a_t \leftarrow$ \anomalyagent's output
\Procedure{MultiAgentTraining}{$S_i$}
    \State Initialize $\pi_\theta$
    \State Initialize $\pi_\psi$
    \State $\pi_{\psi_{fixed}} = \psi$ 
    \While {$i < N$ (episodes)} 
       \State $R_{SC_i} = J(\theta,\psi_{fixed})$
       \State $\overline{R}_{SC_t} = \frac{1}{N} \sum_{i=1}^{N} R_{SC_i}$ \Comment{Cummulative \sysname reward}
       \State $\theta \leftarrow \theta - \alpha \nabla_{\theta} \overline{R}_{SC_t}$ \Comment{Update \sysname's policy}
    \EndWhile
    \State $\theta_{\theta_{fixed}} = \theta$
    \While {$i < N$ (episodes)}
       \State $R_{AA_i} = J(\psi,\theta_{fixed})$
       \State $\overline{R}_{AA_t} = \frac{1}{N} \sum_{i=1}^{N} R_{AA_i}$ \Comment{Cummulative \anomalyagent reward}
       \State $\psi \leftarrow \psi + \alpha \nabla_\psi \overline{R}_{AA_t}$ \Comment{Update \anomalyagent's policy}
    \EndWhile
    \If {$\frac{d}{dt} \overline{R}_{SC_t}| < \epsilon$ and $(\frac{d}{dt} \overline{R}_{AA_t}| < \epsilon$}
        \State $\theta* \leftarrow \theta$ 
    \EndIf
    \State \Return $\pi_\theta*$   \Comment{Optimal recovery control policy}
\EndProcedure

\end{algorithmic}
\end{algorithm}

The \anomalyagent uses the following reward structure. 
The reward is a function of the intensity of sensor bias injected $R_{disruption}$, the degree of stealthiness $R_{stealthiness}$ (\ie anomaly remains undetected for a long period), and the complexity of the attack $R_{complexity}$ (\ie launched at high-risk scenarios), where $\alpha, \beta, \gamma$ are weights that specify the objectives that the \anomalyagent should prioritize. 
For example, a high value of $\alpha$ denotes launch strong attacks. 

\begin{equation}
R_{AA} = \alpha.R_\mathrm{disruption} + \beta.R_\mathrm{stealthiness} + \gamma.R_\mathrm{complexity}
\end{equation}

Algorithm~\ref{algo:adv-training} shows the steps in adversarial training.  
We train \sysname and \anomalyagent using min-max optimization approach in game theory. 
This approach assumes a zero-sum game, meaning the total return of both players sums to 0, where the success of the \anomalyagent in disrupting the mission is directly equivalent to the \sysname's failure to comply with mission specifications. 
The \sysname aims to minimize disruptions (the mission specification violations) during a mission, while the \anomalyagent aims to maximize it. 
We use a function $J(\theta, \psi)$, to quantify mission disruption, where $\theta$ represents parameters associated with the \sysname's policy and $\psi$ represents parameters associated with the \anomalyagent's policy. 
Training involves alternating updates. First, we fix the \anomalyagent's policy while updating the \sysname's policy using gradient descent to minimize $J(\theta, \psi)$ (Line 8-13). 
Then we fix \sysname's policy while updating the \anomalyagent's policy with gradient ascent to maximize $J(\theta, \psi)$ (Line 15-19).  
If the change in average reward for both players becomes smaller over time (Line 20-21) \ie  
$(\frac{d}{dt} \overline{R}_{\text{SC}}(t))| < \epsilon$, and $(\frac{d}{dt} \overline{R}_{\text{AA}}(t))| < \epsilon$, this indicates both \sysname and \anomalyagent have learned optimal policies. 

\section{Experimental Setup}

\subsection{Reward Functions for Mission Specifications}
\label{appn:reward-func}

The following equations show the reward functions for the mission specifications in Table~\ref{tab:mission-specs} and Table~\ref{tab:ardupilot-mission-specs}. 
The constants $a$ in the following equations represent the Params column and function $f(c)$ represents the Condition for each STL specifications ($S_1..S_n$) as shown in the above tables. 
Finally, $k_1..k_n$ are constants to form a smooth piecewise function. 

\begin{align}
& \rho_(S_1) = \left\{\begin{matrix} \frac{2}{1+e^{-k_1(f(c)-a)}}-1& \mathrm{if}~0<f(c)<a\\ 1 & \mathrm{if}~f(c)>a \end{matrix}\right. \\
& \rho_(S_2) = \left\{\begin{matrix} \frac{1}{1+e^{k_2(f(c)-a)}} & \mathrm{if}~0<f(c)<a\\ 0 & \mathrm{if}~f(c)>a \end{matrix}\right. \\
& \rho(S_3) = \left\{\begin{matrix}0 & \mathrm{if}~f(c)<a\\\frac{1}{1+e^{-k_3(f(c)-a))}} & \mathrm{if}~f(c)>a \end{matrix}\right. \\
& \rho(S_4) = \left\{\begin{matrix}\frac{1}{1+e^{k_4(f(c)-a)}}& \mathrm{if}~f(c)<a\\ 0 & \mathrm{if}~f(c)>a \end{matrix}\right. \\
& \rho_(S_5) = \left\{\begin{matrix}\frac{1}{1+e^{k_5(f(c)-a)}} & \mathrm{if}~0<f(c)<a\\ 0 & \mathrm{if}~f(c)>a \end{matrix}\right. \\
& \rho_(S_6) = \left\{\begin{matrix} \frac{1}{1+e^{-k_6(f(c)-a)}}& \mathrm{if}~0<f(c)<a\\ 1 & \mathrm{if}~f(c)>a \end{matrix}\right. \\
& \rho(S_7) = \left\{\begin{matrix}0 & \mathrm{if}~f(c)<a\\\frac{1}{1+e^{-k_7(f(c)-a))}} & \mathrm{if}~f(c)>a \end{matrix}\right. \\
& \rho(S_8) = \left\{\begin{matrix}\frac{1}{1+e^{k_8(f(c)-a)}}& \mathrm{if}~f(c)<a\\ 0 & \mathrm{if}~f(c)>a \end{matrix}\right. \\
& \rho(S_9) = \left\{\begin{matrix}0 & \mathrm{if}~f(c)<a\\\frac{1}{1+e^{-k_9(f(c)-a))}} & \mathrm{if}~f(c)>a \end{matrix}\right. \\
& \rho(S_{10}) = \left\{\begin{matrix}\frac{1}{1+e^{k_{10}(f(c)-a)}}& \mathrm{if}~f(c)<a\\ 0 & \mathrm{if}~f(c)>a \end{matrix}\right. \\
& \rho(S_{11}) = \left\{\begin{matrix}0 & \mathrm{if}~f(c)<a\\\frac{1}{1+e^{-k_{11}(f(c)-a))}} & \mathrm{if}~f(c)>a \end{matrix}\right. \\
& \rho(S_{12}) = \left\{\begin{matrix}
 0 & \mathrm{if}~h<1~\mathrm{and}~v<1\\
 \frac{1}{1+e^{k_{12}(f(c)-1)}} \frac{1}{1+e^{k_{12}(f(c)-1)}} & \mathrm{if}~h>1~\mathrm{and}~v>1
 \label{eqn:s12}
\end{matrix}\right.
\end{align}

The reward function for $S_{12}$ is different in drones and rovers, Equation~\ref{eqn:s12} shows the $S_{12}$ reward function for drone, the following equation shows the $S_{12}$ reward function for rover. 

\begin{equation}
 \rho(S_{12}) = \left\{\begin{matrix}
 0 & \mathrm{if}~f(c)<1\\
 \frac{1}{1+e^{k_{12}(f(c)-1)}} & \mathrm{if}~f(c)>1
\end{matrix}\right.
\end{equation}

\subsection{Mission Specifications for ArduPilot}
\label{appn:mission-specs-ardupilot}
Table~\ref{tab:ardupilot-mission-specs} shows the mission specifications used for ArduRover and Aion Rover.  
Note that the mission specifications $S_3, S_4, S_9, S_{10}$, and $S_{11}$ in Table~\ref{tab:mission-specs} do not apply to rovers, hence these specifications are not shown in Table~\ref{tab:ardupilot-mission-specs}.

\begin{table*}[]
\centering
\footnotesize
\caption{Mission Specifications for evaluating \sysname. The Params shows the parameters to be used in enforcing the mission specifications, and the Control Parameters are parameters in ArduRover autopilot that set them.}
\begin{tabular}{l|l|l|l|l|l}
\hline
\textbf{\textbf{ID}} & \textbf{Mission Specification}  & \textbf{Control Parameters} & \textbf{Params} & \textbf{Condition}                    & \textbf{Mission Specification in STL}       \\ \hline
$S_1$     & Avoid collisions                & AVOID\_MARGIN               & 5m              & $get\_proximity(x_t)$                 & $G(get\_proximity(x_t) > 5)$                \\ 
$S_2$     & Do not veer off  a boundary     & GPS\_POS\_X, Y, Z             & 10m             & $input\_pos\_xyz(x_t)$                & $G(input\_pos\_xyz(x_t) < 10)$              \\ 
$S_5$     & Navigate through waypoints      & WPNAV\_RADIUS                & 5m              & $wp\_distance\_to\_destination(x_t)$  & $F_{[t_i, t_j]}(wp\_distance\_to\_destination(x_t) < 5)$ \\ 
$S_6$     & Maintain distance from obstacle & AVOID\_MARGIN               & 5m              & $get\_proximity(x_t)$                 & $G(get\_proximity(x_t)>5)$                  \\ 
$S_7$     & Maintain minimum velocity       & CRUISE\_SPEED\_MIN          & 5m/s            & $input\_vel_accel\_xy(x_t.velocity)$  & $G(input\_vel\_accel\_xy(x_t.velocity)>5)$  \\ 
$S_8$     & Maintain maximum velocity       & CRUISE\_SPEED\_MAX          & 12m/s           & $input\_vel\_accel\_xy(x_t.velocity)$ & $G(input\_vel\_accel\_xy(x_t.velocity)<12)$ \\ 
$S_{12}$  & Stay within geofence            & FENCE\_MARGIN               & 1m              & $check\_fence(x_t)$                   & $G(check\_fence(x_t)>1)$                    \\ \hline
\end{tabular}
\label{tab:ardupilot-mission-specs}
\end{table*}

\subsection{\sysname Training}
\label{appn:sysname-training}
We borrow the neural network architecture for control policy used in the Swift system proposed by Kaufmann \etal ~\cite{swift}. Swift is a Deep-RL based autonomous drone racing system. 
Swift system has won in drone racing contests against human champions. 
Our neural networks have 4 fully connected layers, the size of the input layer is 24, and the output layer is 6. 
The length of training was set to 150,000 steps, with 20,000 steps for $\epsilon$ greedy decaying (decreasing exploration rate). 
Note that we only borrow the neural network architecture from Swift - we use our compliance based reward structure and the generated reward functions to train \sysname. 

\subsection{Operating Environments}
\label{appn:mission-envs}

\begin{figure}[!ht]
\centering
\subfigure[Suburban]{
        \includegraphics[width=0.45\linewidth]{figures/1-sb.png} 
    }
\subfigure[Suburban]{
    \includegraphics[width=0.45\linewidth]{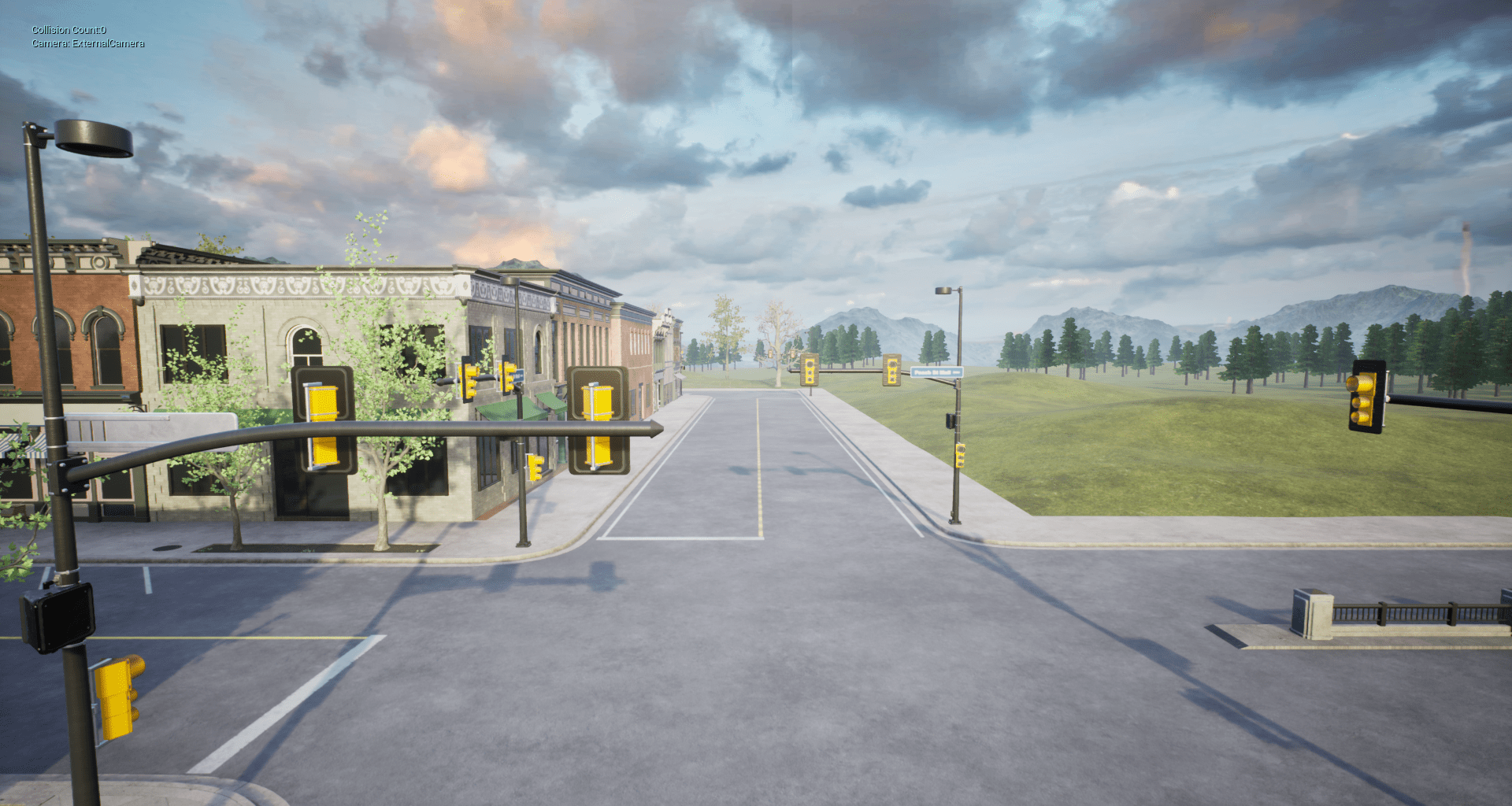}
}
\subfigure[Urban City Areas]{
        \includegraphics[width=0.45\linewidth]{figures/3-cb.png} 
    }
\subfigure[Urban City Streets]{
    \includegraphics[width=0.45\linewidth]{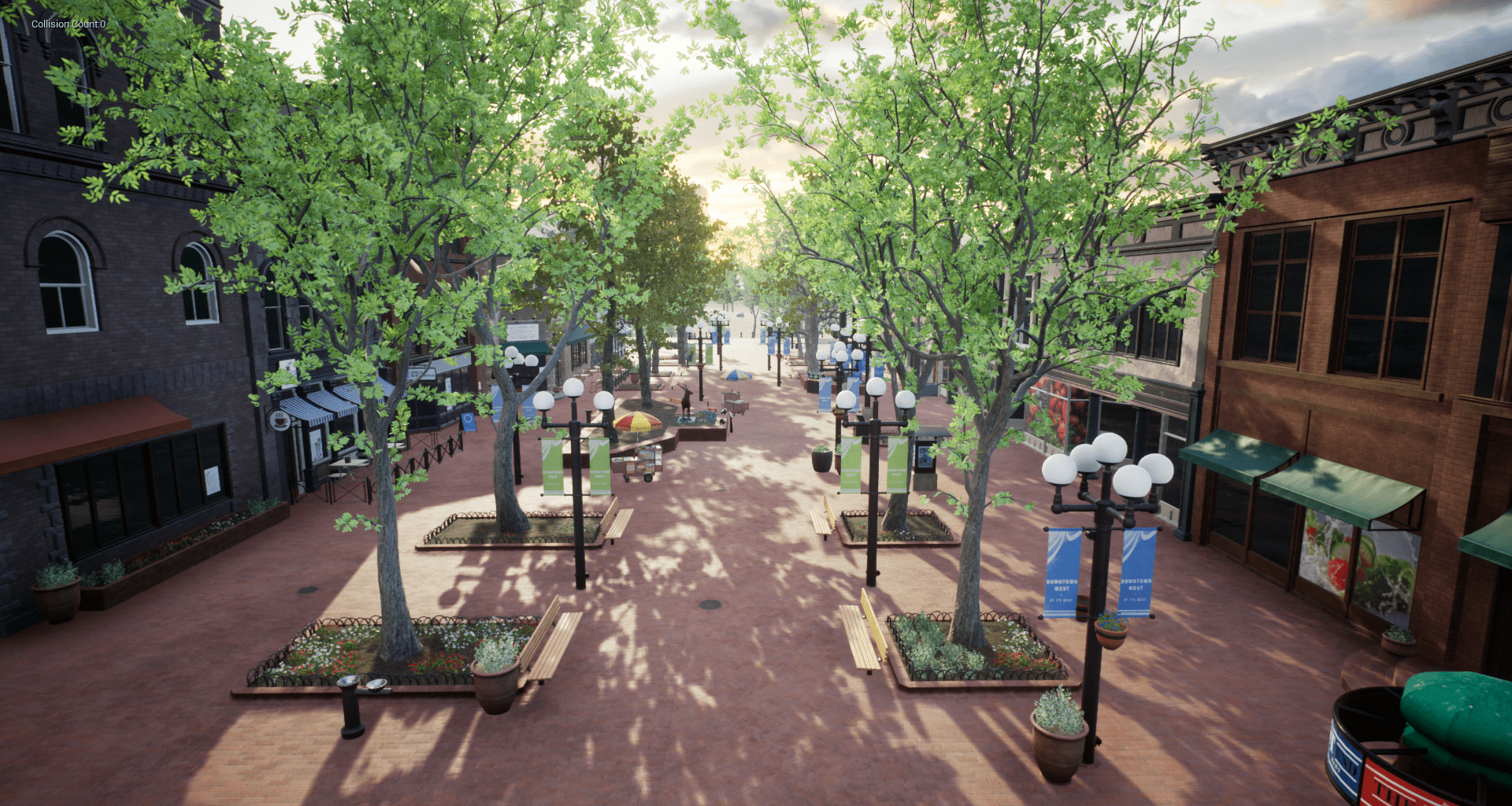}
}
\subfigure[Urban Park]{
        \includegraphics[width=0.45\linewidth]{figures/5-cp.png} 
    }
\subfigure[Urban Green Areas]{
    \includegraphics[width=0.45\linewidth]{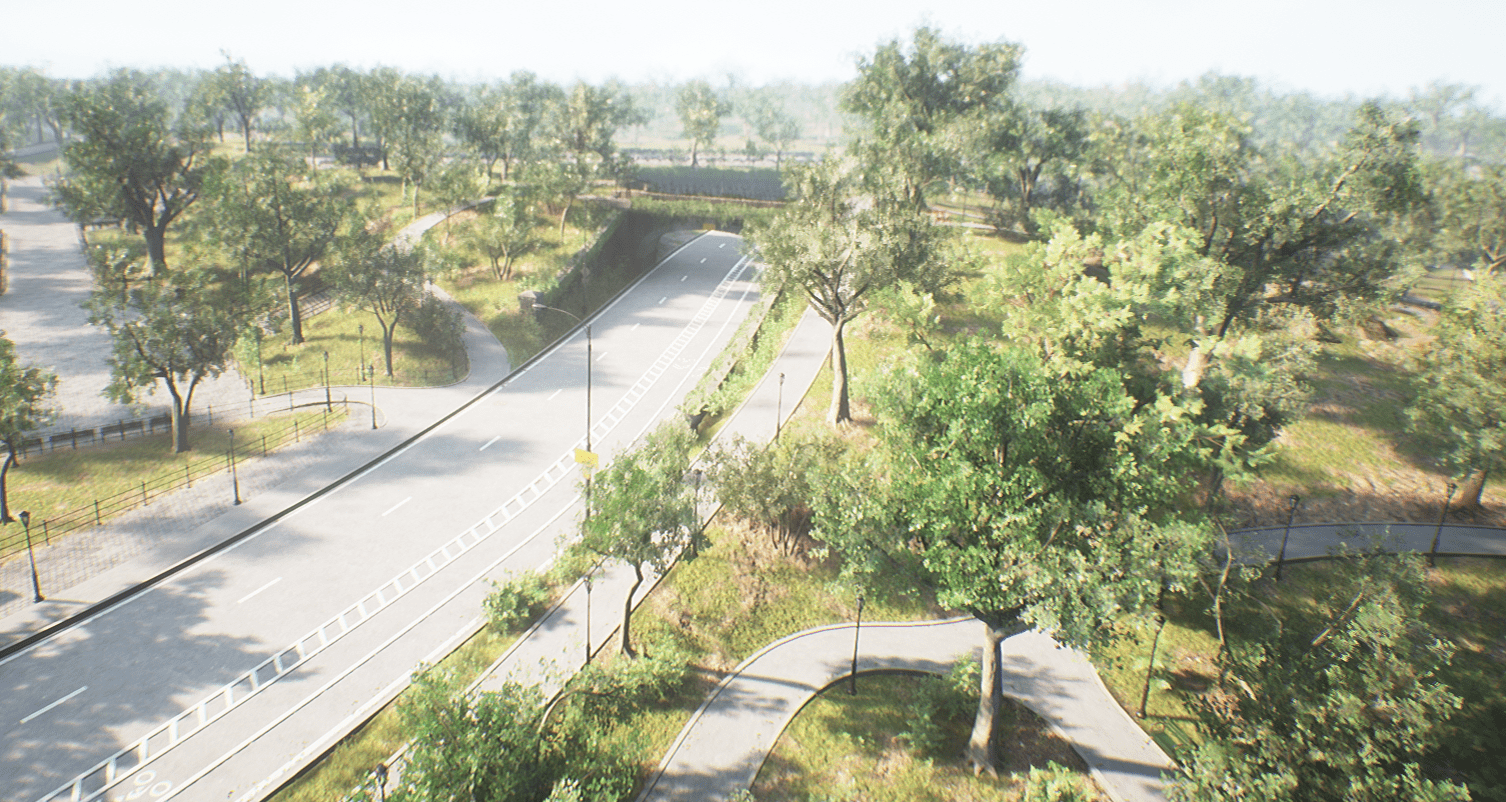}
}
\subfigure[Urban High Rise Areas]{
        \includegraphics[width=0.45\linewidth]{figures/7-dt.png} 
    }
\subfigure[Manufacturing Plant (Indoor)]{
    \includegraphics[width=0.45\linewidth]{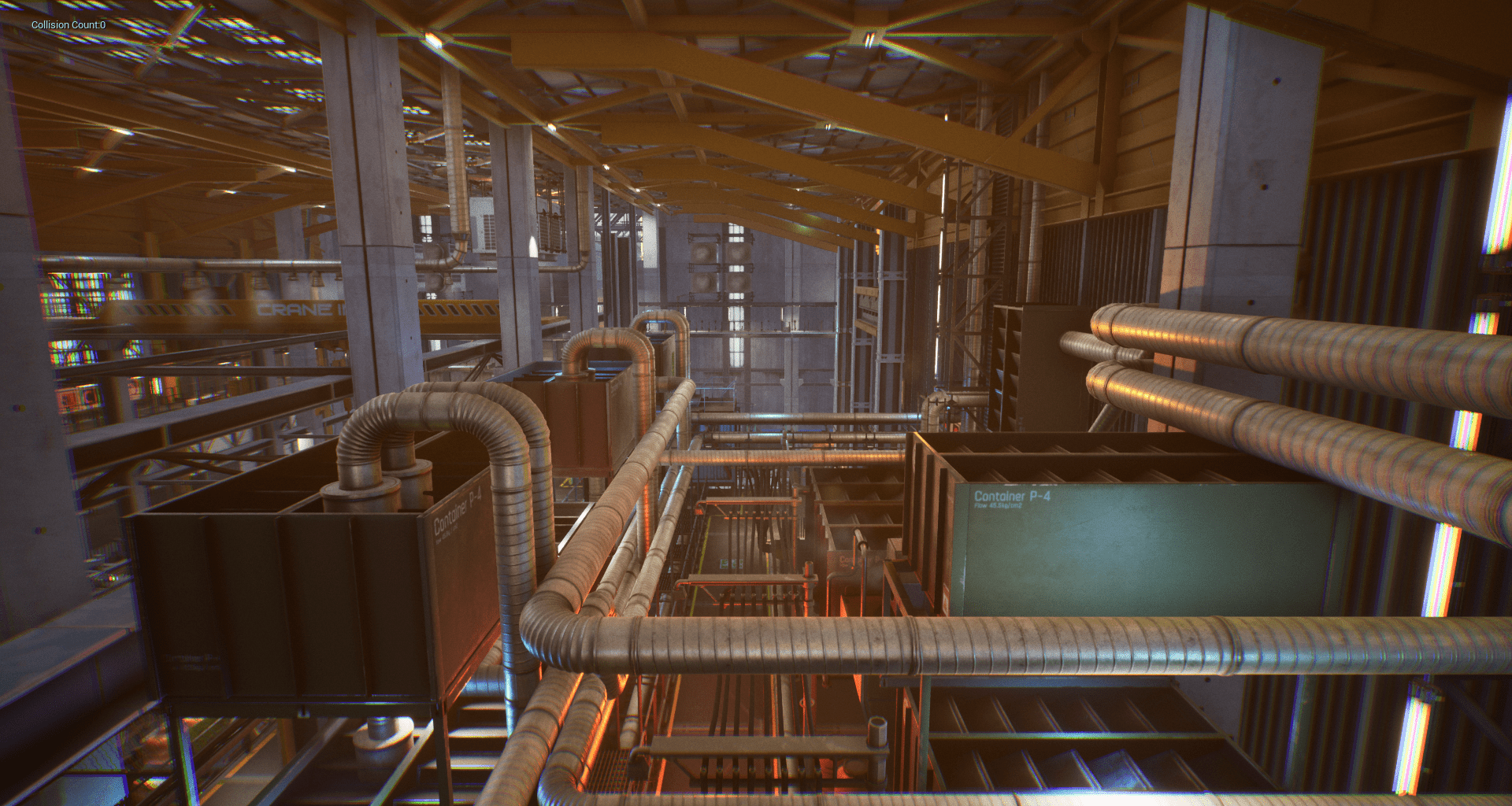}
}
\subfigure[Open Areas (Mountain)]{
        \includegraphics[width=0.45\linewidth]{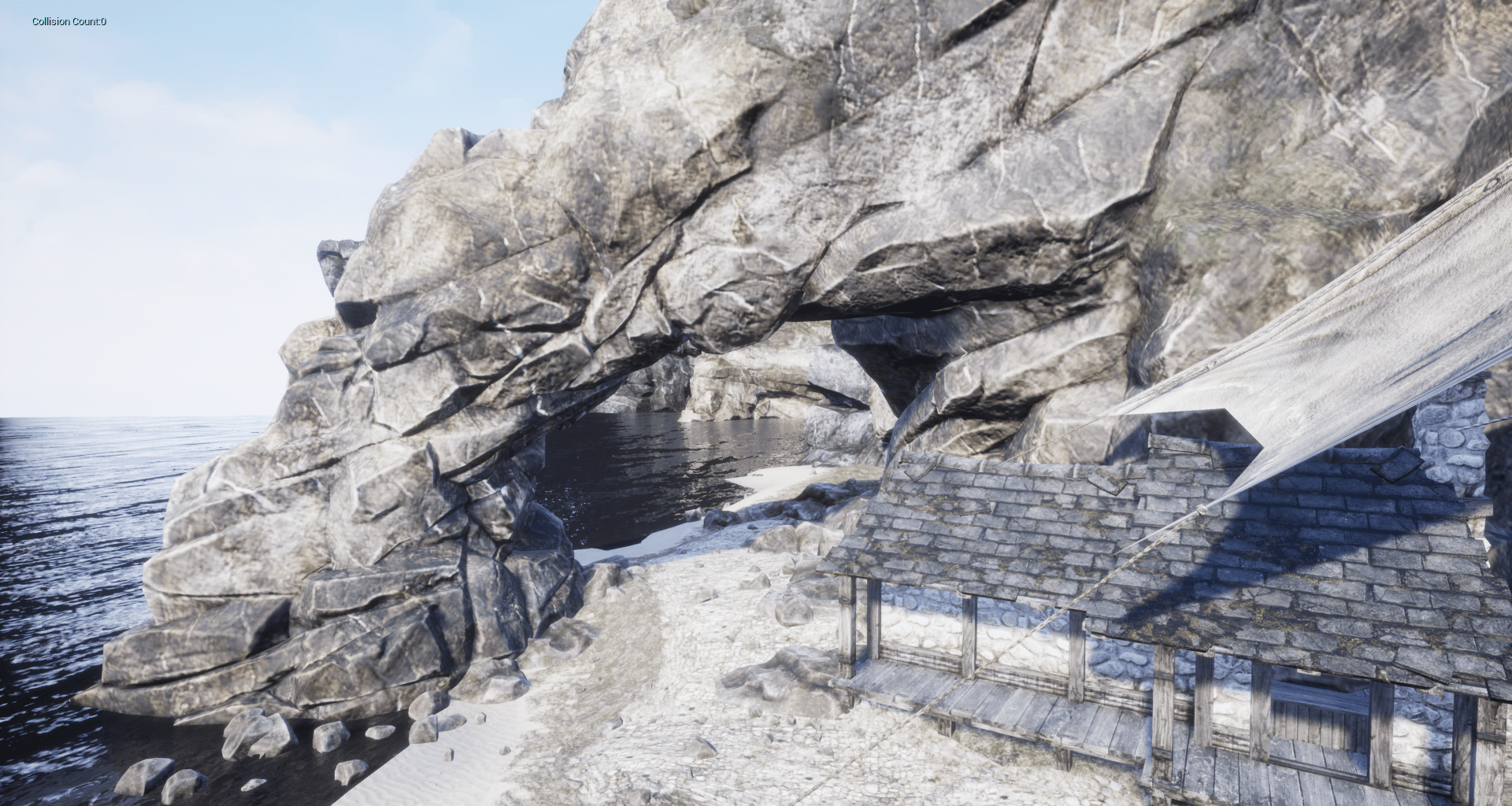} 
    }
\subfigure[Open Areas (Beach)]{
    \includegraphics[width=0.45\linewidth]{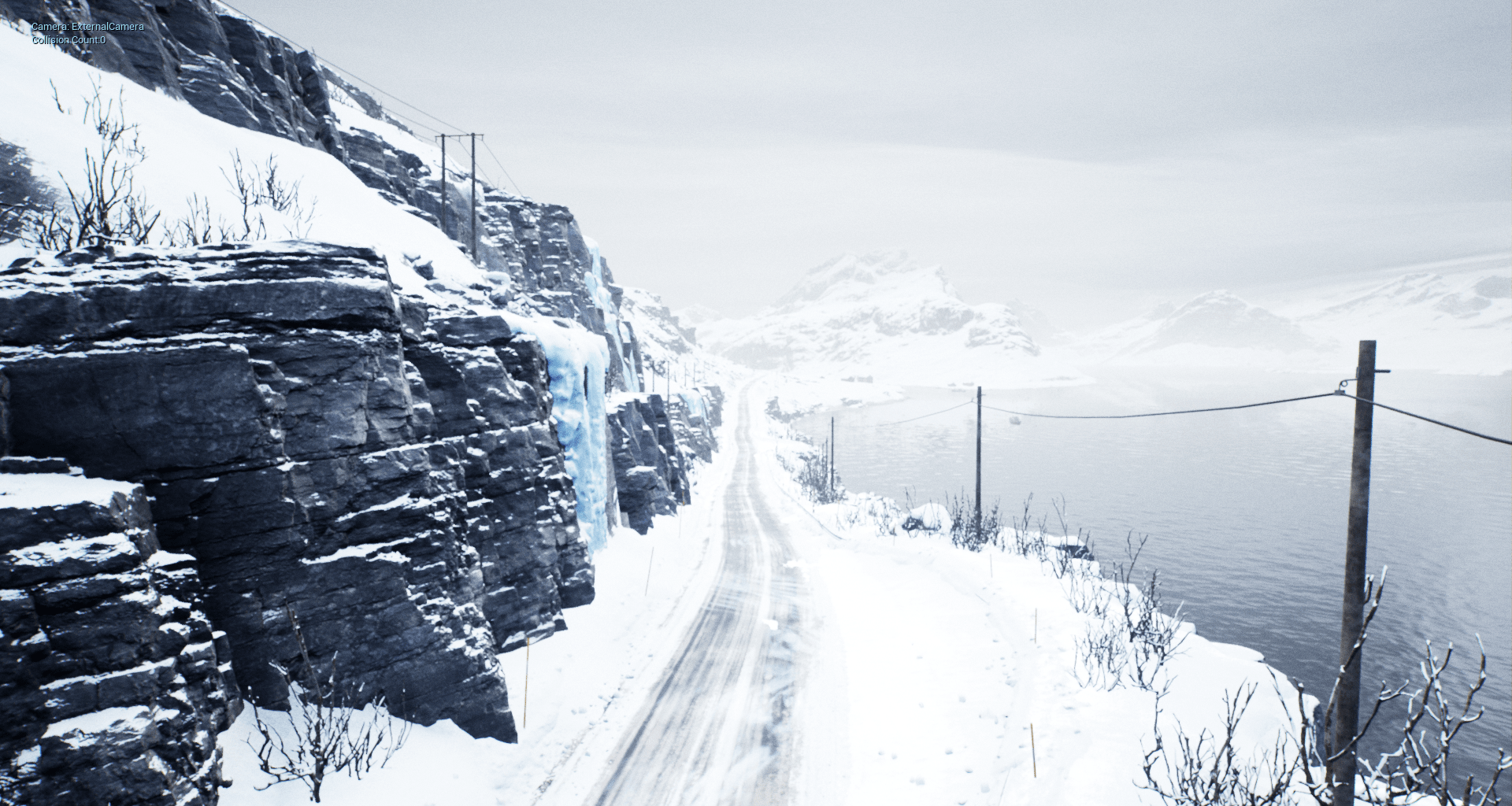}
}
\caption{\sysname is evaluated in 10 different operating environments.}
\label{fig:appn-environments}
\end{figure}

We evaluate \sysname in 10 different operating environments. These environments are open-source unreal engine scenes~\cite{unrealengine}. 
Figure~\ref{fig:appn-environments} shows a snapshot of each environment. We explain the different operating environments below. 
We run RAVs in diverse mission trajectories - straight line, circular, and polygonal paths.

\begin{itemize}
    \item Suburban Residential: This environment is a typical suburban residential area with houses and streets. It includes obstacles like trees, power lines, and electric poles. 
    \item Suburban Streets: This environment represents slightly busier suburban commercial areas. Obstacles include trees, electric poles, and power lines. 
    \item Urban City Areas: This represents typical urban areas with multi-storied buildings and commercial complexes. 
    Obstacles include trees, buildings, urban fixtures like streetlights, and traffic signals. 
    \item Urban City Streets: This represents bustling, narrow urban streets. Obstacles include trees, streetlights, office/restaurant signs, and buildings.    
    \item Urban Park: This environment represents an urban park setting with walking paths and recreational amenities. Obstacles include trees, park benches, and playground equipment.
    \item Urban Green Areas: This environment is a mix of urban streets and urban park environments, with streets cutting through parks. Obstacles include trees, park benches, streetlights, and traffic lights. 
    \item Urban High-rise Areas: Represents typical North American downtown areas with tall buildings and urban infrastructure. Obstacles include construction cranes, building structures,  urban traffic poles, and light posts. 
    \item Manufacturing Plant: This is an indoor environment that represents a typical manufacturing plant. Obstacles include machinery, storage containers, and conveyor belts. 
    \item Open Areas (Mountain): This environment represents a mountainous terrain. Obstacles include rocks and fences.  
    \item Open Areas (Beach): This environment represents a coastline. Obstacles include power lines, electric posts, and rocky cliffs.      
\end{itemize}

In environments A, B, C, D, and G, the private properties are the geofenced areas - the RAVs must not enter the private residential or commercial properties. 
In environments E, F, I, and J the region outside the operation boundaries (Table~\ref{tab:mission-specs}) are geofenced areas.

\subsection{Effectiveness of State Reconstruction}
\label{appn:res-state-recon}

We use State Reconstruction~\cite{delorean} to limit the attack induced sensor perturbations (\S\ref{sec:state-recon}).
We compare state reconstruction~\cite{delorean} with two other methods: 
(1) Sensor fusion~\cite{ekf}: that fuses measurements from multiple sensors to derive robust states under attacks.
We use the sensor fusion implementation in PX4~\cite{px4} to record the effectiveness of sensor fusion. 
(2) Sensor denoising~\cite{unrocker}: this method uses a denoising autoencoder (DAE) to filter the attack induced sensor perturbations. 
We use the sensor denoising techniques proposed by prior work~\cite{unrocker} to record the effectiveness of sensor denoising. 
We use the mean squared error (MSE) as a metric to compare all the above methods. 

\begin{table}[h]
\centering
\footnotesize
\caption{MSE of State Reconstruction in minimizing attack-induced sensor perturbations in RAV's position and attitude.}
\begin{tabular}{l|c|c|c}
\hline
\multicolumn{1}{c|}{\textbf{Metric}} & \multicolumn{1}{c|}{\textbf{\begin{tabular}[c]{@{}c@{}}Sensor \\ Fusion\end{tabular}}} & \multicolumn{1}{c|}{\textbf{\begin{tabular}[c]{@{}c@{}}Sensor \\ Denoising\end{tabular}}} & \multicolumn{1}{c}{\textbf{\begin{tabular}[c]{@{}c@{}}State \\ Reconstruction\end{tabular}}} \\ \hline
MSE - Position  &  18.91                 &   10.33                   &      3.25                     \\ 
MSE - Attitude  &  8.67                  &   4.62                    &      1.28                     \\ \hline
\end{tabular}
\label{tab:state-recon-compare}
\end{table}

We run 10 missions in virtual RAVs in various operating environments and launched overt attacks targeting RAV's GPS and gyroscope measurements.
Table~\ref{tab:state-recon-compare} shows the MSE for the position ($x, y, z$) and attitude (Euler angles $\psi, \theta, \phi$) estimations for all the methods.  
We find that state reconstruction incurs 3X lower MSE for both position and attitude estimations under attacks compared to alternative methods. 
This means that state reconstruction incurs the lowest error in deriving RAV's states under attacks and hence, is the most efficient method for limiting attack-induced sensor perturbations.  
{\em Thus, we use state reconstruction in designing \sysname.}

\section{Results}

\subsection{Proactive vs Reactive Control under Attacks}
\label{appn:mission-spec-compliance}

\begin{figure}[!ht]
\centering
\subfigure[SVR under overt attacks]{
        \includegraphics[width=0.9\linewidth]{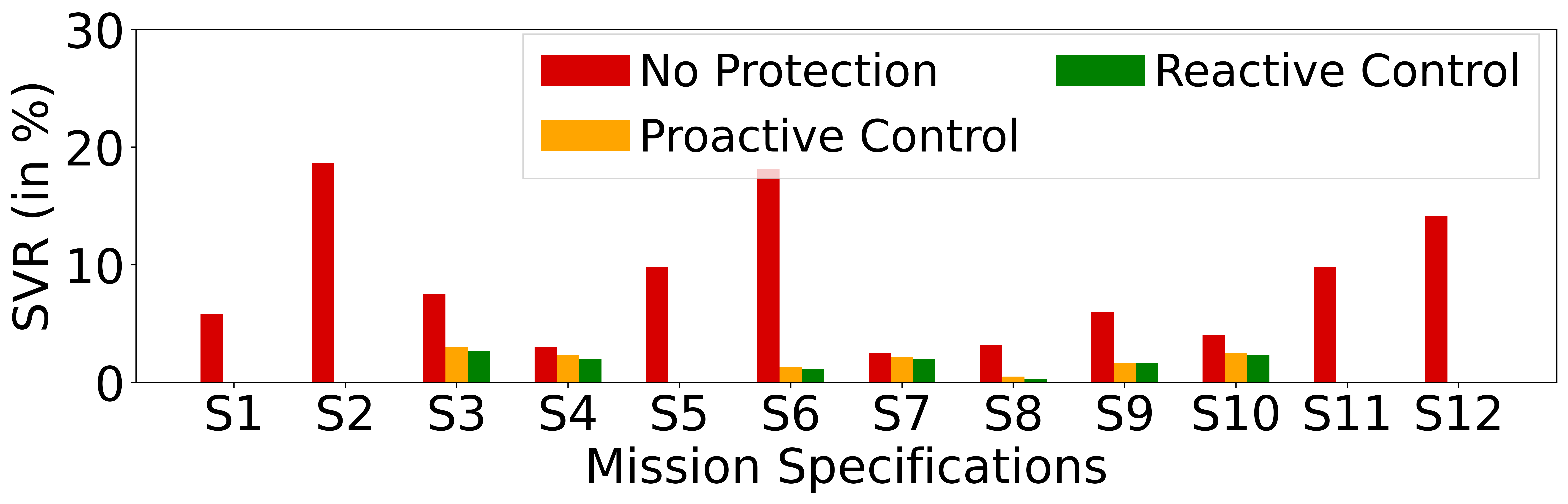} 
        \label{fig:msc-overt}
    }
    
\subfigure[SVR under stealthy attacks]{
    \includegraphics[width=0.9\linewidth]{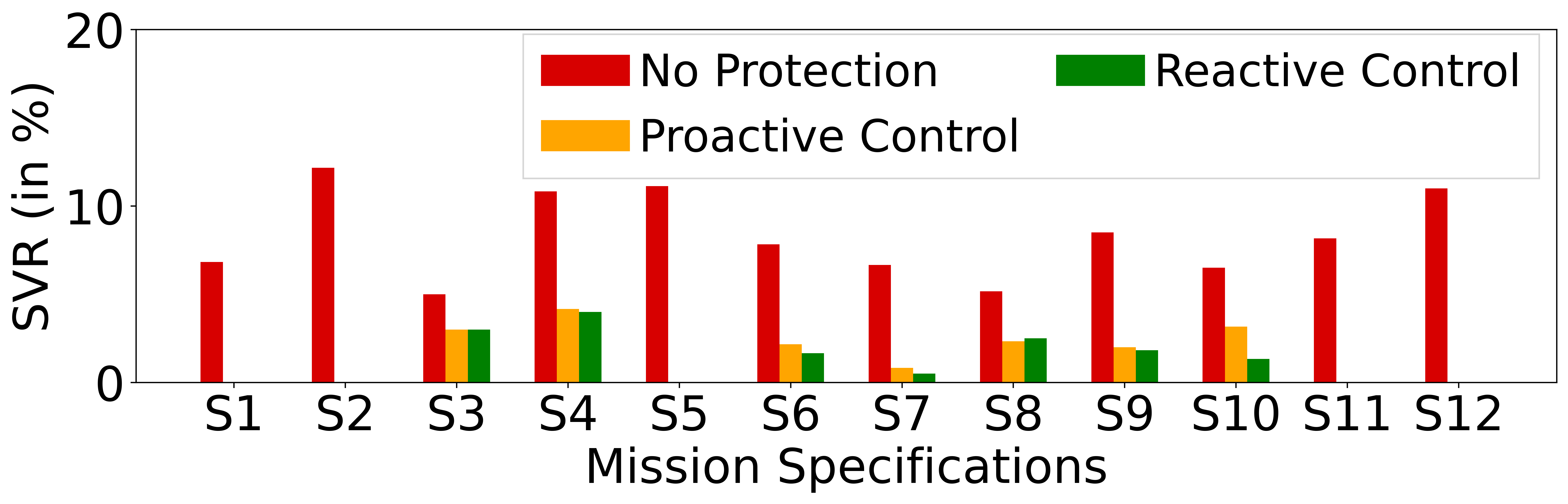}
    \label{fig:msc-stealthy}
}
\caption{SVR of Proactive Control and Reactive Control under both overt and stealthy attacks.}
\label{fig:svr-proactive-reactive}
\end{figure}

Figure~\ref{fig:msc-overt} and Figure~\ref{fig:msc-stealthy} show the distribution of SVR values observed for each mission specification under overt and stealthy attacks respectively. 
We compare the SVR when the RAV was equipped with no protection, \sysname-PC, and \sysname-RC. 

As shown in the figures, both \sysname-PC and \sysname-RC significantly minimized the mission specification violations compared to those observed without any protection. 
Furthermore, both \sysname variants prevented violations of the critical mission specifications $S_1, S_2$, and $S_{12}$.

\subsection{Adaptability in New Operating Conditions}
\label{appn:res-adaptability}

We evaluate \sysname's effectiveness in a new environment that it had not encountered in training. This is a measure of its adaptability. 
We design two experiments:  
(A) Suburban to Urban transition: \sysname is trained in suburban environments, which are less dense, and deployed in high-rise urban areas, which are more complex and dense. 
(B) Urban to Suburban transition: Conversely, \sysname is trained in high-rise urban environments and deployed in a suburban environment. 
Both A and B represent different environments from those in which the \sysname was trained. 

\begin{figure}[!ht]
\centering
\subfigure[Suburban to Urban transition]{
        \includegraphics[width=0.8\linewidth]{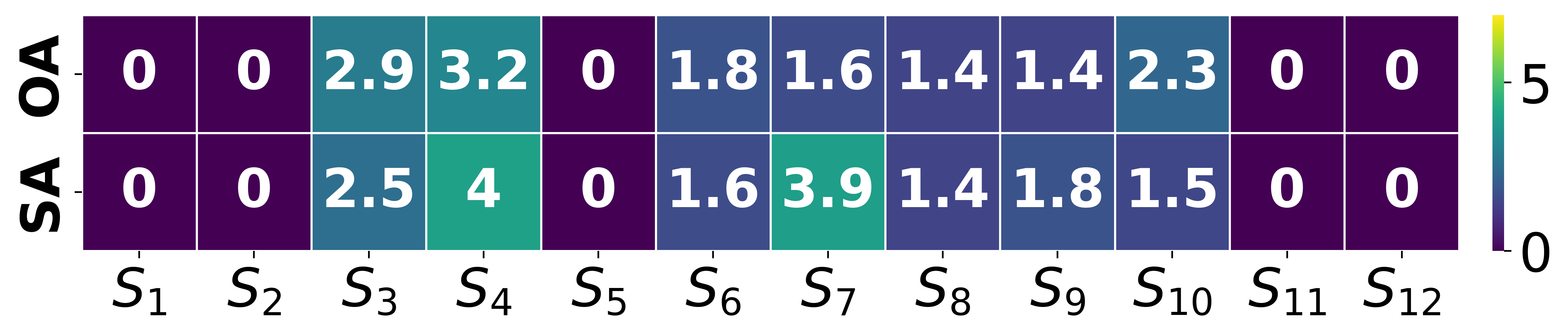} 
        \label{fig:env-seen}
    }
    
\subfigure[Urban to Suburban transition]{
    \includegraphics[width=0.8\linewidth]{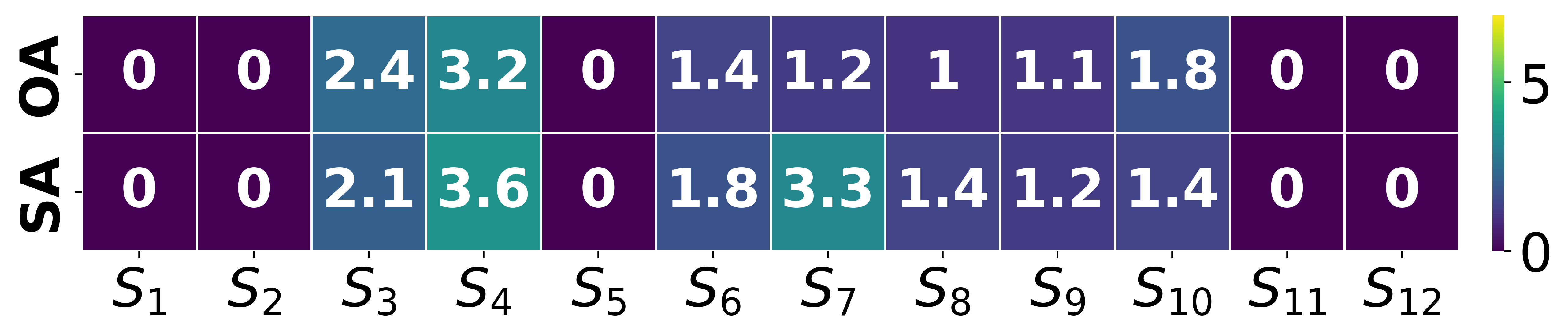}
    \label{fig:env-unseen}
}
\caption{SVR of \sysname when deployed in unseen operating conditions. 
OA: overt attack, SA: stealthy attack}
\label{fig:unseen-test}
\end{figure}

Figure~\ref{fig:unseen-test} shows a heat map of \sysname's SVR in both scenarios. The SVRs in experiment A under overt and stealthy attacks were 14.6\% and 16.7\% respectively. 
In experiment B, the SVRs under overt and stealthy attacks were 12.1\% and 14.8\%. 
The SVR is slightly higher in experiment A than B. 
This is because \sysname in Experiment A has limited exposure to dense and clustered obstacles during its training. 
For example, in Experiment A,  \sysname prevents collision with a building, but sometimes it navigates the drone approaching closer to the building when making turns (violating $S_6$). 
This situation is less prevalent in suburban environments where there are not many obstacles at street corners. 
On the other hand, in Experiment B, when navigating around trees (less prevalent in urban high-rise areas)  \sysname had to adjust the z-axis parameters to determine an optimal path, which resulted in the violation of altitude constraints ($S_3, S_4$).

Overall, \sysname is consistent in preventing mission specification violations in both experiments for the most part. 
This consistency is due to our reward structure that enables learning optimal policy, and deriving appropriate actuator commands. 
{\em Thus, \sysname adapts well to new operating environments.}

\subsection{Effort in Deploying \sysname}
\label{appn:effort}
Recall that there are five steps in \sysname design.  (\S~\ref{sec:design}), and only the first two steps require manual effort. We explain them below.

\smallskip
\noindent 
\textbf{\em Step-1: Defining mission specifications as STL.} 
The goal is to formally define mission specifications, originally expressed in natural language, using STL. 
This process involves manually identifying the components in the generic STL template: \texttt{<temporal operator (condition<parameter)>}. 
These components include the temporal operator, the boolean operator, the conditions to monitor, and constant parameters.

The first step is to define the condition that requires monitoring in the mission specification. For example, if the specification is to avoid collisions, the RAV must maintain a safe distance from the surrounding obstacles. 
The condition to monitor the RAV's distance from obstacles is defined as $obstacle\_distance > \tau$.
where $\tau$ is a constant parameter. 
Next, depending on whether the condition should always be true or only needs to be true once, an appropriate temporal operator is selected to complete the STL specification.
For the above mission specification, temporal operator $G$ is used to express the STL specification as $G(obstacle\_distance > \tau)$ because the RAV must always avoid collisions.

\smallskip
\noindent
\textbf{\em Step-2: Monitor conditions in STL.} 
In this step, we search through the API documentation of the RAV's autopilot software (PX4~\cite{px4} or ArduPilot~\cite{ardupilot}) to identify an API call that can monitor the condition in the STL specification.   
The function $get\_proximity(x_t)$ in ArduRover calculates the RAV's distance from obstacles. 
Thus, the STL specification is completed as $G(get\_proximity(x_t) > \tau)$, where $x_t$ is the RAV's current state.



\subsection{Improving \sysname's Effectiveness}
\label{appn:improving-effectiveness}
\begin{figure}[!ht]
	\centering
	\subfigure[]{
		\includegraphics[width=0.45\linewidth]{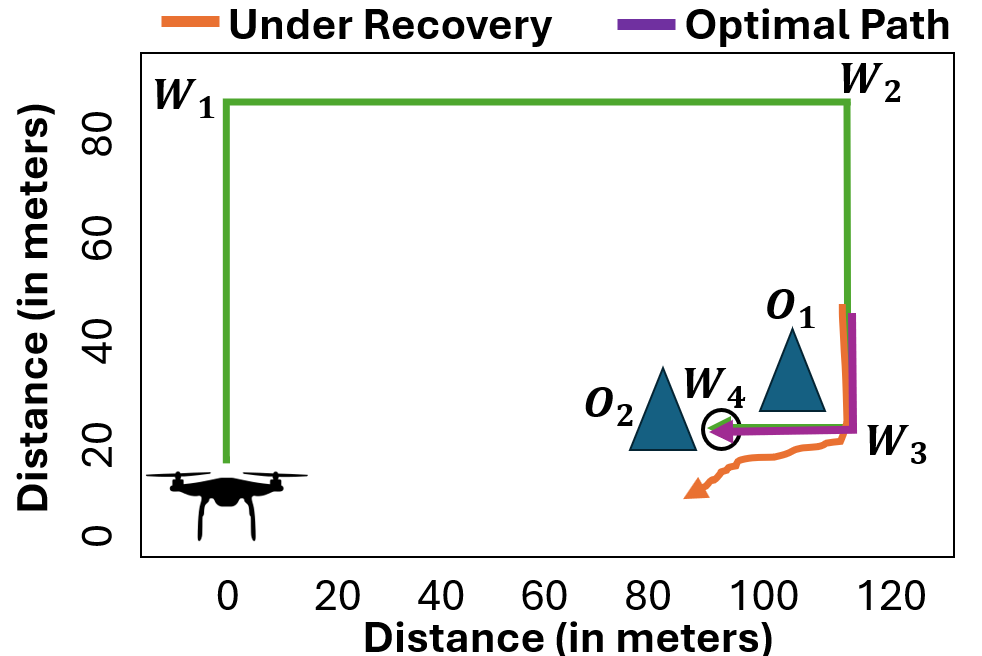} 
		\label{fig:improvement-1}
	}
	\subfigure[]{
		\includegraphics[width=0.45\linewidth]{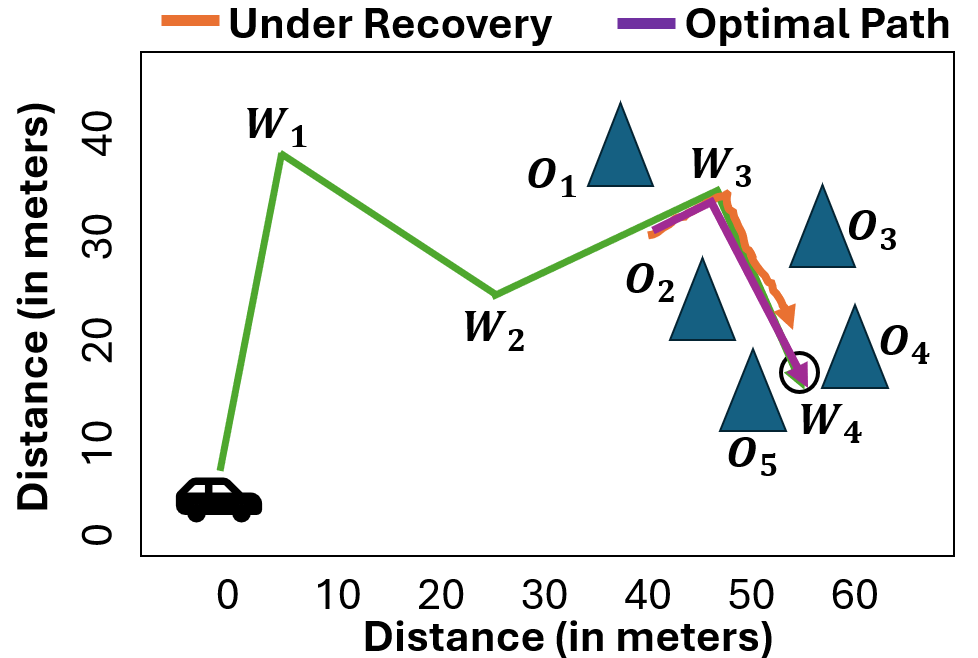}
		\label{fig:improvement-2}
	}
	\caption{Example of scenarios where \sysname failed to maneuver RAVs close to its target. The circle shows the final target, triangles are obstacles, and the mission path is shown in green line. 
		(a) Drone mission with obstacles near target,  
		(b) Rover mission with obstacles clustered near target.}
	\label{fig:improvements}
\end{figure}

Figure~\ref{fig:improvements} shows two examples where \sysname's recovery failed - \sysname could not maneuver the RAV close to its target. We discuss both examples in detail, and explain how \sysname's effectiveness can be improved in such scenarios. 

Figure~\ref{fig:improvement-1} shows a PXCopter drone mission covering waypoints $W_1$, $W_2$, $W_3$, and finally reaching the target $W_4$. 
There are two obstacles near the target $O_1$ and $O_2$. 
A GPS sensor attack was launched on the drone while it was navigating between $W_2$ and $W_3$. 
As shown in Figure~\ref{fig:improvement-1}, \sysname successfully recovered the drone (orange line) and guided it to $W_3$. 
The attack persisted as the drone navigated towards $W_4$. 
\sysname maintained compliance with mission specifications and maneuvered the drone towards $W_4$. 
However, to avoid a collision with obstacles $O_1$ and $O_2$, \sysname maintained a safe distance from them. 
Consequently, the drone landed 7 meters away from the target $W_4$. 
This resulted in a mission failure because the position error from $W_4$ exceeded 5 meters (standard GPS offset~\cite{gps-offset}).

Figure~\ref{fig:improvement-2} shows a ArduRover mission covering waypoints $W_1$, $W_2$, $W_3$, and finally, reaching the target $W_4$. 
There are four obstacles clustered near the target $O_2$, $O_3$, $O_4$, and $O_5$. 
An attack targeting the GPS was launched when the rover was navigating towards $W_3$.
\sysname successfully recovered the rover, maintained mission specification compliance, and maneuvered the rover along the mission trajectory to $W_3$.
However, as the attack persisted during the rover's approach to $W_4$, \sysname maintained a safe distance from the cluster of obstacles and could not maneuver the rover within a  5m distance of $W_4$. 
Consequently, the mission concluded with the rover 8.2 meters away from $W_4$, resulting in a mission failure.

\sysname's recovery can be improved in the above scenarios by using concepts from imitation learning~\cite{imitation-learning, imitation-policy}. 
This includes collecting data that demonstrates the optimal path in a few scenarios where \sysname failed to maneuver the RAV close to the target. 
The purple line in Figure~\ref{fig:improvement-1} and Figure~\ref{fig:improvement-2} shows the optimal path in the above two examples. 
This data serves as supervision for \sysname on how to maneuver RAVs close to the target by imitating the optimal path. 
We will consider this in future work. 

\end{document}